\documentclass[10pt,twocolumn,a4paper]{article}

\usepackage[top=2.0cm, bottom=2.0cm, left=1.8cm, right=1.8cm, columnsep=0.7cm]{geometry}
\usepackage{times}
\usepackage{microtype}

\usepackage{graphicx}
\usepackage{amsmath}
\usepackage{amssymb}
\usepackage{booktabs}
\usepackage{multirow}
\usepackage{subcaption}
\usepackage[dvipsnames]{xcolor}
\usepackage{verbatim}

\usepackage{authblk}

\usepackage{hyperref}
\hypersetup{
    colorlinks=true,
    linkcolor=RoyalBlue,
    citecolor=RoyalBlue,
    urlcolor=RoyalBlue
}

\newcommand{\supptab}[1]{Table~\ref{#1}}
\newcommand{\suppfig}[1]{Figure~\ref{#1}}
\newcommand{\suppsec}[1]{Section~\ref{#1}}
\newcommand{\inSuppTab}[1]{(Supplementary Material, Table~\ref{#1})}

\newcommand{\inSuppSec}[1]{(Supplementary Material, Section~\ref{#1})}

\title{\textbf{\Large Human Pose Estimation by Probabilistic Mixtures of Gaussian and Uniform Distributions}}

\author[1]{\textbf{Santiago Reina Alguacil}\thanks{Corresponding author: \texttt{santiagoreina@uma.es}}}
\author[1,2,3]{\textbf{Ezequiel L\'{o}pez Rubio}}

\affil[1]{\small Department of Computer Languages and Computer Science, University of M\'{a}laga, 29071 M\'{a}laga, Spain}
\affil[2]{\small Instituto de Investigaci\'{o}n Biom\'{e}dica de M\'{a}laga y Plataforma en Nanomedicina -- IBIMA Plataforma BIONAND, 29590 M\'{a}laga, Spain}
\affil[3]{\small ITIS Software, University of M\'{a}laga, 29071 M\'{a}laga, Spain}

\date{}

\begin{document}

\twocolumn[
  \begin{@twocolumnfalse}
    \maketitle
    \begin{abstract}
      Despite advancements in sub-pixel decoding, modern Human Pose Estimation (HPE) architectures remain constrained by deterministic coordinate extraction from discrete heatmaps, which systematically fails under dense occlusions and Out-of-Distribution (OoD) ambiguities. To address these limitations, we propose a post-hoc continuous probabilistic framework that intercepts raw heatmaps and models them as a Gaussian Mixture Model (GMM) augmented with a spatially agnostic uniform distribution, requiring no architectural modifications or retraining. This continuous representation allows us to comprehensively explore the distinct contributions of density modeling to HPE. We show that the spatial expectation of our fitted GMM achieves precision parity with leading sub-pixel decoders without heuristic post-processing. Leveraging this continuous space, our test-time augmentation aggregates probability distributions rather than discrete activations, mitigating multi-scale heatmap poisoning. Furthermore, because our multimodal GMM parameterization isolates competing spatial hypotheses, it naturally interfaces with geometric regularizers such as Markov Random Fields. However, our analysis exposes the inherent vulnerabilities of 2D kinematic priors under perspective foreshortening and visual degradation. Crucially, this density formulation yields significant improvements in uncertainty quantification: our calibrated geometric covariance volume outperforms standard peak-intensity heuristics in OoD anomaly detection under severe occlusion. For in-distribution calibration, dynamically fusing our geometric metric via Adaptive Softmax reduces or matches baseline Area Under the Sparsification Error (AUSE) across benchmarks and degradation levels. Finally, we establish that this continuous parameterization is extensible to multi-view 3D architectures, bypassing the $\mathcal{O}(N^3)$ voxel memory bottleneck.
    \end{abstract}
    \vspace{0.8em}
    \textbf{Keywords:} Human Pose Estimation, Gaussian Mixture Models, Uncertainty Quantification, Out-of-Distribution Detection, Deep Learning.
    \vspace{1.8em}
  \end{@twocolumnfalse}
]

{
  \renewcommand{\thefootnote}{*}
  \footnotetext{Corresponding author: \texttt{santiagoreina@uma.es}. This research was conducted as part of the Bachelor's Thesis of Santiago Reina Alguacil at the University of M\'{a}laga, under the supervision of Ezequiel L\'{o}pez Rubio.}
}

\section{Introduction}
The evolution of Human Pose Estimation (HPE) has been defined by a continuous pursuit of higher spatial fidelity, transitioning from classical pictorial structures \cite{yang2011articulated} to deep Convolutional Neural Networks (CNNs) \cite{he2016deep} and Vision Transformers \cite{xu2022vitpose} capable of dense pixel-wise regression. At the core of this paradigm shift is the spatial heatmap: a two-dimensional grid representing the unnormalized likelihood of a keypoint's location over the input image space. To maximize coordinate fidelity, the research community has engineered architectures, such as the High-Resolution Network (HRNet) \cite{sun2019deep}, which actively maintain dense, high-resolution representations throughout the network depth to prevent the loss of spatial context associated with successive pooling operations.

Parallel to these architectural advancements, the decoding algorithms utilized to extract continuous spatial coordinates from discretized grid predictions have become increasingly mathematically sophisticated. Recognizing that the argmax operation introduces systematic quantization errors bounded by the downsampling stride of the network, state-of-the-art sub-pixel decoders such as Distribution-Aware Coordinate Representation of Keypoint (DARK) \cite{zhang2020distribution} and Unbiased Data Processing (UDP) \cite{huang2020devil} have been developed. These decoders utilize Taylor-series expansions and local Hessian approximations to shift the predicted coordinate to sub-pixel accuracy, pushing deterministic HPE to its absolute geometric limits.

However, this relentless pursuit of absolute spatial precision masks a mathematical flaw: the inherent inability of rigid, deterministic point-estimates to capture structural ambiguity and calibrate spatial uncertainty. By forcing a complex, continuous probability distribution to collapse into a singular coordinate paired with a heuristic maximum activation value, conventional decoding pipelines permanently discard the network's epistemic uncertainty. In canonical, highly visible scenarios featuring isolated individuals, this loss of topological information is relatively benign. Yet, in chaotic real-world environments characterized by severe multi-person occlusions, motion blur, and out-of-frame body parts, the visual evidence is deeply ambiguous. Under these degraded conditions, the network's true spatial belief is often multimodal (e.g., exhibiting symmetric confusions between left and right limbs) or highly diffuse with elevated entropy. Deterministic point-regression fundamentally fails to represent these complex topological states, leading to overconfident errors where the model confidently locks onto the anatomical joints of neighboring individuals.

To overcome this structural bottleneck, we conceptually pivot from deterministic coordinate extraction to continuous probabilistic modeling. We propose intercepting the raw activation heatmaps prior to discrete thresholding, transforming and modeling them explicitly as continuous Gaussian Mixture Models (GMM) augmented with a spatially agnostic Uniform distribution. This formulation explicitly parameterizes the spatial topography of the network's belief, retaining both heteroscedastic uncertainty and multimodal ambiguity. By translating rigid pixels into a continuous parametric space, we embark on a comprehensive exploration to uncover exactly the distinct capabilities enabled by this continuous density representation for modern HPE, structured along two central axes: coordinate precision and uncertainty quantification..

\textbf{Precision and Spatial Structure:} We first establish that transitioning to a continuous probabilistic parameter space does not sacrifice raw coordinate accuracy; the mathematical expectation of our solitary GMM matches the precision of highly optimized decoders like DARK and UDP. Leveraging this continuous domain, we redefine Test-Time Augmentation (TTA). Unlike standard discrete heatmap averaging---which is vulnerable to out-of-bounds ``heatmap poisoning'' when scale augmentations zoom past peripheral keypoints---our continuous aggregation mechanism acts as a scale-aware spatial noise filter. Furthermore, because our framework explicitly models multimodal density landscapes—isolating distinct spatial modes under ambiguity—it naturally interfaces with global structural priors. We incorporate skeletal kinematic constraints via Markov Random Fields (MRF) \cite{yang2011articulated} to enforce global skeletal kinematics, critically analyzing its behavior to reveal a fundamental trade-off: while MRFs act as excellent sub-pixel regularizers under moderate noise, they forcefully propagate anatomical inversions under severe visual degradation due to the strict limitations of 2D projective foreshortening. Finally, we theoretically establish that our parametric representation is naturally extensible to multi-view 3D architectures, offering a direct mathematical pathway to triangulate spatial rays in continuous space, entirely bypassing the $\mathcal{O}(N^{3})$ computational memory curse inherent to dense volumetric voxel grids.

\textbf{Uncertainty Quantification and Calibration:} Beyond coordinate precision, our continuous parameterization provides geometric tools for uncertainty quantification across both Out-of-Distribution (OoD) and In-Distribution regimes. By computing the kinematic dispersion of the Gaussian components, we formulate a geometric metric: the Covariance Volume. In OoD anomaly detection, tracking this geometric volume substantially outperforms heuristic peak confidence \((1 - P_{\text{DARK}})\) when identifying true anatomical absences under severe occlusion, avoiding the confidence collapse common to standard baselines. For In-Distribution calibration, we resolve the sensitivity trade-off between confident and corrupted predictions by dynamically fusing our geometric volume with standard peak activations via an adaptive softmax scheme. This formulation consistently matches or reduces the Area Under the Sparsification Error (AUSE) across benchmarks and degradation levels, delivering a robust, parameter-free inference pipeline.

In summary, the main contributions of this work are:

\begin{itemize}
    \item \textbf{Continuous Probabilistic Decoding and TTA:} We reformulate discrete heatmap decoding into a continuous GMM with a uniform background component, achieving precision parity with state-of-the-art sub-pixel decoders while enabling a scale-aware density aggregation for TTA that eliminates discrete heatmap poisoning.
    \item \textbf{Limits of 2D Geometry:} While integrating a Markov Random Field exploits local multimodal uncertainty to improve overall coordinate precision, 2D kinematic constraints face a topological ceiling, remaining vulnerable to projective ambiguities and symmetric limb swaps under severe occlusion.
    \item \textbf{Geometric Uncertainty and Calibrated Inference:} We establish the Covariance Volume as a reliable metric for OoD anomaly detection under dense occlusions. Paired with our adaptive softmax fusion for in-distribution calibration, this framework systematically optimizes sparsification error (AUSE) without retraining or architectural modifications.
\end{itemize}
\section{Related Work}
The pursuit of robust and highly accurate Human Pose Estimation spans multiple algorithmic domains, from foundational architectural design to advanced sub-pixel decoding and uncertainty quantification. In this section, we review the academic literature across these complementary domains, demonstrating how our continuous GMM + Uniform parameterization bridges the structural gap between rigid deterministic decoders and unconstrained probabilistic regression models.

\subsection{High-Resolution HPE and Sub-pixel Decoding}
The architectural foundation of modern 2D Human Pose Estimation relies on deep neural networks designed to process and output high-resolution spatial feature maps. Early frameworks utilized encoder-decoder structures with successive spatial pooling, which compromised the spatial fidelity necessary for precise joint localization. To address this, the High-Resolution Network (HRNet) \cite{sun2019deep} introduced a paradigm where high-resolution representations are maintained continuously throughout the network, operating parallel to lower-resolution convolutions. This multi-scale fusion allows HRNet to capture both global semantic context and local spatial details, establishing it as a dominant backbone for top-down HPE. More recently, Vision Transformers (ViTs) \cite{xu2022vitpose} have been adapted for pose estimation. Architectures like ViTPose employ plain, non-hierarchical vision transformers to extract features, demonstrating exceptional scalability (up to 1 billion parameters) and setting new state-of-the-art benchmarks on datasets like MS COCO without relying on complex, task-specific structural designs.

Despite these powerful feature extractors, the final output remains a discretized heatmap, which introduces inherent quantization errors when coordinates are extracted via a simple argmax operation. To mitigate this, significant research has focused on post-processing sub-pixel decoders. The Distribution-Aware Coordinate Representation of Keypoint (DARK) \cite{zhang2020distribution} codec addresses this quantization by assuming the underlying distribution follows a 2D Gaussian profile. DARK applies a Gaussian smoothing kernel and utilizes a Taylor-series expansion to approximate the local Hessian matrix around the maximal activation, calculating a sub-pixel shift based on the first and second-order derivatives. Unbiased Data Processing (UDP) \cite{huang2020devil} identified systematic statistical errors in how coordinates are transformed between the image and heatmap spaces. UDP formulates a principled strategy utilizing unit-length-based measurement in continuous space, effectively eliminating the asymmetric spatial drift that plagues standard encoding and decoding pipelines.

While DARK and UDP represent the state-of-the-art in deterministic sub-pixel coordinate extraction, they maximize precision at the direct expense of continuous uncertainty. By explicitly forcing the heatmap to collapse into a single spatial coordinate and discarding the surrounding probability mass, these sub-pixel decoders ignore the complex, multimodal nature of the network's prediction \cite{calibration2023hpe}. If the network predicts two distinct peaks, indicating hesitation between a left and right wrist, DARK and UDP will simply refine the slightly higher peak and discard the second, projecting false confidence. Our framework intercepts the heatmap prior to this deterministic collapse, utilizing Monte Carlo sampling and Gaussian Mixture Modeling to recover precision while preserving the complete multimodal spatial dispersion of the network's belief.
As summarized in Table \ref{tab:sota_comparison}, our proposed continuous formulation bridges a gap in the literature, explicitly transitioning the output paradigm from deterministic sub-pixel coordinates to multimodal probability distributions equipped with geometric Covariance Volume metrics for robust OoD anomaly detection.

\begin{table*}[htbp]
\centering
\caption{Comparison of Uncertainty and Output Paradigms in HPE Frameworks. Our proposed continuous formulation bridges a critical gap in the literature, explicitly transitioning the output paradigm from deterministic point-estimates to multimodal probability distributions with robust OoD rejection mechanisms.}
\label{tab:sota_comparison}
\resizebox{\textwidth}{!}{%
\begin{tabular}{@{}lcccc@{}}
\toprule
\textbf{Feature/Capability} & \textbf{Standard} & \textbf{DARK/UDP} & \textbf{RLE} & \textbf{Proposed (GMM + Uniform)} \\
\midrule
Output Paradigm & Discrete Pixel & Sub-pixel Coordinate & Sub-pixel Coordinate & Continuous Distribution \\
Quantization Mitigation & None & Hessian / Taylor Exp. & Normalizing Flows & MC Sampling \& Uniform Jitter \\
Uncertainty Output & Peak Value ($P_{max}$) & Peak Value ($P_{DARK}$) & Unimodal Variance & Multimodal Covariance Volume \\
Multimodal Awareness & No & No & No & Yes (Explicit) \\
OoD Rejection & Poor (Heuristic) & Poor (Heuristic) & Moderate & High (Covariance Volume) \\
\bottomrule
\end{tabular}%
}
\end{table*}

\subsection{Uncertainty Quantification in HPE}
Recognizing the limitations of heuristic confidence scores (such as the maximum heatmap activation), recent literature has increasingly focused on uncertainty quantification in computer vision. Bayesian Deep Learning partitions uncertainty into two distinct categories: aleatoric uncertainty, which models the inherent observation noise (e.g., motion blur, occlusion), and epistemic uncertainty, which captures model ignorance or lack of data \cite{kendall2017uncertainties, bramlage2023plausible}. In deep networks, epistemic uncertainty is typically estimated via Monte Carlo Dropout \cite{gal2016dropout} or Deep Evidential  Regression \cite{amini2020deep}, while aleatoric uncertainty is learned directly by predicting the variance of a parameterized distribution alongside the coordinate.

In the domain of Human Pose Estimation, the shift towards probabilistic modeling has gained significant traction. Regression-based methods, which map input images directly to joint coordinates, are computationally efficient but traditionally suffer from inferior precision compared to heatmap-based approaches due to the spatial abstraction of fully connected layers. To close this gap, Residual Log-likelihood Estimation (RLE) \cite{li2021human} explores Maximum Likelihood Estimation (MLE) to capture the underlying output distribution of regression networks. Rather than predicting raw coordinates using an arbitrary L1 or L2 loss, RLE utilizes Normalizing Flows to learn the residual change relative to a tractable initial density function (typically a Gaussian or Laplace distribution). This allows the network to dynamically adapt its loss function based on the learned heteroscedastic uncertainty of the data, significantly boosting direct regression performance without introducing test-time overhead.

However, despite these advancements, methodologies like RLE and standard Bayesian uncertainty estimators assume unimodal and isotropic underlying distributions to ensure mathematical tractability during optimization. In chaotic, real-world multi-person scenarios, this assumption collapses. When visual evidence is heavily overlapping, the true probability density is inherently multimodal \cite{rethinking2024crowds}. A unimodal Laplace distribution cannot effectively represent two disjoint, highly probable spatial regions without artificially inflating its variance to encompass both, resulting in an uninformative, over-smoothed uncertainty metric. 

Furthermore, these probabilistic regression techniques require fundamental architectural modifications and end-to-end retraining of the neural network. By formulating the density explicitly as a Gaussian Mixture Model (GMM) at the post-hoc decoding stage, our framework provides the mathematical vocabulary necessary to capture symmetric bimodal confusions without altering the underlying deterministic network weights. Coupled with the orthogonal Uniform component that natively absorbs heavy-tailed Out-of-Distribution background noise, this continuous decoding paradigm resolves the structural limitations that unimodal distributions struggle to characterize, offering a plug-and-play probabilistic upgrade for any standard heatmap-based architecture.

\subsection{Test-Time Augmentation in Vision}
Test-Time Augmentation (TTA) is a standard inferential technique utilized to marginalize over stochastic geometric and photometric perturbations, improving structural robustness without requiring architectural modifications \cite{ayhan2018test}. By evaluating the network across a Cartesian product of transformations, typically horizontal flipping and multi-scale cropping, TTA generates a consensus prediction that is statistically more resilient to isolated inference errors.

In classical top-down HPE paradigms, TTA is executed via a discrete aggregation mechanism. Augmented views are passed through the network, the resulting heatmaps are aligned back to the canonical coordinate system using inverse affine transformations, and the tensors are fused via naive pixel-wise averaging. While effective for minor noise suppression, this discrete averaging is vulnerable to ``heatmap poisoning'' when multi-scale augmentations are introduced. When a bounding box is scaled down to zoom in on the subject, peripheral anatomical joints (e.g., ankles or wrists) may fall outside the physical field of view. Deprived of local visual evidence, the neural network defaults to outputting highly diffuse, entropic distributions. Naively averaging these flat, out-of-bounds heatmaps with accurate predictions from other scales structurally dilutes the legitimate probability mass, degrading overall precision. Our framework resolves this vulnerability by formulating TTA as a continuous, confidence-weighted aggregation pipeline. By evaluating the structural sharpness of each empirical map via an asymptotic sigmoidal penalty prior to fusion, our continuous aggregation dynamically attenuates uninformative scales. This ensures that multi-scale consensus filtering operates strictly as a spatial noise filter rather than a source of entropic contamination, producing a well-conditioned probability density prior to downstream stochastic sampling.

\subsection{Structural Priors \& Graph Models}
Because individual keypoint predictions are executed independently by the convolutional or attention heads, deep networks frequently output anatomically impossible configurations when local visual evidence is compromised. To impose physiological coherence, researchers have extensively explored integrating global structural priors and graph models into the pose estimation pipeline. Classical techniques model the human skeleton as a hierarchical tree graph, employing Markov Random Fields (MRFs) or Conditional Random Fields to execute Belief Propagation \cite{yang2011articulated}. These models define unary potentials based on local network confidence and pairwise potentials based on expected Euclidean bone lengths and joint angle limits, penalizing configurations that violate anatomical norms. More recently, Graph Convolutional Networks (GCNs) \cite{zhao2019semantic} have been deployed to implicitly learn these spatial dependencies directly from data, smoothing out coordinate jitter and inferring the locations of occluded joints by propagating information across the skeletal graph.

While these graph-based regularizers theoretically provide a mechanism for resolving local ambiguity, imposing rigid kinematic priors exclusively in two-dimensional image space exposes a critical vulnerability to projective geometry. Because a 2D camera projection completely collapses the depth axis ($Z$), 2D structural regularizers suffer from projective foreshortening. A physically long anatomical limb segment (such as the thigh or forearm) directed strictly along the optical axis towards the camera lens projects onto the 2D focal plane as a shortened pixel segment. To a 2D MRF relying on Euclidean bone-length priors, this valid foreshortened limb appears as a kinematic violation. Consequently, under extreme visual degradation where local heatmap evidence flattens out, 2D graph models misinterpret projective overlaps, forcefully contorting the skeleton to satisfy erroneous global priors and triggering symmetric limb swaps.

By explicitly isolating bimodal topological states, our GMM probability distributions provide the ideal local hypotheses for an MRF to evaluate. However, by critically analyzing its empirical trade-offs, our work explicitly demonstrates the inherent geometric ceiling of 2D kinematic regularizers. Ultimately, this highlights the necessity of transitioning continuous probabilistic architectures into three-dimensional space. Because our framework parametrically condenses 2D heatmaps into continuous probability distributions, it provides a mathematically elegant foundation for multi-view 3D extensions. By triangulating the continuous 2D spatial rays of our Gaussian components algebraically, future architectures can resolve absolute physical bone lengths in 3D Euclidean space, neutralizing the projective foreshortening ambiguities encountered in 2D formulations and bypassing the $\mathcal{O}(N^{3})$ computational memory bottlenecks of traditional volumetric voxel grids.
\section{Methodology}

\subsection{Theoretical Foundations of Probabilistic Modeling}
This section describes the mathematical and computational framework designed to transform Human Pose Estimation (HPE) from a deterministic coordinate extraction paradigm to continuous, density-aware probabilistic modeling. Unlike conventional methodologies that collapse two-dimensional activation tensors into point estimates, even when employing advanced local gradient approximations like Taylor expansions, this work intercepts the underlying spatial densities to parameterize heteroscedastic uncertainty. This approach aligns with recent research demonstrating that calibrated probability representations outperform rigid Gaussian heatmaps in capturing the true spatial distribution and modeling heteroscedastic uncertainty \cite{li2021human, kendall2017uncertainties}. 

\begin{figure*}[htbp]
    \centering
    \includegraphics[width=\textwidth]{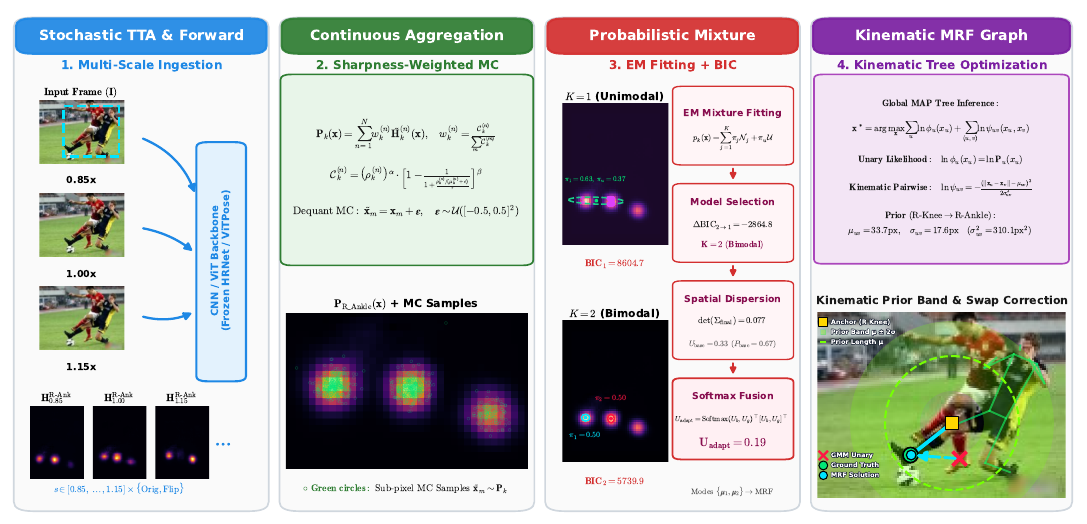}
    \caption{\textbf{Overview of the proposed continuous probabilistic framework for Human Pose Estimation.} 
    \textbf{(1) Stochastic Test-Time Augmentation:} Multi-scale and flipped image crops are passed through a frozen pose backbone. 
    \textbf{(2) Continuous Spatial Aggregation \& Sampling:} Raw heatmaps are aggregated via continuous inverse transforms and confidence weighting, followed by temperature-sharpened Von Neumann rejection sampling ($N=1000$). 
    \textbf{(3) Probabilistic Mixture Modeling:} The EM algorithm concurrently fits candidate models ($K=1$ vs. $K=2$ Gaussians plus Uniform noise), resolved objectively via Bayesian Information Criterion (BIC). 
    \textbf{(4) Kinematic MRF Graph Decoding:} Ambiguous bimodal states are resolved by Maximum a Posteriori (MAP) Belief Propagation over invariant skeletal anchors, producing the final pose with calibrated spatial covariance ellipses ($\boldsymbol{\Sigma}$).}
    \label{fig:methodology_flowchart}
\end{figure*}

The proposed architecture is depicted in Figure \ref{fig:methodology_flowchart}, illustrating the complete four-stage execution pipeline from multi-scale stochastic ingestion to kinematic graph decoding. This modular system can operate in two distinct inference flows to analyze different aspects of the model:
\begin{enumerate}
    \item \textbf{The Solitary Flow (Base Representation):} In its most fundamental form, a single input image is processed through the backbone to obtain raw heatmaps. We then apply sub-pixel Monte Carlo rejection sampling (Stage 2) to extract spatial coordinates, which are parameterized into the GMM+Uniform distribution via EM and BIC selection (Stage 3). This baseline flow (\textit{Image $\rightarrow$ Heatmap $\rightarrow$ Sampling $\rightarrow$ GMM+Uniform}) bypasses TTA and MRF decoding, ensuring geometric parity with optimized deterministic sub-pixel decoders (e.g., DARK), while establishing the foundational parameters for continuous geometric uncertainty analysis.
    \item \textbf{The Global Stochastic Flow (TTA + MRF):} To analyze the model's multimodal behavior under structural ambiguity, the pipeline integrates continuous aggregation and geometric priors across the full pipeline (Figure \ref{fig:methodology_flowchart}, Stages 1--4). Multiple augmented image views (incorporating various geometric transformations such as horizontal flips and scaling) are aggregated via Test-Time Augmentation (TTA) before sampling. The resulting GMM distributions are then passed through a Markov Random Field (MRF, Stage 4) to study the trade-offs of enforcing skeletal kinematics. This comprehensive flow (\textit{Image $\rightarrow$ TTA Heatmaps $\rightarrow$ Sampling $\rightarrow$ GMM+Uniform $\rightarrow$ MRF}) allows us to thoroughly investigate how local uncertainty parameters interact with global anatomical constraints.
\end{enumerate}

To deeply understand the mechanisms by which the proposed framework translates raw heatmaps into spatial coordinates, it is essential to ground the methodology in the foundational principles of probability theory, Bayesian inference, and mixture modeling. The theoretical underpinnings of this research draw extensively from the formulations presented in Christopher M. Bishop's seminal text, \textit{Pattern Recognition and Machine Learning} \cite{bishop2006pattern}.

In the domain of pattern recognition, probability theory provides a consistent framework for the quantification and manipulation of uncertainty. When a deep neural network processes an image to localize an anatomical keypoint, the output is not a definitive location but a spatial distribution over a 2D grid. Let us denote the spatial coordinates as a continuous random variable $\mathbf{x} = (x, y)^\top$. The neural network outputs an unnormalized probability density function \(p(\mathbf{x} \mid \mathcal{D})\), where $\mathcal{D}$ represents the observed image data.

From a frequentist perspective, the heatmap represents the relative frequency of the keypoint appearing at a specific pixel if the identical observation were repeated infinitely. However, a Bayesian perspective is far more appropriate for the complexities of pose estimation. Here, \(p(\mathbf{x} \mid \mathcal{D})\) quantifies the predictive spatial density and captures the spatial aleatoric uncertainty regarding the joint's location conditioned on the observed visual evidence. When a keypoint is occluded, the degree of belief is spread across a wider spatial area, resulting in a distribution with high variance and high entropy.

The standard procedure of extracting the maximum value (the mode) of the distribution, denoted as $\mathbf{x}_{MAP} = \arg\max_{\mathbf{x}} p(\mathbf{x} \mid \mathcal{D})$, and even subsequent deterministic sub-pixel refinements utilizing local Hessian approximations, completely discards the global shape of the distribution. If the distribution is a perfect, isotropic Gaussian, the mode is a sufficient statistic. However, if the distribution is highly skewed, heavy-tailed, or multimodal due to anatomical symmetries, the mode represents an incomplete and potentially highly misleading summary of the model's belief.

To capture the complex topography of the spatial heatmaps, particularly the presence of multiple peaks, we transcend this deterministic bottleneck by modeling the density explicitly as a Gaussian Mixture Model (GMM) augmented by a Uniform distribution to handle robust outliers.

\subsection{Raw Heatmap Interception and Affine Mapping}
The prerequisite for downstream probabilistic analysis is the unadulterated preservation of the probability mass inferred by the network. State-of-the-art approaches such as Unbiased Data Processing (UDP) \cite{huang2020devil} or Distribution-Aware Coordinate Representation of Keypoints (DARK) \cite{zhang2020distribution} resort to Taylor series approximations to achieve sub-pixel localization, but this analytical process destroys the original multimodal spatial landscape of the activation tensor.

To circumvent this decoding bottleneck, the framework implements a dynamic interception at the output layer. Each heatmap $\mathbf{H}_k$ corresponding to joint $k$ is isolated and normalized to satisfy the Kolmogorov axioms, transforming it into a discrete empirical probability mass function (PMF) \(\hat{\mathbf{H}}_k\):
\begin{equation}
\hat{\mathbf{H}}_k(x, y) = \frac{\max(0, \mathbf{H}_k(x, y))}{\sum_{i=1}^W \sum_{j=1}^H \max(0, \mathbf{H}_k(i, j))}
\end{equation}

Mapping the covariance matrices and spatial means from the cropped heatmap domain to the absolute coordinate system of the high-resolution image requires a bidirectional affine transformation free of systematic biases. The forward transformation is governed by the equation $\mathbf{x}_{img} = \mathbf{A} \cdot \mathbf{x}_{hm} + \mathbf{b}$. However, underlying codec implementations generate discrepancies often overlooked in the literature. Specifically, for MSRA-based frameworks, the scale matrix is defined as $\mathbf{A}_{MSRA} = \text{diag}(\mathbf{s}_{input} / \mathbf{s}_{hm})$, constituting an integer proportion. Conversely, modern models utilizing the UDP codec demand a fractional scale $\mathbf{A}_{UDP} = \text{diag}((\mathbf{s}_{input} - 1) / (\mathbf{s}_{hm} - 1))$. Ignoring this subtle mathematical distinction induces asymmetric spatial drift during the back-projection of uncertainty ellipsoids. Our system dynamically incorporates these codec-specific affine calibration parameters to compose the Jacobian matrix, ensuring that the continuous density back-projection preserves spatial fidelity and directional orientation upon coordinate scaling.

\subsection{Robust Spatial Aggregation via Scale-Aware Continuous TTA}
Because our formulation intercepts the heatmap to extract continuous probability densities rather than relying on discrete point estimates, it inherently enables the application of Test-Time Augmentation (TTA). Test-Time Augmentation is conceptually elevated from a mere deterministic booster to an empirical integration mechanism for mitigating spatial noise \cite{ayhan2018test}. Given an ensemble of geometric transformations (spatial scaling and horizontal reflections), the network explores perturbed perspectives of the input scene.

However, the inclusion of multi-scale perturbations presents an algorithmic vulnerability termed \textit{heatmap poisoning}. In a top-down paradigm, scaling the bounding box zooms the crop. When scaling down ($s < 1.0$), peripheral joints frequently exceed the field of view (FOV). Deprived of local visual evidence, the neural network collapses, producing a flat entropic distribution. Naively averaging these uniform maps destroys the legitimate density peaks obtained at other scales.

To eradicate this phenomenon, we propose a continuous aggregation weighted by structural confidence. Instead of applying rigid thresholds that induce discontinuities in the likelihood function, we evaluate the structural sharpness of each empirical map. For the $n$-th variant of keypoint $k$, we extract the supremum $\rho_k^{(n)} = \max(\hat{\mathbf{H}}_k^{(n)})$ and the expected mean $\mu_k^{(n)} = \mathbb{E}[\hat{\mathbf{H}}_k^{(n)}]$. A continuous confidence score $\mathcal{C}_k^{(n)}$ is defined via an asymptotic sigmoidal penalty:
\begin{equation}
\mathcal{C}_k^{(n)} = \left(\rho_k^{(n)}\right)^\alpha \cdot \left(1 - \frac{1}{1 + \frac{\rho_k^{(n)} / (\mu_k^{(n)} + \epsilon)}{\tau}}\right)^\beta,
\end{equation}
Where $\tau$ is the sharpness scale factor, $\epsilon$ is an infinitesimal regularization constant, and $\alpha, \beta \ge 0$ are hyperparametric exponents that modulate the aggressiveness of the attenuation. A diffuse, poisoned heatmap originating from an out-of-frame joint yields $\rho_k^{(n)} \approx \mu_k^{(n)}$, driving the sigmoidal ratio toward zero and nullifying its contribution weight without relying on non-differentiable hard gating.

Prior to aggregation, the raw prediction maps $\hat{\mathbf{H}}_k^{(n)}$ are aligned into a common canonical reference frame. Our framework performs an inverse geometric transformation via bilinear grid sampling, dynamically accounting for codec-aware spatial scaling factors (e.g., fractional disparities between MSRA and UDP coordinate conventions). Inversely warping the maps with an \textit{align\_corners=False} convention ensures that the spatial probability mass preserves its sub-pixel metric alignment, yielding the canonical maps $\tilde{\mathbf{H}}_k^{(n)}$. Finally, the confidence scores across all $N$ augmentation variants are normalized to produce a valid convex combination:
\begin{equation}
w_k^{(n)} = 
\begin{cases} 
\displaystyle \frac{\mathcal{C}_k^{(n)}}{\sum_{m=1}^N \mathcal{C}_k^{(m)}}, & \text{if } \sum_{m=1}^N \mathcal{C}_k^{(m)} > \epsilon_{\text{tol}}, \\[10pt]
\displaystyle \frac{1}{N}, & \text{otherwise},
\end{cases}
\end{equation}
where $\epsilon_{\text{tol}}$ represents a numerical safety threshold. When all scales fail to capture the joint, the aggregation gracefully defaults to uniform averaging. The fused probabilistic consensus $\mathbf{P}_k(\mathbf{x})$ is then synthesized as the convex weighted sum:
\begin{equation}
P_k(\mathbf{x}) = \sum_{n=1}^N w_k^{(n)} \tilde{\mathbf{H}}_k^{(n)}(\mathbf{x}),
\label{eq:confidence_weighting}
\end{equation}
effectively filtering out poisoned predictions while consolidating multi-scale evidence into a unified, continuous probability distribution.

\subsection{Stochastic Dequantization and Monte Carlo Sampling}
\label{subsec:methodology_sampling}
Given that the fused empirical distributions $P_k(\mathbf{x})$ reside on a discrete spatial lattice, directly fitting a continuous statistical model suffers from severe quantization errors. The approximation of continuous two-dimensional densities is achieved through a vectorized Von Neumann Rejection Sampling method. By processing large batches of candidate coordinates concurrently, this strategy draws unbiased samples from the underlying spatial probability mass function (PMF).

To modulate the map's entropy and focus the sampling process on the highest density modes, a thermodynamic parameter $T \in (0, 1]$ is injected. The smoothed density is theoretically formulated as $P_T(\mathbf{x}) \propto P_k(\mathbf{x})^{1/T}$. However, to guarantee numerical stability and prevent overflow during extreme sharpening ($T \ll 1$), this modulation is implemented in log-space:
\begin{equation}
    P_T(\mathbf{x}) = \frac{\exp(\frac{1}{T}\log(P_k(\mathbf{x}) + \epsilon))}{\sum \exp(\frac{1}{T}\log(P_k(\mathbf{x}) + \epsilon))}
\end{equation}
Subsequently, extracting integer candidates $\{\mathbf{x}_m\}_{m=1}^M$ from the grid is inevitably prone to covariance collapse. If multiple samples fall on the same pixel coordinate pair, the empirical dispersion matrix loses full rank and becomes singular, rendering matrix inversion ill-defined and destabilizing the Mahalanobis distance evaluations required during EM clustering.

To mitigate this intrinsic limitation of discretized spaces, a sub-pixel de-quantization step is applied post-sampling. A uniform white noise component \(\boldsymbol{\epsilon} \sim \mathcal{U}([-0.5, 0.5]^2)\) is injected into each retained coordinate. The smoothed random variable $\tilde{\mathbf{x}}_m = \mathbf{x}_m + \boldsymbol{\epsilon}$ recovers the continuous support domain $\mathbb{R}^2$, faithfully modeling the uncertainty inherent to the underlying feature map resolution and immunizing the subsequent maximization step against numerical instabilities.

From a theoretical standpoint, the injection of this uniform noise introduces an additive isotropic bias to the empirical dispersion matrix. Due to the statistical independence of the injected noise, the empirical dispersion matrix undergoes an additive isotropic shift equal to the variance of the continuous uniform perturbation: \(\operatorname{Cov}(\boldsymbol{\epsilon}) = \frac{(0.5 - (-0.5))^2}{12}\mathbf{I} = \frac{1}{12}\mathbf{I} \approx 0.0833\mathbf{I}\).

Crucially, we do not subtract this variance bias post-hoc. Instead, we deliberately retain it as a principled lower-bound regularization penalty that encapsulates the fundamental quantization limit of the 2D discrete lattice (i.e., the spatial discretization uncertainty inherent to coordinate localization within a \(1 \times 1\) pixel cell). Consequently, this structural variance lower-bound seamlessly complements the subsequent Tikhonov regularization, ensuring strict positive definiteness without compromising the geometric fidelity of the uncertainty ellipsoids.

\subsection{Continuous Density Modeling via Robust Gaussian Mixtures}
The core analytical contribution of this framework is the parametric condensation of keypoint densities into a Gaussian Mixture Model (GMM). Unlike conventional regression methods that force unimodality, the mixture representation explicitly captures symmetric bilateral ambiguities (e.g., spatial confusion between contralateral left and right joints), while providing a mechanism to absorb Out-of-Distribution (OoD) noise from erroneous background detections.

The hybrid probabilistic model assumes the sampled spatial data originates from a mixture of \(K\) Gaussian components and a spatially agnostic uniform noise process, formalized as:
\begin{equation}
p(\tilde{\mathbf{x}} \mid \Theta) = \sum_{k=1}^K \pi_k \mathcal{N}(\tilde{\mathbf{x}} \mid \boldsymbol{\mu}_k, \boldsymbol{\Sigma}_k) + \pi_u \mathcal{U}(\tilde{\mathbf{x}} \mid A),
\end{equation}
where \(A\) denotes the spatial area of the input bounding box, ensuring a constant background density. The parameter set \(\Theta = \{\pi_k, \boldsymbol{\mu}_k, \boldsymbol{\Sigma}_k\}_{k=1}^K \cup \{\pi_u\}\), subject to \(\sum_{k=1}^K \pi_k + \pi_u = 1\), is optimized via the Expectation-Maximization (EM) algorithm.

\begin{figure*}[htbp]
    \centering
    \includegraphics[width=\textwidth]{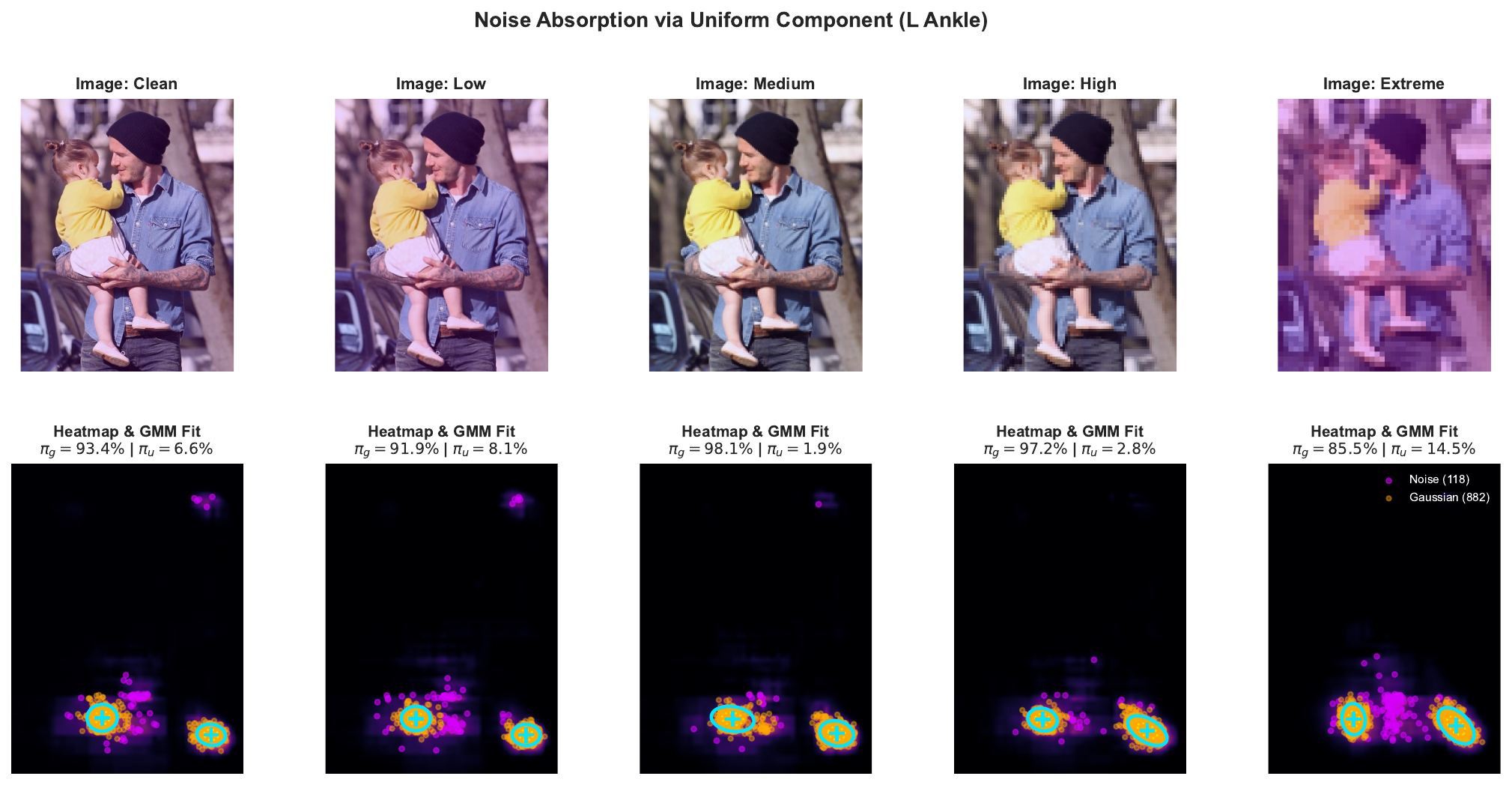}
    \caption{\textbf{Empirical motivation and mechanics of the orthogonal Uniform component ($\pi_{\text{uniform}}$).} 
    \textbf{Top Row:} Input image instances under escalating resolution loss ($\text{Clean } 1.00\times \rightarrow \text{Extreme } 0.062\times$). 
    \textbf{Bottom Row:} Corresponding activation heatmaps, Monte Carlo candidate samples, and fitted continuous components for the left ankle. In structured visual regimes ($\le \text{High}$), spatial evidence concentrates cleanly around genuine anatomical modes, assigning $\ge 91.9\%$ of the probability mass to the Gaussian components ($\pi_g$) and leaving uniform absorption minimal ($\pi_u \le 8.1\%$). Under extreme resolution destruction ($\text{Extreme } 0.062\times$), unstructured background noise contaminates the feature map. Rather than forcing a Gaussian to expand and corrupt the true coordinate expectation, the EM algorithm naturally diverts diffuse, non-Gaussian outliers into the uniform component ($\pi_u = 14.51\%$, magenta dots), preserving the compact geometric integrity of the genuine anatomical modes (cyan/orange ellipses).}
    \label{fig:methodology_uniform_motivation}
\end{figure*}

The inclusion of the orthogonal uniform component $\mathcal{U}(\tilde{\mathbf{x}}|A)$ is mathematically essential to guarantee the geometric stability of the Gaussian parameters. In unconstrained visual environments or under severe degradation, activation heatmaps often exhibit diffuse background noise alongside true keypoint peaks. If a standard, pure Gaussian mixture were fitted, the Expectation-Maximization (EM) algorithm would attempt to cover these distant, scattered outliers by artificially inflating the covariance matrices $\boldsymbol{\Sigma}_k$ or allocating spurious Gaussian components. 

As visually demonstrated in Figure \ref{fig:methodology_uniform_motivation}, our hybrid formulation resolves this structural vulnerability by acting as a probability sink. While anatomical activations are tightly clustered by the Gaussian components (orange dots), diffuse background samples (magenta dots) are naturally assigned to the uniform component $\pi_u$. Under extreme degradation (Figure \ref{fig:methodology_uniform_motivation}, Extreme $0.062\times$), $\pi_{\text{uniform}}$ dynamically increases to absorb $14.51\%$ of the spatial noise mass, allowing the primary Gaussian ellipses to maintain their true physical orientation and sub-pixel localization accuracy without covariance explosion.

\subsubsection{Initialization and Tikhonov Regularization}
Given the sensitivity of the Expectation-Maximization (EM) algorithm to initial conditions, the Gaussian means are seeded using the \textit{k-means++} algorithm \cite{arthur2007k} to ensure dispersed and stable initial cluster centers, falling back to uniform random sampling only if the spatial distribution is severely degenerate. Furthermore, components whose mixing weight drops below a numerical threshold ($\pi_k < 10^{-3}$) are actively monitored to prevent dead components.

During the M-step (Maximization), the estimation of covariance matrices $\Sigma_k$ is susceptible to ill-conditioning derived from components capturing minuscule groups of collinear outliers. Consequently, we apply dynamic Tikhonov regularization driven by spectral decomposition. If the minimum eigenvalue $\lambda_{min}(\Sigma_k)$ of any component falls below an operational threshold $\lambda_{reg}$, a stabilizing term is orthogonally injected:
\begin{equation}
    \tilde{\Sigma}_k = \Sigma_k + (\lambda_{reg} - \min(0, \lambda_{min})) I
\end{equation}
This algebraic step preserves the strict positive definiteness of the matrices, allowing the valid tracing of anatomical uncertainty ellipsoids.

\subsubsection{Model Selection via Bayesian Information Criterion}
To automatically adjudicate the optimal cardinality of the Gaussian set ($K=1$ unimodal vs. $K=2$ bimodal), we employ a complexity penalty through the Bayesian Information Criterion (BIC). This criterion balances the log-likelihood growth ($\mathcal{L}$) against parametric parsimony ($p$ free parameters for $M$ Monte Carlo samples):
\begin{equation}
    BIC = p \ln(M) - 2\mathcal{L}
\end{equation}
By relying on the BIC, the formulation acts as a conservative structural gatekeeper. It heavily penalizes unnecessary complexity, ensuring that the model transitions to a bimodal mixture ($K=2$) only when the spatial evidence of structural ambiguity (e.g., a potential limb swap) is mathematically overwhelming, thereby preventing the noise-induced overfitting of spurious components.

\subsection{Global Constraint Inference with Markov Random Fields (MRF)}
Thanks to the density modeling provided by the GMM (which supplies both multimodal hypothesis centers and their spatial variances), our framework provides a principled interface for probabilistic graphical models to resolve complex spatial ambiguities. While the GMM locally isolates bimodal hypotheses, it lacks the global spatial context required to resolve anatomical inversions. To impose inter-articular kinematic coherence, the architecture maps the human body as a hierarchically structured tree graph $\mathcal{G} = (\mathcal{V}, \mathcal{E})$, rooted at the facial axis (nose), where the edges $\mathcal{E}$ dictate the adjacencies of the COCO skeleton \cite{lin2014microsoft}.

Crucially, our formulation selectively targets ambiguity: confident unimodal predictions ($K=1$) bypass variable optimization to act as fixed structural anchors, while the MRF exclusively resolves nodes exhibiting bimodal uncertainty ($K=2$). We seek to infer the Maximum a Posteriori (MAP) configuration for these ambiguous joint states $\mathbf{Y}$, where the state space of node $i$ is dictated by its local Gaussian centers $\mathbf{y}_i \in \{\boldsymbol{\mu}_1^{(i)}, \boldsymbol{\mu}_2^{(i)}\}$. For numerical stability, the inference is performed in log-space, maximizing the sum of unary and pairwise potentials:
\begin{equation}
\arg\max_{\mathbf{Y}} \left( \sum_{i \in \mathcal{V}} \log \phi_i(\mathbf{y}_i) + \sum_{(i, j) \in \mathcal{E}} \log \psi_{ij}(\mathbf{y}_i, \mathbf{y}_j) \right)
\label{eq:mrf_map}
\end{equation}

The unary potential \(\phi_i\) evaluates the viability of candidate \(\mathbf{y}_i = \boldsymbol{\mu}_k^{(i)}\) in log-space by combining its GMM prior \(\pi_k^{(i)}\) with its modal peak density, formalized as \(\log \phi_i(\mathbf{y}_i) = \log \pi_k^{(i)} - \frac{1}{2} \log \det(\boldsymbol{\Sigma}_k^{(i)}) - \log(2\pi)\), penalizing hypotheses with high spatial dispersion.

The binary potential $\psi_{ij}$ imposes physiological constraints on bone lengths, acting as a parametric Gaussian prior over the spatial configuration. To ensure anatomical accuracy, the expected bone length ratios ($\mu_{ij}^{bone}$) and their standard deviations ($\sigma_{ij}^{bone}$) were empirically extracted from visible instances in the COCO \texttt{train2017} dataset. The affinity of a connection is evaluated by projecting the Euclidean distance between candidate hypotheses against this prior, formulated as:

\begin{equation}
\psi_{ij}(y_i, y_j) = \exp \left( - \frac{ \left( \Vert y_i - y_j \Vert_2 - (\mu_{ij}^{bone} \cdot \sqrt{A_{box}}) \right)^2 }{ 2 \left( \sigma_{ij}^{bone} \cdot \sigma_{mult} \cdot \sqrt{A_{box}} \right)^2 } \right)
\label{eq:mrf_binary}
\end{equation}

where $A_{box}$ denotes the bounding box area of the detected person, dynamically de-normalizing the ratios into absolute pixel distances. To account for natural postural foreshortening and 2D projection artifacts, the standard deviation is relaxed by a multiplicative tolerance factor ($\sigma_{mult}=2.0$). Operating in log-space (Equation \ref{eq:mrf_map}), this pairwise potential reduces to a scaled squared-error penalty, perfectly integrating with the unary uncertainties without risking numerical underflow.

\subsection{Uncertainty Metrics Formulation}
The true power of our formulation emerges when evaluating Out-of-Distribution (OoD) data and catastrophic failures. To translate our mathematical parameters into standardized uncertainty metrics, we explicitly decouple multimodal spatial ambiguity from catastrophic background noise.

First, to quantify anatomical and structural uncertainty (e.g., contralateral limb swaps or spatial blur), we extract a continuous geometric metric directly from the Gaussian components. Because the variance of a uniform distribution spanning the entire bounding box is orders of magnitude larger than localized joint activations, incorporating it into the total dispersion calculation would completely saturate the covariance ellipse, masking the subtle geometric ambiguities captured by the Gaussians. Thus, we isolate this spatial metric from the uniform component.

Because the GMM is initially fitted in the cropped heatmap domain, the component means ($\boldsymbol{\mu}_k$) and covariances ($\Sigma_k$) must first be projected into the absolute image coordinate system using the inverse affine transformation of the bounding box. Operating in this absolute image space guarantees dimensional consistency. Let $\tilde{\pi}_k = \pi_k / \sum_{j=1}^K \pi_j$ be the normalized mixing weights of the $K$ Gaussians. The global spatial expectation of the mixture is $\boldsymbol{\mu}_{global} = \sum_{k=1}^K \tilde{\pi}_k \boldsymbol{\mu}_k$. Applying the Law of Total Variance, the total spatial covariance matrix of the Gaussian components is formulated as:
\begin{equation}
    \Sigma_{total} = \sum_{k=1}^K \tilde{\pi}_k \left( \Sigma_k + (\boldsymbol{\mu}_k - \boldsymbol{\mu}_{global})(\boldsymbol{\mu}_k - \boldsymbol{\mu}_{global})^T \right)
    \label{eq:total_variance}
\end{equation}
This formulation provides an analytical advantage: it combines the intrinsic intra-component variance ($\Sigma_k$) with the inter-component geometric displacement. When the model exhibits bilateral ambiguity (e.g., competing hypotheses between left and right wrists), the spatial distance between candidate means acts as a severe penalty, quadratically expanding the covariance volume specifically along the axis of confusion.

To ensure this geometric measure is anatomically meaningful across different joints, we kinematically calibrate the global covariance using the joint-specific COCO tolerance constants $\kappa_j$ \cite{lin2014microsoft} and the squared object scale $s^2$ (derived from the person's total bounding box area in image space):
\begin{equation}
    \Sigma_{final} = \frac{1}{(s \cdot \kappa_j)^2} \Sigma_{total} + \epsilon I
    \label{eq:sigma_final}
\end{equation}
where $\epsilon=10^{-6}$ is added to the diagonal to guarantee strict positive definiteness. The scalar spatial uncertainty is defined by the generalized variance of this calibrated distribution, formally evaluated via its determinant: \(U_{\text{spatial}} = \det(\boldsymbol{\Sigma}_{\text{final}})\).

Second, the mixing weight of the uniform distribution, \(\pi_u\) (or \(\pi_{\text{uniform}}\)), serves as an orthogonal indicator of visual absence. When local visual evidence is destroyed by occlusion or sensor degradation, the EM algorithm diverts the unstructured probability mass into this uniform background sink. Rather than quantifying local spatial dispersion, \(\pi_u\) provides a dedicated indicator of keypoint non-visibility, effectively isolating Out-of-Distribution (OoD) noise and preventing spurious background activations from artificially inflating the geometric covariance volume.

\subsection{Adaptive Uncertainty Fusion}
To capitalize on the volumetric alerts of the GMM without degrading the baseline's robustness in canonical scenarios, we formulate standardized uncertainty metrics. Because the calibrated geometric volume \(\det(\boldsymbol{\Sigma}_{\text{final}})\) is not bounded within \([0, 1]\) and lacks a direct interpretation as a bounded error probability, we project it into the interval \([0, 1]\) using a parametric exponential transfer function modulated by a sensitivity hyperparameter \(\beta\):
\begin{equation}
    U_{\text{gmm}} = 1 - \exp\left(-\beta \cdot \det(\boldsymbol{\Sigma}_{\text{final}})\right).
    \label{eq:projection}
\end{equation}

To seamlessly combine the strengths of the baseline confidence and the structural awareness of the GMM ($U_{gmm}$), we establish a common metric space. Let $P_{DARK} \in [0, 1]$ denote the maximum confidence score extracted by the deterministic sub-pixel decoder (i.e., the peak activation intensity of the refined heatmap). The baseline heuristic uncertainty is naturally defined as the complement of this confidence, $U_{base} = 1 - P_{DARK}$. We evaluate three adaptive fusion strategies to integrate these signals:

\begin{enumerate}
    \item \textbf{Max-Pooling (Pessimistic Approach):} Assumes the worst-case scenario by taking the maximum uncertainty. It acts as a stable upper bound that triggers when volumetric anomalies exist:
    \begin{equation}
        U_{adapt} = \max(U_{base}, U_{gmm})
    \end{equation}
    
    \item \textbf{Multimodal Gating (Discrete and Hybrid):} Routes uncertainty dynamically based on the mixture cardinality selected by the BIC. The discrete variant selects \(U_{\text{gmm}}\) exclusively when bimodal ambiguity is detected (\(K=2\)), falling back to \(U_{\text{base}}\) under unimodality. To account for highly dispersed unimodal predictions, a hybrid variant incorporates a safety threshold \(\tau\), triggering the volumetric uncertainty if \((K=2) \lor (U_{\text{gmm}} > \tau)\).
    
    \item \textbf{Adaptive Softmax Fusion:} A continuous, weighted aggregation that organically balances the contributions based on their relative exponential magnitudes. This approach inherently resolves the rigid hyperparameter sensitivity common in discrete gating or max-pooling:
    \begin{equation}
        U_{adapt} = \frac{\exp(U_{base}) \cdot U_{base} + \exp(U_{gmm}) \cdot U_{gmm}}{\exp(U_{base}) + \exp(U_{gmm})}
    \end{equation}
\end{enumerate}

To ensure optimal calibration across differing environmental complexities, the sensitivity hyperparameter $\beta$ (and the threshold $\tau$ for the hybrid gating) is determined via a logarithmic grid search ($\beta \in [10^{-6}, 10^{4}]$). The optimization objective is defined mathematically as the minimization of the global Area Under the Sparsification Error (AUSE) curve.

\section{Experiments and Results}

\subsection{Experimental Setup}

\subsubsection{Models and Datasets}
To ensure the generalizability of our findings, we evaluate our probabilistic formulation across distinct pose estimation architectures. The primary evaluation presented in this manuscript utilizes \textbf{HRNet-W32} \cite{sun2019deep} (High-Resolution Network). To guarantee our baseline represents the state-of-the-art in deterministic coordinate extraction, this model is configured to utilize the \textbf{DARK} (Distribution-Aware Coordinate Representation) \cite{zhang2020distribution} unbiased sub-pixel decoder natively within the MMPose framework. 

Furthermore, to explicitly confirm that our continuous parametric aggregation scales across different architectural paradigms, we also validate our framework on \textbf{ResNet50} \cite{he2016deep} (a standard convolutional baseline also using DARK) and \textbf{ViTPose-Small} \cite{xu2022vitpose} (a vision transformer configured with the \textbf{UDP} \cite{huang2020devil} codec). The comprehensive results for these additional architectures mirror the behavioral improvements observed in HRNet and are detailed in the Supplementary Material.

To ensure a fair comparison across these architectures, we designed a framework-agnostic \textit{Adapter Layer} following SOLID principles (specifically Dependency Inversion). Rather than coupling our probabilistic pipeline to a specific library, all models are integrated via an \texttt{MMPoseAdapter} that standardizes their raw outputs into a unified \texttt{StandardizedHeatmap} format. All models operate under a top-down paradigm with a fixed input resolution of $256 \times 192$, guaranteeing that any observed differences in stochastic behavior stem entirely from the network's intrinsic architecture rather than external framework variations or resolution disparities.

We benchmark the models on three datasets of varying complexity:
\begin{itemize}
    \item \textbf{COCO val2017} \cite{lin2014microsoft}: The standard benchmark for general human pose estimation, primarily featuring well-framed and isolated individuals.
    \item \textbf{CrowdPose} \cite{li2019crowdpose}: Specifically selected for its high density of overlapping individuals, essential for evaluating robustness under crowding and complex multi-person interactions.
    \item \textbf{OCHuman} \cite{zhang2019pose2seg}: Characterized by severe, heavy occlusions. This dataset is critical for stressing the model's structural integrity and its vulnerability to predicting invisible or obstructed joints.
\end{itemize}

\begin{figure}[htbp]
    \centering
    \includegraphics[width=\columnwidth]{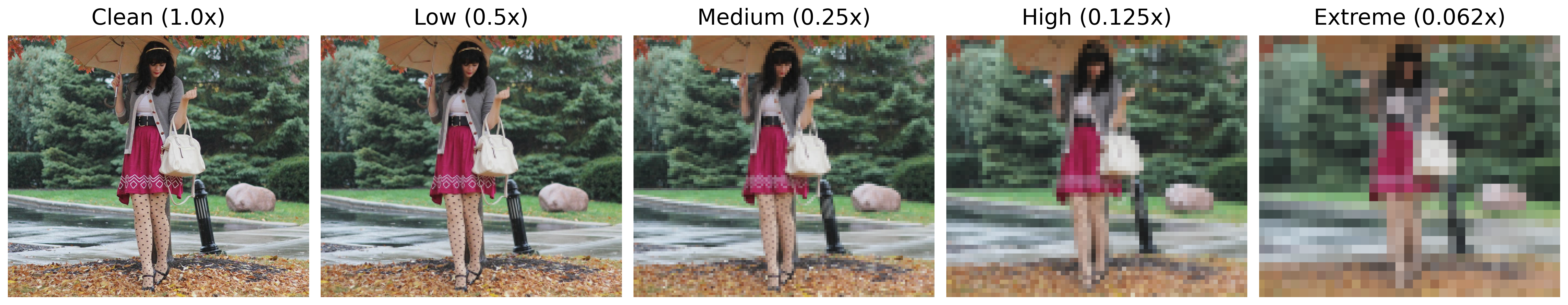}
    \caption{Example of the resolution degradation applied to simulate real-world low-quality inputs. Rather than injecting artificial Gaussian noise or uniform blur, we apply deterministic downsampling to organically reduce spatial information.}
    \label{fig:degradation_example}
\end{figure}

To simulate the challenges of real-world deployment on low-quality camera feeds, we introduce a \textit{Degradation Benchmark}. As illustrated in Figure \ref{fig:degradation_example}, we aggressively downsample the input images (resolution loss) rather than applying synthetic noise. This forces the network to infer structural geometry from constrained pixel evidence. To ensure statistical robustness and comparative consistency, our evaluation protocol samples exactly 500 images per dataset. Consequently, the benchmark is executed ensuring 500 independent inferences for every specific degradation level across every dataset (e.g., 500 for \textit{Clean} COCO, 500 for \textit{Clean} OCHuman, \dots, 500 for \textit{Low} COCO, 500 for \textit{Low} OCHuman, etc.). This ensures that the observed metric trajectories track purely architectural resilience rather than variations in sample size.

\subsubsection{Probabilistic Modeling Configuration}
To translate discrete heatmaps into continuous probability distributions, we perform Monte Carlo Rejection Sampling ($N=1000$ samples per keypoint). To ensure empirical reproducibility across all stochastic evaluations, a global pseudo-random seed (\texttt{seed=42}) is enforced throughout the sampling pipeline. Prior to sampling, a temperature scaling of $T=0.3$ is applied to the heatmaps to concentrate samples around the highest confidence peaks, avoiding excessive diffusion. Furthermore, we apply sub-pixel uniform dequantization (adding a continuous jitter \(\boldsymbol{\epsilon} \sim \mathcal{U}([-0.5, 0.5]^2)\) to the discrete coordinate grids). This is mathematically necessary to prevent the collapse of the EM algorithm into singular covariance matrices (as justified in Section \ref{subsec:methodology_sampling}).

The sampled points are clustered via the Expectation Maximization (EM) algorithm to fit a Gaussian Mixture Model (GMM). We cap the process at 100 iterations or a convergence threshold of $10^{-3}$ in log-likelihood. To further secure numerical stability, Tikhonov regularization (\(\lambda_{\text{reg}} = 10^{-4}\)) is added to the covariance diagonals. Model selection between unimodal and bimodal states is governed by the Bayesian Information Criterion (BIC), penalizing overfitting.

\subsubsection{Test-Time Augmentation (TTA) Configuration}
Our TTA pipeline generates a cartesian product of geometric augmentations, comprising horizontal flips and multi-scale crops ($0.85\times, 0.925\times, 1.0\times, 1.075\times, 1.15\times$). To maintain architectural control, we explicitly disable the native \textit{flip\_test} TTA enabled by default in the MMPose framework (\texttt{test\_cfg=\{\}}). Because our external pipeline explicitly generates both flipped and scaled image views, allowing the framework to perform internal flips would result in redundant double-augmentations and bypass our custom aggregation logic. 

Specifically, we eschew naive pixel-wise averaging for the entire suite. As detailed in the Methodology, while non-scale augmentations (flips) are aggregated standardly, scale augmentations are processed using our scale-aware confidence-weighted aggregation to actively prevent \textit{heatmap poisoning} (i.e., avoiding contamination when a keypoint falls outside the field of view during a zoomed-in crop). To  penalize these out-of-bounds flat distributions, we configure the empirical parameters of our confidence weighting formulation with a peak exponent $\alpha = 1.5$, a sharpness exponent $\beta = 1.5$, and a sharpness scale divisor \(\tau = 10.0\). Only after this dynamically weighted heatmap aggregation do we extract Monte Carlo samples to fit the continuous GMM or decode the standard baseline coordinate.

\subsubsection{Evaluation Metrics}
To comprehensively evaluate both deterministic precision and probabilistic reliability, we establish a multi-tiered evaluation protocol:

\textbf{Geometric Precision and Anatomical Coherence:}
Standard coordinate accuracy is measured using the Object Keypoint Similarity (OKS) \cite{lin2014microsoft}. To evaluate structural integrity, we explicitly track \textbf{Limb Swaps} (instances where the model confuses contralateral joints, such as predicting a left wrist at the right wrist's coordinates). To isolate true anatomical confusions from random background misdetections, our metric enforces visibility constraints and requires the contralateral match to exceed a minimum threshold (OKS $> 0.3$). It tracks errors at two levels of severity:
\begin{enumerate}
    \item \textit{Loose Swaps:} Any instance where the predicted keypoint has a higher OKS match with the contralateral (opposite) ground truth joint than the ipsilateral (correct) one.
    \item \textit{Catastrophic Swaps (Strict):} Extreme topological failures where the contralateral OKS exceeds the ipsilateral OKS by a massive margin ($> 0.7$). This ensures we only flag high-confidence, definitive anatomical inversions, completely filtering out minor sub-pixel coordinate noise.
\end{enumerate}

\textbf{Uncertainty Calibration (In-Distribution):} To assess how accurately our uncertainty metrics rank spatial errors, we compute the \textbf{Area Under the Sparsification Error (AUSE)} curve \cite{calibration2023hpe}. Lower AUSE values denote superior calibration, indicating that the model's highest uncertainty predictions perfectly correlate with its largest spatial errors. Furthermore, to evaluate continuous reliability, we compute the \textbf{Expected Calibration Error (ECE)}, correlating empirical error (\(1 - \text{OKS}\)) against the geometric covariance volume across distinct mixture cardinality regimes (\(K=1\) vs. \(K=2\)).

\textbf{Out-of-Distribution (OoD) Anomaly Detection:} To evaluate the model's robustness against severe degradation and missing joints, we treat OoD detection as a binary classification task (visible vs. catastrophically occluded or absent joints). We report the \textbf{Area Under the Receiver Operating Characteristic (AUROC)} curve, where a higher AUROC demonstrates that the uncertainty metric successfully isolates catastrophic structural failures.

\subsection{Ablation Study: Continuous Estimation and Stochastic Aggregation}

\subsubsection{Ablation on Continuous Point Estimation (The Solitary GMM)}
The first ablation of our framework isolates the impact of replacing the deterministic point-estimate with the expected value (the mean of the primary Gaussian component). Our baseline utilizes the state-of-the-art in deterministic sub-pixel decoding: the \textbf{DARK} (Distribution-Aware Coordinate Representation) codec, which applies a Taylor expansion over the heatmap's maximal activation to derive sub-pixel coordinates via the local Hessian.

To evaluate this, we compare the Object Keypoint Similarity (OKS) of this optimized deterministic baseline against our solitary GMM formulation on clean, uncorrupted images (the \textit{00\_Baseline\_Clean} condition) across all three datasets.

\begin{table}[htbp]
    \centering
    \caption{Baseline Precision (Clean Images): Comparing the Taylor-expanded DARK decoder against our continuous GMM expectation.}
    \label{tab:baseline_precision}
    \resizebox{\columnwidth}{!}{
    \begin{tabular}{lccc}
        \toprule
        \textbf{Dataset} & \textbf{DARK (Argmax)} & \textbf{Ours (GMM)} & \textbf{$\Delta$ OKS} \\
        \midrule
        COCO       & \textbf{0.7079} & 0.7077 & -0.0002 \\
        CrowdPose  & \textbf{0.8278} & 0.8270 & -0.0008 \\
        OCHuman    & \textbf{0.6281} & 0.6280 & -0.0001 \\
        \bottomrule
    \end{tabular}
    }
\end{table}

As detailed in Table \ref{tab:baseline_precision}, the probabilistic parametrization maintains strict parity with the DARK decoder. The observed differences are negligible (ranging from -0.0001 to -0.0008 OKS). This provides a validation of our stochastic methodology: while DARK simulates continuity by shifting a discrete pixel using local Hessian approximations, our method explicitly recovers the continuous support domain \(\mathbb{R}^2\) through uniform dequantization and evaluates the true spatial expectation (\(\boldsymbol{\mu}_1\)).

\begin{figure*}[htbp]
    \centering
    \includegraphics[width=\textwidth]{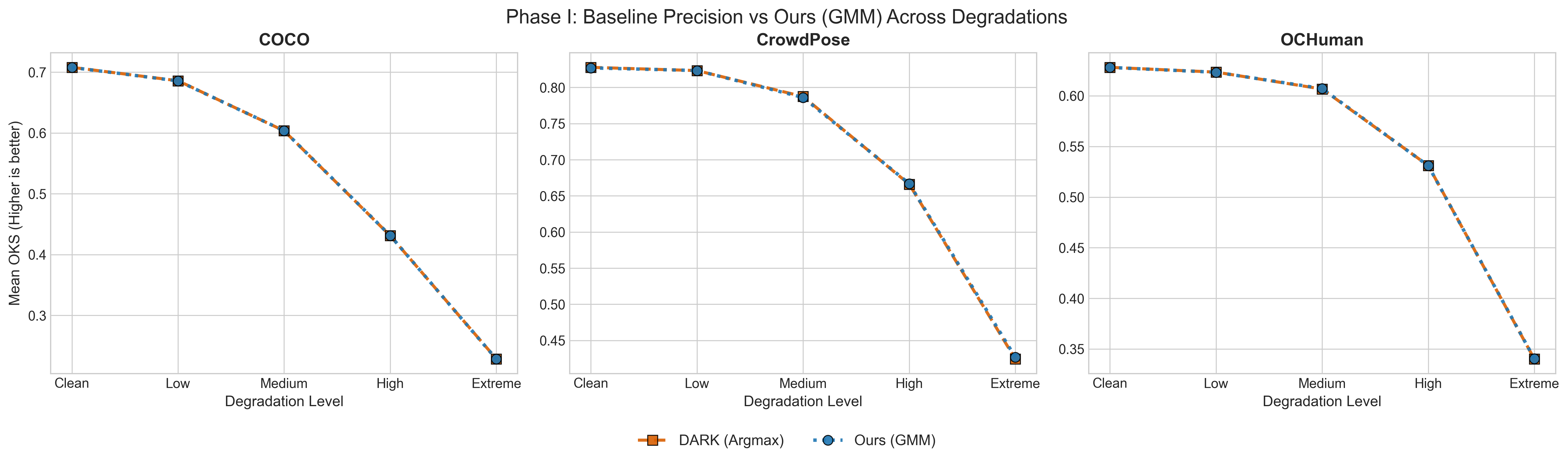}
    \caption{Ablation on Continuous Point Estimation: Baseline Precision across visual degradations. The continuous GMM expectation (blue) maintains geometric parity with the DARK sub-pixel decoder (orange) across all levels of corruption.}
    \label{fig:baseline_degradation_precision}
\end{figure*}

Furthermore, this geometric resilience extends beyond clean images. Figure \ref{fig:baseline_degradation_precision} traces the evolution of the OKS as visual degradation intensifies. Across all levels of downsampling and resolution loss, the continuous GMM Mean perfectly mirrors the trajectory of the Taylor-expanded DARK estimates. Consequently, our formulation successfully unlocks a continuous probabilistic parameter space without sacrificing the raw geometric precision of state-of-the-art deterministic decoders.

\begin{figure*}[htbp]
    \centering
    \includegraphics[width=\textwidth]{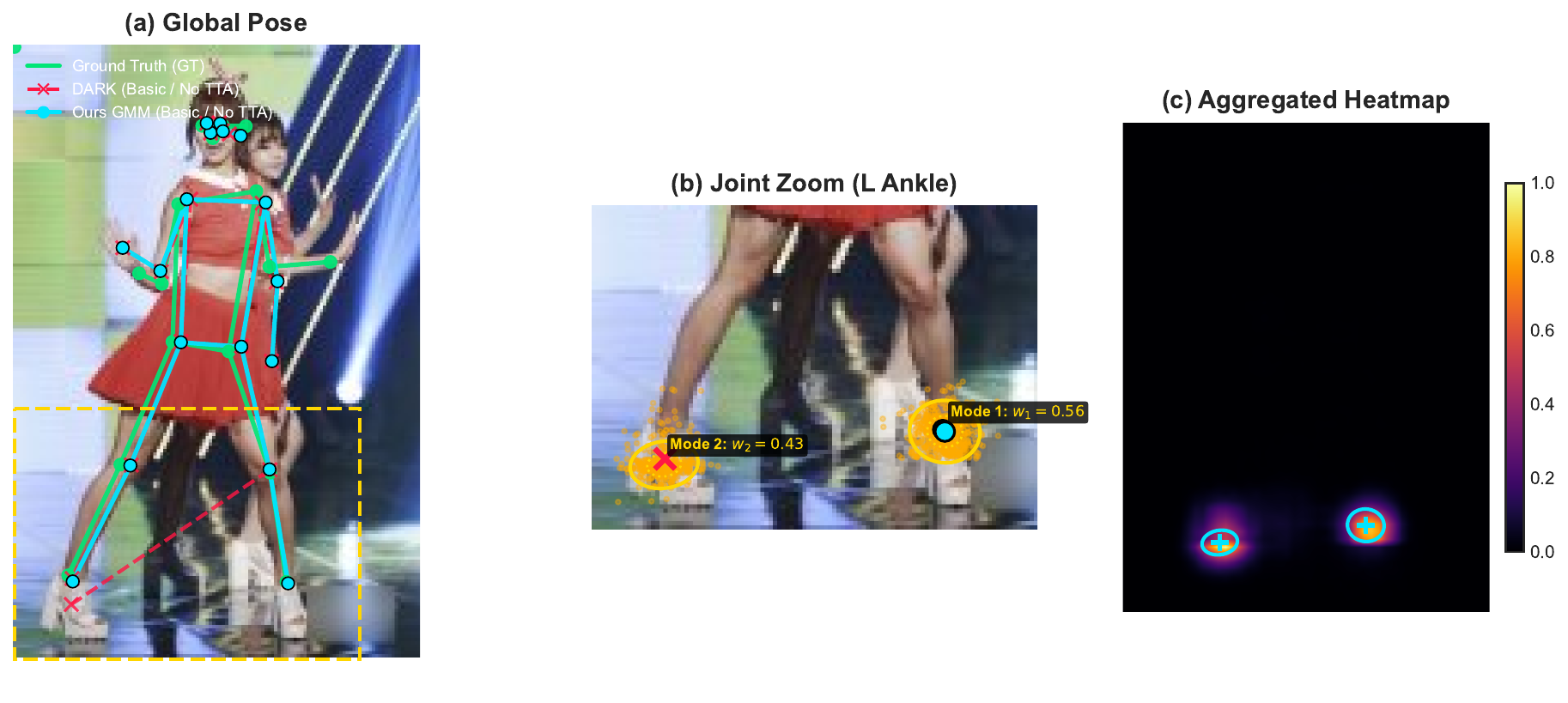}
    \caption{\textbf{Motivation for bimodal modeling ($K=2$).} 
    \textbf{(a)} Global pose estimation on an occluded instance from OCHuman. The deterministic baseline (\textbf{DARK}, dashed red line) suffers from an ankle swap due to coordinate collapse, whereas our continuous formulation (\textbf{Ours}, solid cyan line) preserves the correct anatomical alignment against the Ground Truth (solid green line). 
    \textbf{(b)} High-resolution zoom on the lower limb region. The Monte Carlo samples (amber dots) delineate a dual activation mass, where while DARK prematurely snaps to the contralateral false peak ($\text{Mode } 2: w_2 = 0.43$, red cross), our GMM fits two Gaussian components ($K=2$), retaining both spatial hypotheses and placing primary confidence on the correct mode ($\text{Mode } 1: w_1 = 0.56$, solid gold $2\sigma$ ellipse). 
    \textbf{(c)} Heatmap demonstrating the dual-peak topography intercepted prior to deterministic thresholding.}
    \label{fig:qualitative_swap_resolution}
\end{figure*}

Beyond raw coordinate precision, the continuous GMM provides explicit multimodal awareness during solitary, single-pass inference without requiring external augmentations. As illustrated in Figure \ref{fig:qualitative_swap_resolution}, when spatial heatmaps exhibit symmetric visual ambiguities due to overlapping lower limbs (Figure \ref{fig:qualitative_swap_resolution}c), standard deterministic decoders like DARK are forced into a premature, irreversible coordinate collapse. Because DARK relies exclusively on a second-order Taylor refinement around the global discrete maximum, it rigidly commits to whichever peak yields a marginally higher activation, readily confusing contralateral joints and triggering a catastrophic limb swap (Figure \ref{fig:qualitative_swap_resolution}a).

In contrast, our formulation intercepts the raw activation before this point-collapse and fits a bimodal Gaussian mixture ($K=2$, Panel b). Rather than discarding the secondary hypothesis, the GMM explicitly parameterizes both candidate regions along with their spatial extents ($\text{Mode } 1: w_1 = 0.56$, $\text{Mode } 2: w_2 = 0.43$). This continuous parameterization prevents artificial overconfidence, disambiguate symmetric limbs and retain structural uncertainty.

\subsubsection{Ablation on Stochastic Aggregation (Test-Time Augmentation)}
While the baseline performance establishes the deterministic capability of the network, Test-Time Augmentation (TTA) exploits the stochastic potential of the model by providing multiple geometric perspectives.

\begin{figure*}[htbp]
    \centering
    \includegraphics[width=\textwidth]{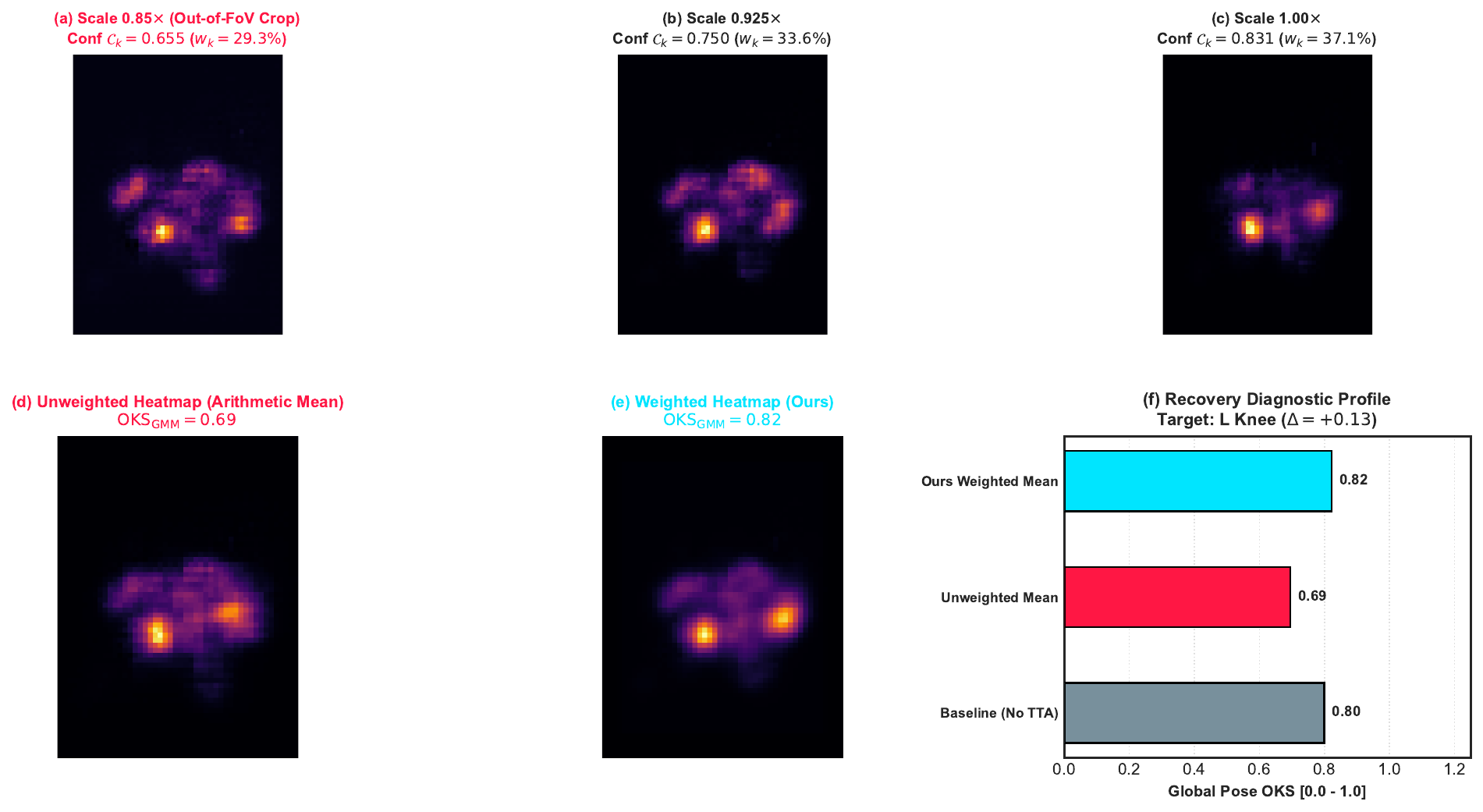}
    \caption{\textbf{Anatomy and mitigation of heatmap poisoning during multi-scale Test-Time Augmentation (TTA).} 
    \textbf{(a)} Representative zoomed-in crop ($0.85\times$) where the target left knee falls outside the physical field of view, causing the network to emit a diffuse, poisoned heatmap ($\mathcal{C}_k = 0.655$, $w_k = 29.3\%$). 
    \textbf{(b--c)} In-bounds scales ($0.925\times$ and $1.00\times$) capturing genuine visual evidence with sharp activation modes ($\mathcal{C}_k \ge 0.750$). 
    \textbf{(d)} Traditional unweighted arithmetic averaging across all TTA scales: the flat, out-of-bounds map dilutes the genuine peak, spreading background noise and degrading localization ($\mathrm{OKS}_{\mathrm{GMM}} = 0.69$). 
    \textbf{(e)} Our quality-gated continuous aggregation: sharpness and peak confidence dynamically penalize the corrupted scale, preserving a sharp, well-defined mode ($\mathrm{OKS}_{\mathrm{GMM}} = 0.82$). 
    \textbf{(f)} Global pose recovery profile across all 17 keypoints: unweighted averaging degrades overall pose accuracy below the baseline ($0.80 \rightarrow 0.69$), whereas our weighted aggregation achieves $+0.13$ OKS recovery, successfully restoring the poisoned target joint from near-zero accuracy to $\mathrm{OKS}_k = 0.99$.}
    \label{fig:qualitative_heatmap_poisoning}
\end{figure*}

Moreover, standard discrete TTA is susceptible to \textit{heatmap poisoning}. As illustrated in Figure \ref{fig:qualitative_heatmap_poisoning}, when multi-scale scaling forces a peripheral joint outside the crop's field of view (Panel a), standard pixel-wise averaging dilutes the genuine activation peak with flat noise (Panel d, dropping OKS to $0.69$). In contrast, our scale-aware confidence weighting (Equation~\eqref{eq:confidence_weighting}) naturally attenuates the uninformative scale, preserving a sharp, well-conditioned probability distribution (Panel e, recovering OKS to $0.82$).

To garantee a fair comparison, our \textbf{DARK (Discrete TTA)} baseline does not naively average final predicted coordinates. Instead, we compute the standard baseline by first averaging the aligned heatmaps utilizing our scale-aware logic to prevent out-of-view heatmap poisoning, and subsequently extracting the keypoint utilizing the DARK decoder. We contrast this state-of-the-art discrete aggregation against our probabilistic framework, which aggregates the distributions directly in the continuous parameter space of the GMMs.

\textbf{For an exhaustive quantitative breakdown across all datasets and degradation levels, we refer the reader to \suppsec{sec:supp_ablations} (\supptab{tab:supp_tta_comprehensive}).} The empirical results reveal an architectural trade-off:

\begin{itemize}
    \item \textbf{Raw Precision vs. Continuous Density Filtering:} In terms of pure geometric precision, the \textbf{DARK (Discrete TTA)} maintains a marginal advantage over the GMM across most canonical scenarios (yielding negligible OKS differences ranging from $0.0002$ to $0.0020$). However, by fusing probability distributions rather than discrete spatial activations, the GMM acts as an effective spatial filter. It systematically reduces minor bilateral ambiguities across almost all evaluated conditions, successfully rescuing 16 loose swaps in COCO High and 18 in OCHuman Extreme.
...
    \item \textbf{Catastrophic Inversions (Strict Swaps):} In relatively clean datasets (COCO, CrowdPose), the sharp confidence of the deterministic baseline is slightly more resilient against strict left-right inversions. Conversely, in structurally chaotic environments characterized by severe multi-person occlusions (OCHuman), the continuous aggregation mitigates severe errors, successfully eliminating up to 8 catastrophic swaps under extreme degradation compared to the discrete baseline.
\end{itemize}

Ultimately, while deterministic sub-pixel decoders maximize pure spatial precision in isolated contexts, our continuous TTA operates as a spatial consensus filter in chaotic, overlapping environments.

\subsection{Ablation on Global Structural Priors (The MRF Trade-off)}
The final phase of our architecture introduces a Markov Random Field (MRF) to enforce global kinematic constraints across independently sampled keypoints. Unlike the GMM, which operates within local coordinate frames, the MRF contextualizes the global skeletal structure, penalizing biomechanically implausible configurations through pairwise kinematic potentials.

To isolate the exact impact of this global constraint, we compare our continuous TTA aggregation (independent GMMs) against the same pipeline augmented with the MRF inference layer. Crucially, we also contextualize these results against \textbf{DARK (Discrete TTA)}.

\textbf{The comprehensive empirical results detailing this comparison across all degradations are provided in \suppsec{sec:supp_ablations} (\supptab{tab:supp_mrf_comprehensive}).} This analysis uncovers an unexpected algorithmic trade-off. While the unconstrained GMM effectively minimizes catastrophic inversions by relying on local density evidence, integrating the MRF exposes two opposing operational regimes:

\begin{enumerate}
    \item \textbf{Global Precision Regularizer:} The MRF acts as soft kinematic regularizer, pulling noisy local coordinate predictions slightly closer to their anatomically plausible centers. This global fine-tuning allows the MRF to consistently increase the OKS, occasionally surpassing the highly engineered \textbf{DARK (Discrete TTA)} baseline in raw coordinate precision (e.g., achieving 0.7138 OKS on COCO Clean).
    \item \textbf{The Cost of Structural Rigidity (Kinematic Error Propagation):} Enforcing rigid skeletal priors in structurally chaotic scenarios induces widespread error propagation. In heavily occluded scenes, when multiple connected keypoint hypotheses are compromised, the MRF pairwise potentials displace an otherwise accurate local GMM prediction across the image plane to satisfy a globally distorted skeleton. Under extreme degradation in dense environments (CrowdPose), this structural rigidity proves counterproductive, yielding 175 strict swaps, exceeding both the unconstrained GMM (171) and \textbf{DARK (Discrete TTA)} (169), due to 2D perspective foreshortening ambiguities (detailed in Section \ref{sec:mrf_discussion}).
\end{enumerate}

\begin{figure}[htbp]
    \centering
    \includegraphics[width=\linewidth]{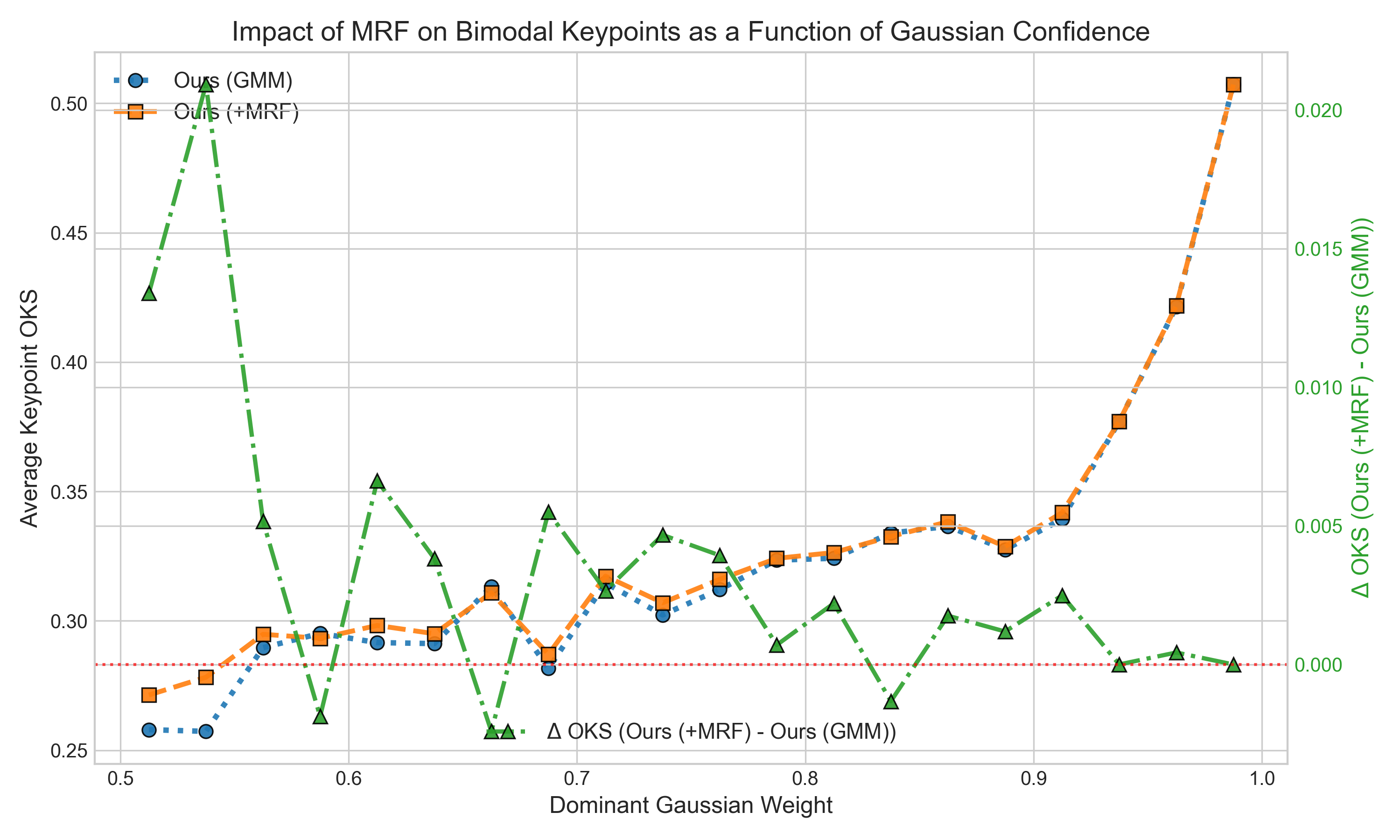}
    \caption{Impact of the MRF decoder on bimodal keypoints as a function of the dominant Gaussian weight. The MRF provides the highest precision gains when the GMM is highly uncertain ($\sim 0.5 - 0.6$), effectively leveraging structural priors to resolve local ambiguity. As the GMM confidence approaches 1.0, the MRF correctly converges to the GMM's local prediction ($\Delta$ OKS approaches zero).}
    \label{fig:mrf_weight_oks}
\end{figure}

To further unpack the mechanics of this MRF intervention, we analyze its effect specifically on \textbf{bimodal keypoints} (where the GMM detects exactly $K=2$ components and actively triggers the MRF decoder). Figure \ref{fig:mrf_weight_oks} plots the precision improvement ($\Delta$ OKS) against the confidence of the dominant Gaussian component ($\max(\pi_1, \pi_2)$).

The empirical behavior perfectly mirrors theoretical expectations for a Belief Propagation network. When the visual evidence is split evenly (Maximum Ambiguity, $\pi_{max} \approx 0.5 - 0.6$), the MRF imposes its anatomical prior to deterministically select the structurally viable component, yielding the peak precision gain. Conversely, when the GMM is overwhelmingly confident (Absolute Certainty, $\pi_{max} > 0.95$), the MRF simply ratifies the GMM's choice, and the $\Delta$ OKS converges exactly to zero.

To validate this selective optimization, we conducted a decoupled evaluation comparing independent GMM distributions against the MRF decoder across a pooled sample of 85,255 keypoints (detailed in \suppsec{sec:supp_ablations}, \supptab{tab:mrf_decoupled_breakdown}). The empirical evidence confirms two structural mechanisms:
\begin{itemize}
    \item \textbf{Invariant Anchor Points ($K=1$):} In 74.7\% of the instances ($N=63,725$), the network predicts a solitary mode; these nodes strictly bypass optimization, maintaining an invariant difference of $\Delta\text{OKS} = 0.0000 \pm 0.0000$ and eliminating unnecessary computational drift.
    \item \textbf{Targeted Ambiguity Resolution ($K=2$):} Across the 21,522 ambiguous nodes ($25.3\%$), the MRF delivers a consistent net precision improvement, achieving coordinate gains of $+2.54\%$ on COCO Clean ($0.4945 \rightarrow 0.5199$ OKS), $+1.40\%$ on CrowdPose Medium, and a net pooled gain of $+0.40\%$ across all degradation levels. 
\end{itemize}
Furthermore, the frequency of bimodal states acts as a direct metric of scene complexity, expanding from $4.9\%$ in clean images to $59.4\%$ under extreme degradation in COCO, and reaching $27.3\%$ in baseline OCHuman instances due to intrinsic occlusions.

\subsection{Uncertainty and Out-of-Distribution Detection}

Having established the foundational evaluation metrics, we evaluate the efficacy of our geometric uncertainty parameters (calibrated covariance volume \(\det(\boldsymbol{\Sigma}_{\text{final}})\) and uniform mixing weight \(\pi_u\)) against standard probabilistic baselines (the peak confidence complement \(1 - P_{\text{DARK}}\) and heatmap Shannon entropy). \textbf{All comprehensive quantitative tables, Expected Calibration Error (ECE) curves, ROC profiles, and Kernel Density Estimations (KDE) supporting this section are detailed in \suppsec{sec:supp_calibration}, \ref{sec:supp_fusion}, and \ref{sec:supp_ood}.}

\subsubsection{In-Distribution Uncertainty Calibration}
The primary objective of In-Distribution calibration is to determine whether a model's uncertainty metric correctly ranks its spatial errors, evaluated via the Area Under the Sparsification Error (AUSE) curve. 

The empirical evaluation \inSuppSec{sec:supp_calibration} demonstrates distinct operational regimes depending on keypoint visibility and occlusion density. In canonical and unoccluded scenarios (COCO and CrowdPose), the baseline confidence complement, \textbf{DARK (\(1 - P_{\text{DARK}}\))}, achieves the lowest sparsification error across global, visible, and ocluded sets (e.g., AUSE of 0.0581 on COCO and 0.0515 on CrowdPose). While local peak intensity provides a well-calibrated ranking under direct visibility, its performance degrades under severe occlusion. Specifically, within the occluded subset of OCHuman, the continuous geometric representation of \textbf{Ours (\(\det(\boldsymbol{\Sigma}_{\text{final}})\))} yields an AUSE of 0.2203, outperforming both the \textbf{DARK (\(1 - P_{\text{DARK}}\))} baseline (0.2636) and \textbf{DARK (Entropy)} (0.3130).

Furthermore, stratifying calibration by mixture cardinality reveals that the covariance volume is not merely a discrete indicator of ambiguity. Isolating bimodal hypotheses (\(K=2\)) demonstrates a near-perfect monotonic relationship between the log-determinant volume and empirical localization failure (\(1 - \text{OKS}\)).

\subsubsection{Adaptive Uncertainty Fusion}
To capitalize on the volumetric alerts of the GMM without degrading the baseline's robustness in canonical scenarios, we evaluated adaptive fusion strategies by projecting the geometric volume into the $[0, 1]$ space using a sensitivity hyperparameter $\beta$ (Equation \ref{eq:projection}).

As demonstrated in the logarithmic Grid Search \inSuppSec{sec:supp_fusion}, hard-routing mechanisms like \textbf{Ours (Max-Pool)} suffer from a structural conflict: clean scenarios demand a minuscule penalty, while severely occluded joints require massive sensitivity ($\beta \approx 39.08$). Relying on a static $\beta$ with discrete Max-Pooling results in suboptimal fusion.

\begin{figure}[htbp]
    \centering
    \includegraphics[width=\linewidth]{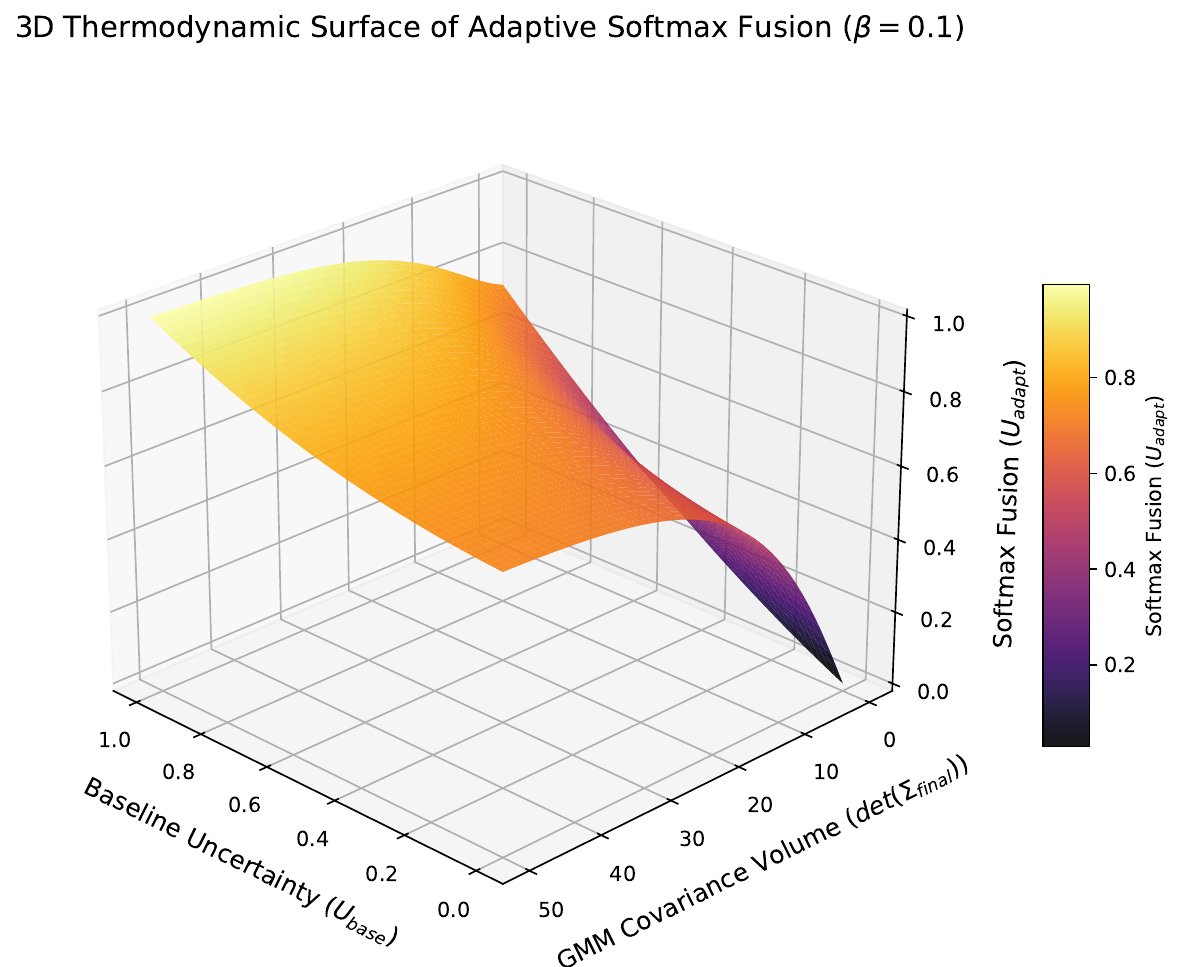}
    \caption{3D Thermodynamic Surface of the Adaptive Softmax Fusion ($\beta=0.1$). This continuous aggregation avoids the non-differentiable ridges of Max-Pooling. When the geometric volume is near zero, the surface delegates the output to the baseline confidence. However, when structural ambiguity expands the GMM volume, the exponential weighting dynamically overrides the baseline, pushing the fused uncertainty toward 1.0. This smooth transition establishes an universal sweet spot.}
    \label{fig:softmax_surface}
\end{figure}

To circumvent this oracle barrier, we transition to the continuously weighted \textbf{Ours (Softmax)}. As depicted in the 3D response surface (Figure \ref{fig:softmax_surface}), this aggregation organically weights the uncertainty contributions: when the GMM volume is compact, it smoothly follows the baseline confidence \(U_{\text{base}}\). When spatial ambiguity triggers volumetric expansion, the exponential weighting dynamically elevates the fused uncertainty. This continuous geometry yields a dataset-agnostic operating point at \(\beta = 0.1\), which uniformly optimizes AUSE across all datasets and degradation levels, preserving baseline calibration in clean regimes while integrating geometric risk warnings under severe visual corruption.

\subsubsection{Out-of-Distribution (OoD) Anomaly Detection}
OoD detection evaluates the model's capacity to identify catastrophic failures, specifically tasked with discriminating between visible joints and missing or severely occluded ones ($vis=0$).

\begin{figure*}[htbp]
    \centering
    \includegraphics[width=\textwidth]{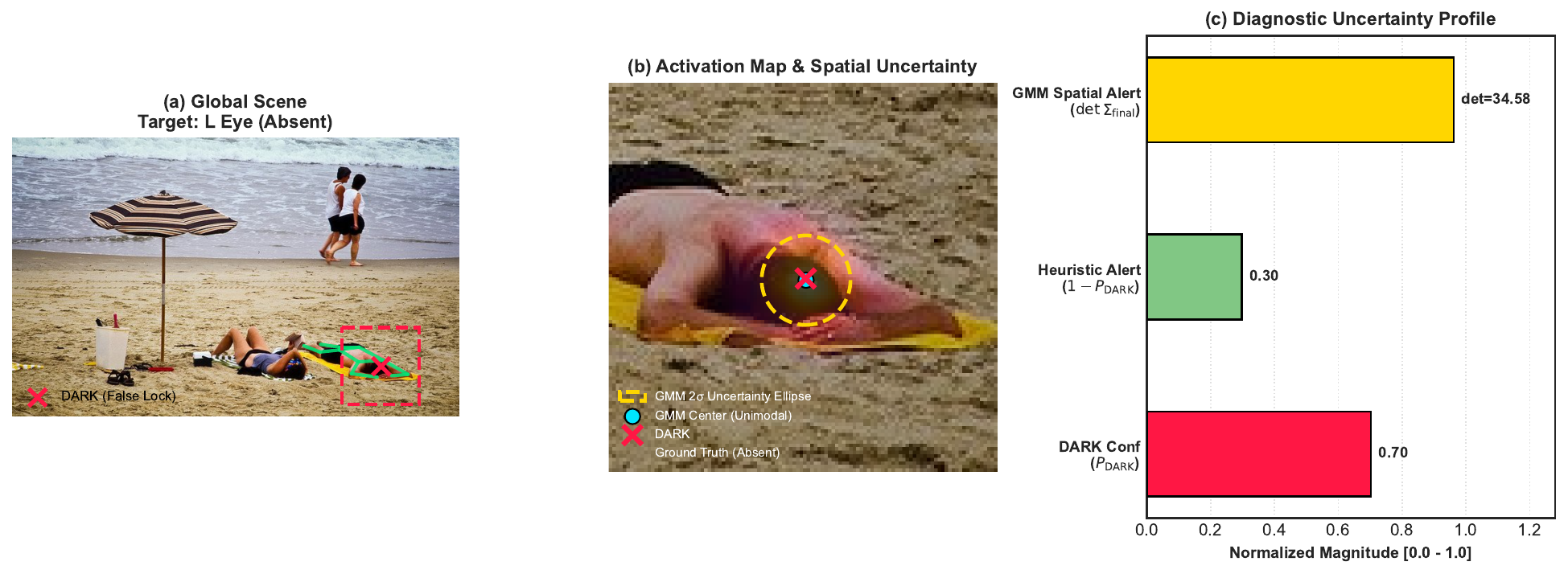}
    \caption{\textbf{Heuristic confidence vs. volumetric spatial alert in Out-of-Distribution (OoD) absence detection.} 
    \textbf{(a)} Scene from CrowdPose where the target keypoint (left eye) is completely absent. The deterministic baseline (\textbf{DARK}) commits a false lock onto the background (red cross). 
    \textbf{(b)} Spatial activation overlay and continuous GMM fit. Despite the absence of visual evidence, DARK reports a misleadingly confident peak ($P_{\text{DARK}} = 0.70$). In stark contrast, our continuous formulation captures the underlying spatial diffusion, yielding an expanded uncertainty ellipse ((\(\det(\boldsymbol{\Sigma}_{\text{final}}) = 34.58\), dashed gold ring). 
    \textbf{(c)} Diagnostic uncertainty profile. While the baseline heuristic alert ($1 - P_{DARK} = 0.30$) projects false confidence, our calibrated geometric volume alert ((\(\det(\boldsymbol{\Sigma}_{\text{final}}) = 34.58\))) saturates to maximum uncertainty, reliably flagging the missing keypoint.}
    \label{fig:qualitative_ood_crowd}
\end{figure*}

The global quantitative analysis \inSuppSec{sec:supp_ood} exposes a vulnerability in standard peak-confidence heuristics (\(1 - P_{\text{DARK}}\)). In dense, overlapping environments (CrowdPose, OCHuman), when the model fits a unimodal distribution (\(K=1\)), \textbf{DARK (\(1 - P_{\text{DARK}}\))} systematically collapses, yielding an AUROC of 0.38 on CrowdPose (substantially below random chance, indicating inverted confidence rankings). As illustrated by the CrowdPose failure case in Figure \ref{fig:qualitative_ood_crowd}, when a target joint is absent (\(\text{vis} = 0\), Figure \ref{fig:qualitative_ood_crowd}a), the deterministic network erroneously locks onto background clutter, producing a high activation peak of \(P_{\text{DARK}} = 0.70\) (Figure \ref{fig:qualitative_ood_crowd}b). The resulting baseline alert (\(1 - P_{\text{DARK}} = 0.30\)) fails to identify this structural failure (Figure \ref{fig:qualitative_ood_crowd}c).

In stark contrast, \textbf{Ours (Volume)} evaluates the continuous spatial dispersion of the activation mass rather than relying on a discrete peak intensity. When local visual evidence is absent (\(\text{vis} = 0\)), Monte Carlo sampling across diffuse activations produces a broadly dispersed coordinate distribution, driving a substantial expansion of the calibrated covariance volume (\(\det(\boldsymbol{\Sigma}_{\text{final}}) = 34.58\), Figure \ref{fig:qualitative_ood_crowd}b). This volumetric inflation effectively penalizes confident yet spurious activations where the baseline collapses. Consequently, on CrowdPose, \textbf{Ours (Volume)} achieves an AUROC of 0.9435 on unimodal predictions (\(K=1\)), where \textbf{DARK (\(1 - P_{\text{DARK}}\))} plunges to 0.3838, and maintains a dominant global AUROC of 0.8788 (versus 0.5160 for the baseline). In parallel, the orthogonal uniform mixture weight \(\pi_u\) provides a dedicated non-geometric metric, rising under severe visual corruption to signal complete anatomical absence.

\section{Discussion and Limitations}

The empirical evaluation of our probabilistic framework demonstrates the analytical advantages of transitioning from deterministic coordinate extraction to continuous spatial modeling in Human Pose Estimation. By parameterizing activation heatmaps as Gaussian Mixture Models (GMM) augmented with an orthogonal uniform distribution, the framework captures the spatial dispersion of keypoint likelihood while exposing critical algorithmic trade-offs.

\subsection{The Continuous Paradigm: Precision and Stochastic Aggregation}
Our initial analysis confirms that evaluating the spatial expectation of the primary Gaussian mode (\(\boldsymbol{\mu}_1\)) achieves geometric parity with state-of-the-art deterministic sub-pixel decoders, such as DARK (Table \ref{tab:baseline_precision}). However, the inferential divergence between discrete coordinate decoders and continuous density aggregation becomes explicit during Test-Time Augmentation (TTA).

As detailed in \supptab{tab:supp_tta_comprehensive}, standard TTA retains a marginal advantage in absolute coordinate precision across canonical datasets, leading by $0.0002$ to $0.0020$ OKS on COCO and CrowdPose. Nevertheless, evaluating structural robustness highlights the specific utility of continuous density aggregation. By fusing continuous probability distributions rather than discrete spatial grids, the GMM operates as a spatial filter, systematically mitigating minor bilateral ambiguities (Loose Swaps) across evaluated conditions (e.g., eliminating 16 loose swaps under High degradation in COCO and 18 under Extreme degradation in OCHuman).

Conversely, regarding catastrophic anatomical inversions (Strict Swaps), the continuous formulation behaves as a context-dependent regularizer. In clean or isolated scenes (COCO, CrowdPose), where high activation peaks rarely mislead the decoder, the GMM exhibits minor increases in strict swaps (\(+1\) to \(+3\)) compared to the confident deterministic baseline. However, in complex multi-person scenes characterized by severe inter-subject occlusions (OCHuman), continuous density aggregation outperforms standard averaging, successfully eliminating 8 catastrophic swaps under extreme degradation ($249$ vs. $257$).

\subsection{Kinematic Constraints and the 2D MRF Trade-off}
\label{sec:mrf_discussion}
The explicit modeling of mixture cardinality states (\(K = 1\) vs. \(K = 2\)) naturally facilitates the integration of global skeletal priors, such as Markov Random Fields (MRF), to resolve local coordinate ambiguities. As evidenced by our ablations in the Supplementary Material (\supptab{tab:supp_mrf_comprehensive} and \supptab{tab:mrf_decoupled_breakdown}) and Figure \ref{fig:mrf_weight_oks}, the MRF operates as a global regularizer in canonical scenarios, refining noisy predictions toward anatomically plausible centers (e.g., reaching $0.7138$ OKS on COCO Clean in \supptab{tab:supp_mrf_comprehensive}, and yielding a \(+2.54\%\) OKS improvement on ambiguous bimodal nodes in \supptab{tab:mrf_decoupled_breakdown}).

\begin{figure*}[htbp]
    \centering
    \includegraphics[width=\textwidth]{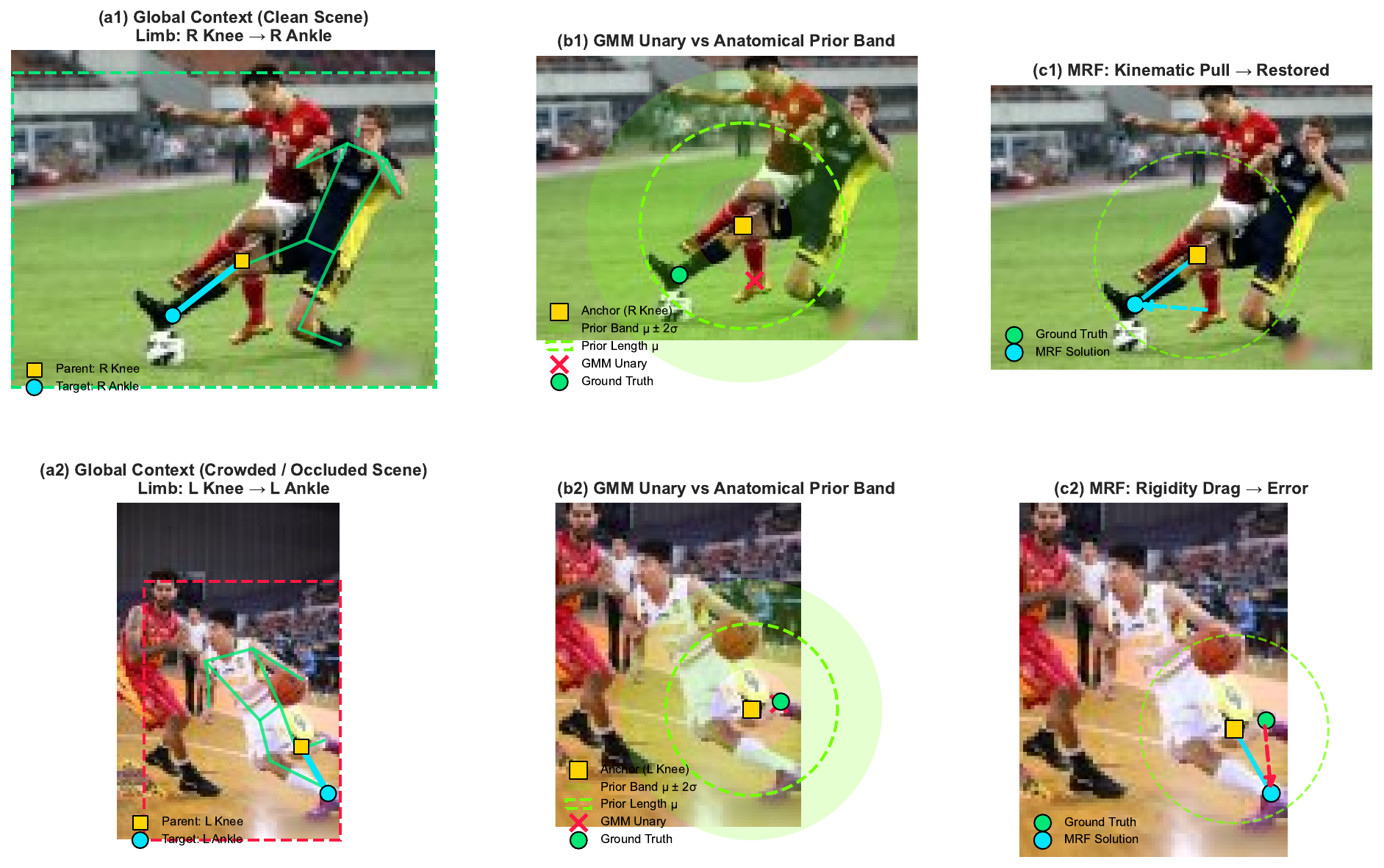}
    \caption{\textbf{The limitation of 2D kinematic regularizers: Constructive alignment vs. Projective foreshortening error propagation.} 
    \textbf{Top Row (Constructive Kinematic Guidance in Canonical Poses):} In un-foreshortened poses \textbf{(a1)}, local GMM sampling jitter leaves the initial prediction retracted (\(d = 18\,\text{px}\), red cross in \textbf{b1}) relative to the empirical prior band (\(\mu = 34\,\text{px}\), green dashed ring). In \textbf{(c1)}, the pairwise kinematic potential guides the ankle toward the anatomically expected configuration (\(d = 26\,\text{px}\), cyan arrow), improving localization accuracy. 
    \textbf{Bottom Row (Destructive Rigidity under Projective Foreshortening):} In scenes with acute camera perspective \textbf{(a2)}, the lower limb extends along the optical axis, naturally projecting as a compressed segment in 2D (\(d = 8\,\text{px}\) to Ground Truth, \textbf{b2}). While the independent GMM localizes this foreshortened joint accurately, the rigid 2D Euclidean prior (\(\mu = 24\,\text{px}\)) penalizes the perspective shortening as an anatomical violation. Consequently, the MRF pairwise potential displaces the coordinate by \(19\,\text{px}\) across the image plane (\textbf{c2}, red dashed arrow) to satisfy the distorted 2D graph, inducing a catastrophic strict swap.}
    \label{fig:qualitative_mrf_tradeoff}
\end{figure*}

However, as visually demonstrated in Figure \ref{fig:qualitative_mrf_tradeoff}, imposing skeletal priors exclusively in two-dimensional image coordinates exposes a limitation of projective geometry. Because the perspective projection mapping \(\mathbb{R}^3 \to \mathbb{R}^2\) discards depth along the optical axis (\(Z\)), pairwise potentials defined solely over Euclidean image distances cannot mathematically distinguish between a biomechanically invalid bone length and an anatomically sound limb undergoing severe \textit{projective foreshortening}.

When a limb extends along the optical axis toward the camera lens (Figure \ref{fig:qualitative_mrf_tradeoff}, Bottom Row), its physical length is compressed on the 2D focal plane (d = 8 px).  While the unconstrained GMM preserves the local visual evidence and correctly centers its mode on the foreshortened joint, the pairwise potential (Equation~\eqref{eq:mrf_binary}) penalizes this valid configuration against an unforeshortened prior mean (\(\mu = 24\,\text{px}\)). The graph optimization consequently displaces the coordinate by \(19\,\text{px}\), overriding a correct local prediction and inducing a catastrophic limb swap. This explains the quantitative degradation observed in \supptab{tab:supp_mrf_comprehensive}, where enforcing global 2D kinematics under extreme corruption increases strict swaps on CrowdPose to $175$ (exceeding both unconstrained GMM at $171$ and DARK at $169$), and on OCHuman Extreme to $254$ (compared to $249$ for GMM).

This failure mode highlights the inherent geometric ceiling of 2D kinematic regularizers. It provides the primary theoretical motivation for extending our continuous probabilistic framework into three-dimensional Euclidean space (Section \ref{sec:future_3d}), where absolute physical bone dimensions can be evaluated invariantly to camera viewpoint and perspective foreshortening.

\subsection{The Out-of-Distribution Paradigm and Adaptive Fusion}
A central analytical contribution of this framework is the isolation and quantification of Out-of-Distribution (OoD) keypoints, exposing a vulnerability in standard heuristic confidence metrics (\(1 - P_{\text{DARK}}\)) within crowded scenes. As demonstrated in \suppfig{fig:ood_combined_matrix}e and \supptab{tab:ood_global_auc}, when the deterministic baseline outputs a single coordinate (\(K=1\)) under heavy occlusion in CrowdPose, its confidence ranking systematically collapses, yielding an AUROC of 0.3838 (substantially below random chance). In these instances, the neural network confidently locks onto background clutter or limbs from neighboring individuals, assigning high confidence to catastrophic spatial errors.

The explicitly parameterized geometric covariance volume \(\det(\boldsymbol{\Sigma}_{\text{final}})\) fundamentally resolves this failure mode. By evaluating continuous spatial dispersion across the entire activation mass rather than relying on local peak height, the calibrated covariance volume inflates under diffuse evidence, achieving a dominant AUROC of 0.9435 on unimodal predictions (\(K=1\)) in CrowdPose (\suppfig{fig:ood_combined_matrix}e). Furthermore, when competing candidate modes emerge, the calibrated covariance volume \(\det(\boldsymbol{\Sigma}_{\text{final}})\) (Equation~\eqref{eq:sigma_final}) scales quadratically with inter-component displacement along the axis of bilateral confusion, as formulated in the total covariance matrix (Equation~\eqref{eq:total_variance}). In parallel, the orthogonal uniform weight \(\pi_u\) provides a mathematically decoupled indicator of visual absence, rising systematically as image degradation escalates (\suppfig{fig:uniform_weight_all}).

To integrate this volumetric awareness into the standard prediction flow, the Adaptive Softmax Fusion dynamically balances baseline confidence \(U_{\text{base}}\) with continuous GMM dispersion \(U_{\text{gmm}}\). As depicted in the continuous response surface (Figure \ref{fig:softmax_surface}), this aggregation organically resolves the structural sensitivity conflicts encountered when optimizing static thresholds in Max-Pooling \inSuppTab{tab:optimal_betas_combined}. Consequently, setting a static operating point of \(\beta = 0.1\) establishes a robust inference pipeline that optimizes or matches the Sparsification Error (AUSE) across all datasets and degradation levels without requiring oracle tuning \inSuppTab{tab:ause_fusion_final}.

\subsection{Limitations and Future 3D Extensibility}
\label{sec:future_3d}
Despite its strong analytical performance, the proposed methodology introduces a notable computational overhead. While our framework requires zero additional learnable parameters, \supptab{tab:computational_cost} and \suppfig{fig:supp_computational_breakdown} detail the comprehensive latency, throughput (FPS), stage-by-stage profiling, and peak GPU memory (VRAM) across all evaluated backbones on standard COCO resolution (\(256 \times 192\)).

In our research prototype, activation heatmaps are transferred to host RAM to execute post-processing entirely on the CPU, preserving modularity and framework independence. As evidenced by the profiling breakdown, while continuous TTA aggregation (\(3.2\text{--}3.6\,\text{ms}\)) and Belief Propagation over the MRF kinematic tree (\(1.58\text{--}1.62\,\text{ms}\)) introduce negligible overhead, the sequential CPU execution of vectorized Monte Carlo sampling (\(80\text{--}100\,\text{ms}\)), iterative Expectation-Maximization (EM) optimization, and BIC model selection across all 17 keypoints consumes \(1311.7\text{--}1391.6\,\text{ms}\) per instance. Consequently, in its current unoptimized, single-threaded Python implementation on CPU, the full pipeline operates at \(0.5\text{--}0.7\,\text{FPS}\), making it unsuitable for real-time edge deployment.

Crucially, this latency bottleneck is localized in the iterative mixture fitting rather than in graph kinematics or memory transfers. Because Monte Carlo sampling and GMM clustering are conditionally independent across the 17 anatomical joints, the post-processing module is decoupled across keypoints and fully parallelizable. Implementing the EM algorithm as native batch-tensor GPU operations would substantially reduce post-processing latency by evaluating all keypoints concurrently, providing an immediate pathway toward interactive frame rates. Furthermore, the memory footprint remains exceptionally modest (\(118\text{--}265\,\text{MB}\) peak VRAM across all backbones), confirming that continuous uncertainty extraction does not impose GPU memory bottlenecks.

However, this mathematical formulation lays the groundwork for a future extension: Multi-View 3D Human Pose Estimation. Standard multi-view approaches project 2D heatmaps from various cameras into massive 3D volumetric grids to find spatial intersections \cite{tu2020voxelpose}, a process that suffers heavily from the curse of dimensionality (\(\mathcal{O}(N^3)\) memory scaling). Because our framework parameterizes 2D heatmaps into continuous probability distributions, it is inherently resolution-agnostic. By crossing the mathematical rays of 2D GMMs from multiple calibrated views, one could directly triangulate a continuous 3D GMM (\(\boldsymbol{\mu} \in \mathbb{R}^3\), \(\boldsymbol{\Sigma} \in \mathbb{R}^{3 \times 3}\)) algebraically, circumventing the creation of dense voxel grids entirely. Crucially, applying our MRF algorithm directly over these 3D Gaussian components would evaluate Euclidean distances in physical space, rendering the kinematic prior invariant to 2D camera projections and neutralizing the foreshortening ambiguities encountered in the 2D formulation (Section \ref{sec:mrf_discussion}).
\section{Conclusions}

This work introduces a principled probabilistic framework that transitions Human Pose Estimation from deterministic coordinate extraction to continuous, density-aware spatial modeling. By intercepting raw activation heatmaps and fitting a parametric Gaussian Mixture Model coupled with an orthogonal uniform noise sink, our approach matches the sub-pixel geometric precision of highly optimized decoders (such as DARK) while unlocking a continuous parameter space for heteroscedastic uncertainty quantification.

Our comprehensive empirical evaluations reveal several insights into continuous density modeling under visual corruption. First, aggregating probability densities directly within the continuous parameter domain serves as a spatial consensus filter during Test-Time Augmentation (TTA). This continuous aggregation mitigates minor bilateral ambiguities (loose swaps) across diverse benchmarks and eliminates catastrophic anatomical inversions under severe multi-person occlusion.

Second, parameterizing explicit mixture cardinality states (\(K=1\) vs. \(K=2\)) establishes an objective interface for kinematic graph optimization. Our analysis delineates a clear operational boundary for 2D geometric priors: while pairwise kinematic potentials act as effective sub-pixel regularizers in canonical poses, enforcing rigid 2D skeletal structures induces error propagation when perspective foreshortening compresses limb projections along the optical axis. This geometric limitation provides a rationale for extending continuous mixture formulations into three-dimensional Euclidean space.

Third, our framework exposes a failure mode in conventional heuristic confidence metrics (\(1 - P_{\text{DARK}}\)) within crowded scenes. We demonstrate that deterministic baselines collapse under dense occlusions by assigning high confidence to misplaced background activations. In contrast, the calibrated covariance volume \(\det(\boldsymbol{\Sigma}_{\text{final}})\) inflates under diffuse evidence and inter-component dispersion, reliably detecting Out-of-Distribution failures. In parallel, the orthogonal uniform weight \(\pi_u\) provides a decoupled indicator of anatomical absence.

Finally, by continuously weighting these complementary uncertainty signals via Adaptive Softmax Fusion, the proposed pipeline establishes a parameter-agnostic operating point (\(\beta = 0.1\)) that optimizes error calibration across all datasets without requiring image-specific oracle tuning. Looking forward, this continuous formulation establishes a memory-efficient foundation for Multi-View 3D Pose Estimation, enabling direct algebraic ray triangulation that circumvents the volumetric complexity (\(\mathcal{O}(N^3)\)) of dense voxel grids while resolving 2D perspective foreshortening.
\section*{CRediT Authorship Contribution Statement}
\textbf{Santiago Reina Alguacil:} Conceptualization, Methodology, Software, Validation, Formal analysis, Investigation, Data curation, Writing - original draft, Writing - review \& editing, Visualization. \\
\textbf{Ezequiel L\'{o}pez Rubio:} Conceptualization, Methodology, Validation, Resources, Writing - review \& editing, Supervision, Project administration.

\section*{Data and Code Availability}
The benchmark datasets analyzed in this study are publicly available in the MS COCO (\url{https://cocodataset.org}), CrowdPose (\url{https://github.com/Jeff-sjtu/CrowdPose}), and OCHuman (\url{https://github.com/liruilong940607/OCHumanApi}) repositories. The complete source code, configuration files, and evaluation scripts supporting the findings of this study are openly accessible on GitHub at \url{https://github.com/santiago-git19/probabilistic-pose-gmm.git}.

\section*{Declaration on the Use of Generative AI Tools}
The authors acknowledge the use of Gemini as an auxiliary tool during the preparation of this work. Specifically, the model provided assistive support in preliminary text drafting, language and style refinement, source code documentation, and routine scripting for data visualization and figure rendering. 

The core conceptual methodology, mathematical derivations (including the continuous GMM parameterization, scale-aware TTA aggregation, and kinematic MRF decoding), primary software architecture, and experimental benchmarks were conceived, implemented, and executed entirely by the authors. All AI-assisted textual suggestions, code annotations, and plotting routines were critically reviewed, modified, and validated by the authors, who assume full and unreserved responsibility for the integrity, correctness, and reproducibility of all results and code presented in this manuscript.

\section*{Acknowledgements}
This work is based on research developed as part of the Bachelor's Thesis of S.~Reina-Alguacil at the Department of Computer Languages and Computer Science, University of M\'{a}laga. This work is partially supported by the Autonomous Government of Andalusia (Spain) under project PPRO-TIC163-G-2023 (TIC163-G-FEDER) and grant DGP\_PIDI\_2024\_00462; by the Ministry of Science and Innovation of Spain, grant PID2022-136764OA-I00; and by the Instituto de Salud Carlos III, project PI25/02129 (co-financed by the European Union). It includes funds from the European Regional Development Fund (ERDF). The authors acknowledge the computer resources and technical assistance provided by the SCBI center of the University of M\'alaga, the donation of an RTX A6000 GPU by NVIDIA Corporation, and the support of the Universidad de M\'alaga and the Instituto de Investigaci\'on Biom\'edica de M\'alaga y Plataforma en Nanomedicina-IBIMA Plataforma BIONAND.

\bibliographystyle{IEEEtran}
\bibliography{references/references}

\clearpage
\appendix
\clearpage
\onecolumn 

\setcounter{section}{0}
\setcounter{table}{0}
\setcounter{figure}{0}
\setcounter{equation}{0}

\renewcommand{\thesection}{\Alph{section}}
\renewcommand{\thetable}{S\arabic{table}}
\renewcommand{\thefigure}{S\arabic{figure}}
\renewcommand{\theequation}{S\arabic{equation}}

\begin{center}
    \vspace*{0.5cm}
    {\LARGE \textbf{Supplementary Material for:}} \\[0.6em]
    {\Large \textbf{Human Pose Estimation by Probabilistic Mixtures of Gaussian and Uniform Distributions}} \\[1.2em]
    {\large Santiago Reina Alguacil\(^{1}\), Ezequiel L\'{o}pez Rubio\(^{1,2,3}\)} \\[0.8em]
    {\small \(^{1}\)Department of Computer Languages and Computer Science, University of M\'{a}laga, Spain} \\
    {\small \(^{2}\)IBIMA Plataforma BIONAND, M\'{a}laga, Spain} \quad
    {\small \(^{3}\)ITIS Software, University of M\'{a}laga, Spain} \\
    \vspace*{1.5cm}
\end{center}

\section{Extended Ablation Studies (HRNet-W32)}
\label{sec:supp_ablations}
This section provides the detailed quantitative breakdown of the ablation studies discussed in the main manuscript, detailing performance across all datasets (COCO, CrowdPose, OCHuman) and all levels of visual degradation.

Table \ref{tab:supp_tta_comprehensive} details the impact of continuous stochastic aggregation compared to discrete Test-Time Augmentation. Table \ref{tab:supp_mrf_comprehensive} isolates the effect of the Markov Random Field (MRF) global structural prior.

\begin{table*}[htbp]
\centering
\small
\resizebox{\textwidth}{!}{%
\setlength{\tabcolsep}{4.5pt} 
\begin{tabular}{@{}llccc|ccc|ccc@{}}
\toprule
\multirow{2}{*}{\textbf{Dataset}} & \multirow{2}{*}{\textbf{Degradation}} & \multicolumn{3}{c|}{\textbf{OKS}} & \multicolumn{3}{c|}{\textbf{Loose Swaps}} & \multicolumn{3}{c}{\textbf{Strict Swaps}} \\ \cmidrule(l){3-11}
 & & \textbf{DARK} & \textbf{GMM} & \textbf{$\Delta$} & \textbf{DARK} & \textbf{GMM} & \textbf{$\Delta$} & \textbf{DARK} & \textbf{GMM} & \textbf{$\Delta$} \\ \midrule

\multirow{5}{*}{COCO} & Baseline (Clean) & \textbf{0.7137} & 0.7129 & \textcolor{red}{-0.0008} & 98 & \textbf{97} & \textcolor{teal}{-1} & 18 & 18 & 0 \\
 & Resize Low & \textbf{0.6912} & 0.6907 & \textcolor{red}{-0.0005} & 116 & \textbf{113} & \textcolor{teal}{-3} & 26 & 26 & 0 \\
 & Resize Medium & \textbf{0.6082} & 0.6076 & \textcolor{red}{-0.0007} & 215 & \textbf{214} & \textcolor{teal}{-1} & \textbf{66} & 67 & \textcolor{red}{+1} \\
 & Resize High & \textbf{0.4485} & 0.4465 & \textcolor{red}{-0.0020} & 404 & \textbf{388} & \textcolor{teal}{-16} & \textbf{135} & 138 & \textcolor{red}{+3} \\
 & Resize Extreme & \textbf{0.2505} & 0.2494 & \textcolor{red}{-0.0011} & 438 & \textbf{437} & \textcolor{teal}{-1} & 139 & 139 & 0 \\ \midrule
\multirow{5}{*}{CrowdPose} & Baseline (Clean) & \textbf{0.8361} & 0.8355 & \textcolor{red}{-0.0005} & \textbf{150} & 151 & \textcolor{red}{+1} & 38 & \textbf{36} & \textcolor{teal}{-2} \\
 & Resize Low & \textbf{0.8312} & 0.8302 & \textcolor{red}{-0.0009} & 155 & \textbf{153} & \textcolor{teal}{-2} & \textbf{36} & 38 & \textcolor{red}{+2} \\
 & Resize Medium & \textbf{0.7956} & 0.7954 & \textcolor{red}{-0.0002} & 198 & \textbf{191} & \textcolor{teal}{-7} & \textbf{53} & 54 & \textcolor{red}{+1} \\
 & Resize High & \textbf{0.6761} & 0.6755 & \textcolor{red}{-0.0005} & 336 & \textbf{329} & \textcolor{teal}{-7} & \textbf{102} & 103 & \textcolor{red}{+1} \\
 & Resize Extreme & \textbf{0.4389} & 0.4383 & \textcolor{red}{-0.0007} & \textbf{497} & 498 & \textcolor{red}{+1} & \textbf{169} & 171 & \textcolor{red}{+2} \\ \midrule
\multirow{5}{*}{OCHuman} & Baseline (Clean) & 0.6356 & 0.6356 & 0.0000 & 441 & \textbf{437} & \textcolor{teal}{-4} & 135 & \textbf{132} & \textcolor{teal}{-3} \\
 & Resize Low & \textbf{0.6305} & 0.6299 & \textcolor{red}{-0.0006} & 443 & \textbf{442} & \textcolor{teal}{-1} & \textbf{127} & 131 & \textcolor{red}{+4} \\
 & Resize Medium & 0.6100 & \textbf{0.6106} & \textcolor{teal}{+0.0006} & 467 & \textbf{458} & \textcolor{teal}{-9} & \textbf{126} & 128 & \textcolor{red}{+2} \\
 & Resize High & \textbf{0.5342} & 0.5337 & \textcolor{red}{-0.0005} & 544 & \textbf{540} & \textcolor{teal}{-4} & 139 & \textbf{138} & \textcolor{teal}{-1} \\
 & Resize Extreme & 0.3574 & \textbf{0.3578} & \textcolor{teal}{+0.0005} & 738 & \textbf{720} & \textcolor{teal}{-18} & 257 & \textbf{249} & \textcolor{teal}{-8} \\ \bottomrule
\end{tabular}%
}
\vspace{2mm}
\caption{Ablation on Stochastic Aggregation: Comparison of DARK (Discrete TTA) vs. continuous GMM aggregation across three datasets and five degradation levels. The analysis highlights an architectural trade-off: while the discrete DARK baseline achieves slightly higher raw OKS, continuous aggregation acts as an effective spatial filter, reducing minor bilateral ambiguities (Loose Swaps). For catastrophic inversions (Strict Swaps), the GMM yields minor regressions in canonical datasets (COCO, CrowdPose) but outperforms the discrete baseline in severely occluded environments (OCHuman) (\textbf{DARK}: Discrete TTA, \textbf{GMM}: Ours (GMM)).}
\label{tab:supp_tta_comprehensive}
\end{table*}

\begin{table*}[htbp]
\centering
\small
\resizebox{\textwidth}{!}{%
\setlength{\tabcolsep}{4.5pt} 
\begin{tabular}{@{}llcccc|cccc|cccc@{}}
\toprule
\multirow{2}{*}{\textbf{Dataset}} & \multirow{2}{*}{\textbf{Degradation}} & \multicolumn{4}{c|}{\textbf{OKS}} & \multicolumn{4}{c|}{\textbf{Loose Swaps}} & \multicolumn{4}{c}{\textbf{Strict Swaps}} \\ \cmidrule(l){3-14}
 & & \textbf{DARK} & \textbf{GMM} & \textbf{+MRF} & \textbf{$\Delta_{M,G}$} & \textbf{DARK} & \textbf{GMM} & \textbf{+MRF} & \textbf{$\Delta_{M,G}$} & \textbf{DARK} & \textbf{GMM} & \textbf{+MRF} & \textbf{$\Delta_{M,G}$} \\ \midrule

\multirow{5}{*}{COCO}
 & Baseline (Clean) & 0.7137 & 0.7129 & \textbf{0.7138} & \textcolor{teal}{+0.0008} & 98 & \textbf{97} & 101 & \textcolor{red}{+4} & \textbf{18} & \textbf{18} & 20 & \textcolor{red}{+2} \\
 & Resize Low & \textbf{0.6912} & 0.6907 & 0.6911 & \textcolor{teal}{+0.0004} & 116 & \textbf{113} & \textbf{113} & 0 & \textbf{26} & \textbf{26} & 28 & \textcolor{red}{+2} \\
 & Resize Medium & \textbf{0.6082} & 0.6076 & 0.6071 & \textcolor{red}{-0.0004} & 215 & \textbf{214} & 215 & \textcolor{red}{+1} & \textbf{66} & 67 & 70 & \textcolor{red}{+3} \\
 & Resize High & \textbf{0.4485} & 0.4465 & 0.4481 & \textcolor{teal}{+0.0016} & 404 & \textbf{388} & 394 & \textcolor{red}{+6} & \textbf{135} & 138 & 142 & \textcolor{red}{+4} \\
 & Resize Extreme & 0.2505 & 0.2494 & \textbf{0.2524} & \textcolor{teal}{+0.0030} & 438 & 437 & \textbf{435} & \textcolor{teal}{-2} & 139 & 139 & \textbf{135} & \textcolor{teal}{-4} \\ \midrule
\multirow{5}{*}{CrowdPose}
 & Baseline (Clean) & \textbf{0.8361} & 0.8355 & 0.8354 & \textcolor{red}{-0.0001} & \textbf{150} & 151 & 153 & \textcolor{red}{+2} & 38 & \textbf{36} & 39 & \textcolor{red}{+3} \\
 & Resize Low & \textbf{0.8312} & 0.8302 & 0.8307 & \textcolor{teal}{+0.0005} & 155 & \textbf{153} & 155 & \textcolor{red}{+2} & \textbf{36} & 38 & 40 & \textcolor{red}{+2} \\
 & Resize Medium & 0.7956 & 0.7954 & \textbf{0.7971} & \textcolor{teal}{+0.0018} & 198 & \textbf{191} & 197 & \textcolor{red}{+6} & \textbf{53} & 54 & 58 & \textcolor{red}{+4} \\
 & Resize High & 0.6761 & 0.6755 & \textbf{0.6767} & \textcolor{teal}{+0.0012} & 336 & \textbf{329} & 338 & \textcolor{red}{+9} & \textbf{102} & 103 & 106 & \textcolor{red}{+3} \\
 & Resize Extreme & \textbf{0.4389} & 0.4383 & 0.4389 & \textcolor{teal}{+0.0006} & 497 & 498 & \textbf{493} & \textcolor{teal}{-5} & \textbf{169} & 171 & 175 & \textcolor{red}{+4} \\ \midrule
\multirow{5}{*}{OCHuman}
 & Baseline (Clean) & \textbf{0.6356} & 0.6356 & 0.6345 & \textcolor{red}{-0.0011} & 441 & 437 & \textbf{434} & \textcolor{teal}{-3} & 135 & \textbf{132} & 134 & \textcolor{red}{+2} \\
 & Resize Low & 0.6305 & 0.6299 & \textbf{0.6319} & \textcolor{teal}{+0.0020} & 443 & \textbf{442} & \textbf{442} & 0 & \textbf{127} & 131 & 134 & \textcolor{red}{+3} \\
 & Resize Medium & 0.6100 & 0.6106 & \textbf{0.6120} & \textcolor{teal}{+0.0014} & 467 & \textbf{458} & 461 & \textcolor{red}{+3} & \textbf{126} & 128 & 131 & \textcolor{red}{+3} \\
 & Resize High & 0.5342 & 0.5337 & \textbf{0.5343} & \textcolor{teal}{+0.0005} & 544 & \textbf{540} & 542 & \textcolor{red}{+2} & 139 & \textbf{138} & 142 & \textcolor{red}{+4} \\
 & Resize Extreme & 0.3574 & 0.3578 & \textbf{0.3579} & \textcolor{teal}{+0.0001} & 738 & 720 & \textbf{718} & \textcolor{teal}{-2} & 257 & \textbf{249} & 254 & \textcolor{red}{+5} \\ \bottomrule
\end{tabular}%
}
\vspace{2mm}
\caption{Ablation on Global Structural Priors: Isolating the effect of the MRF structural constraint applied over continuous GMM probability distributions, compared against DARK (Discrete TTA). While the MRF generally improves global coordinate precision (OKS) relative to the independent GMM, enforcing this global skeletal prior increases bilateral swap errors (both Loose and Strict Swaps), occasionally exceeding the swap error rates of discrete TTA in crowded, degraded scenarios (\textbf{DARK}: Discrete TTA, \textbf{GMM}: Ours (GMM), \textbf{+MRF}: Ours (+MRF)).}
\label{tab:supp_mrf_comprehensive}
\end{table*}

\begin{table*}[htbp]
\centering
\caption{\textbf{Decoupled Performance Breakdown of the MRF Structural Prior on HRNet-W32.} Evaluating the net impact of the pairwise kinematic constraints over 85,255 keypoints. Unimodal nodes ($K=1$) act as invariant anchors ($\Delta\text{OKS}=0$), while the MRF selectively optimizes bimodal ambiguities ($K=2$), delivering consistent precision gains.}
\label{tab:mrf_decoupled_breakdown}
\resizebox{\textwidth}{!}{%
\begin{tabular}{ll|ccc|ccc}
\toprule
\multirow{2}{*}{\textbf{Dataset}} & \multirow{2}{*}{\textbf{Degradation}} & \multicolumn{3}{c|}{\textbf{Unimodal Anchors ($K=1$)}} & \multicolumn{3}{c}{\textbf{Ambiguous Nodes ($K=2$)}} \\
 & & \textbf{Ratio (\%)} & \textbf{Ours (TTA)} & \textbf{$\Delta_{\text{MRF}}$} & \textbf{Ratio (\%)} & \textbf{Ours (TTA)} & \textbf{$\Delta_{\text{MRF}}$ (Gain)} \\
\midrule
\multirow{5}{*}{COCO} 
 & Baseline (Clean) & 95.1\% & $0.8851 \pm 0.18$ & $0.0000$ & 4.9\%  & $0.4945 \pm 0.38$ & \textbf{+0.0254 (+2.54\%)} \\
 & Resize Low       & 93.6\% & $0.8649 \pm 0.20$ & $0.0000$ & 6.4\%  & $0.4517 \pm 0.38$ & \textbf{+0.0021 (+0.21\%)} \\
 & Resize Medium    & 86.1\% & $0.8002 \pm 0.27$ & $0.0000$ & 13.9\% & $0.3106 \pm 0.37$ & $-0.0001$ ($-0.01\%$) \\
 & Resize High      & 67.1\% & $0.6843 \pm 0.34$ & $0.0000$ & 32.9\% & $0.2106 \pm 0.33$ & \textbf{+0.0060 (+0.60\%)} \\
 & Resize Extreme   & 40.6\% & $0.4948 \pm 0.38$ & $0.0000$ & 59.4\% & $0.1285 \pm 0.26$ & \textbf{+0.0056 (+0.56\%)} \\
\midrule
\multirow{5}{*}{CrowdPose} 
 & Baseline (Clean) & 91.0\% & $0.8761 \pm 0.22$ & $0.0000$ & 9.0\%  & $0.4639 \pm 0.41$ & $-0.0009$ ($-0.09\%$) \\
 & Resize Low       & 90.4\% & $0.8719 \pm 0.22$ & $0.0000$ & 9.5\%  & $0.4697 \pm 0.41$ & \textbf{+0.0030 (+0.30\%)} \\
 & Resize Medium    & 87.5\% & $0.8451 \pm 0.25$ & $0.0000$ & 12.4\% & $0.4664 \pm 0.40$ & \textbf{+0.0140 (+1.40\%)} \\
 & Resize High      & 77.5\% & $0.7760 \pm 0.29$ & $0.0000$ & 22.4\% & $0.3437 \pm 0.38$ & \textbf{+0.0061 (+0.61\%)} \\
 & Resize Extreme   & 53.1\% & $0.6344 \pm 0.35$ & $0.0000$ & 46.9\% & $0.2229 \pm 0.33$ & \textbf{+0.0045 (+0.45\%)} \\
\midrule
\multirow{5}{*}{OCHuman} 
 & Baseline (Clean) & 72.7\% & $0.7145 \pm 0.37$ & $0.0000$ & 27.3\% & $0.4576 \pm 0.44$ & $-0.0032$ ($-0.32\%$) \\
 & Resize Low       & 72.7\% & $0.7070 \pm 0.37$ & $0.0000$ & 27.3\% & $0.4579 \pm 0.44$ & \textbf{+0.0069 (+0.69\%)} \\
 & Resize Medium    & 73.1\% & $0.6833 \pm 0.38$ & $0.0000$ & 26.9\% & $0.4425 \pm 0.43$ & \textbf{+0.0068 (+0.68\%)} \\
 & Resize High      & 71.8\% & $0.6163 \pm 0.39$ & $0.0000$ & 28.2\% & $0.3487 \pm 0.40$ & \textbf{+0.0019 (+0.19\%)} \\
 & Resize Extreme   & 59.3\% & $0.4417 \pm 0.39$ & $0.0000$ & 40.7\% & $0.2468 \pm 0.35$ & \textbf{+0.0007 (+0.07\%)} \\
\midrule
\multicolumn{2}{l|}{\textbf{Pooled Total (85,255 pts)}} & \textbf{74.7\%} & $\mathbf{0.7380 \pm 0.34}$ & $\mathbf{0.0000}$ & \textbf{25.3\%} & $\mathbf{0.3186 \pm 0.39}$ & $\mathbf{+0.0040 \ (+0.40\%)}$ \\
\bottomrule
\end{tabular}%
}
\end{table*}

\section{In-Distribution Uncertainty Calibration}
\label{sec:supp_calibration}
This section details the empirical evaluation of Area Under the Sparsification Error (AUSE) and Expected Calibration Error (ECE), stratifying by mixture cardinality ($K=1$ vs. $K=2$) across visual degradation levels. We show that standard heuristic confidences degrade under severe occlusion, whereas explicit geometric modeling provides a calibrated uncertainty signal.

\begin{table*}[htbp]
\centering
\resizebox{\textwidth}{!}{
\begin{tabular}{l|ccc|ccc|ccc}
\toprule
 & \multicolumn{3}{c|}{\textbf{All (Global)}} & \multicolumn{3}{c|}{\textbf{Visible}} & \multicolumn{3}{c}{\textbf{Occluded}} \\
\textbf{Dataset} & \textbf{DARK ($1-P_{DARK}$)} & \textbf{DARK (Entropy)} & \textbf{Ours ($\det(\boldsymbol{\Sigma}_{\text{final}})$)} & \textbf{DARK ($1-P_{DARK}$)} & \textbf{DARK (Entropy)} & \textbf{Ours ($\det(\boldsymbol{\Sigma}_{\text{final}})$)} & \textbf{DARK ($1-P_{DARK}$)} & \textbf{DARK (Entropy)} & \textbf{Ours ($\det(\boldsymbol{\Sigma}_{\text{final}})$)} \\
\midrule
\textbf{COCO} & \textbf{0.0581} & 0.1437 & 0.0945 & \textbf{0.0570} & 0.1404 & 0.0902 & \textbf{0.1026} & 0.2037 & 0.1322 \\
\textbf{CrowdPose} & \textbf{0.0515} & 0.1591 & 0.0911 & \textbf{0.0379} & 0.1262 & 0.0657 & \textbf{0.0599} & 0.1845 & 0.1067 \\
\textbf{OCHuman} & \textbf{0.1573} & 0.2373 & 0.2143 & \textbf{0.1491} & 0.2320 & 0.2002 & 0.2636 & 0.3130 & \textbf{0.2203} \\
\bottomrule
\end{tabular}
}
\caption{Global Sparsification Error (AUSE). Lower values indicate better calibration. While the baseline performs better in clean, isolated scenarios, the geometric GMM volume ($\det(\boldsymbol{\Sigma}_{\text{final}})$) achieves lower sparsification error under severe multi-person occlusions (e.g., 0.2203 vs. 0.2636 on OCHuman Occluded).}
\label{tab:ause_general_combined}
\end{table*}

\begin{table*}[htbp]
\centering
\resizebox{\textwidth}{!}{
\begin{tabular}{l|cc|cc|cc}
\toprule
 & \multicolumn{2}{c|}{\textbf{COCO}} & \multicolumn{2}{c|}{\textbf{CrowdPose}} & \multicolumn{2}{c}{\textbf{OCHuman}} \\
\textbf{Degradation Level} & \textbf{DARK ($1-P_{DARK}$)} & \textbf{Ours ($\det(\boldsymbol{\Sigma}_{\text{final}})$)} & \textbf{DARK ($1-P_{DARK}$)} & \textbf{Ours ($\det(\boldsymbol{\Sigma}_{\text{final}})$)} & \textbf{DARK ($1-P_{DARK}$)} & \textbf{Ours ($\det(\boldsymbol{\Sigma}_{\text{final}})$)} \\
\midrule
\textbf{00\_Baseline\_Clean} & \textbf{0.0527} & 0.0654 & \textbf{0.0422} & 0.0766 & \textbf{0.1381} & 0.2037 \\
\textbf{01\_Resize\_Low} & \textbf{0.0544} & 0.0701 & \textbf{0.0432} & 0.0766 & \textbf{0.1429} & 0.2069 \\
\textbf{02\_Resize\_Medium} & \textbf{0.0689} & 0.0899 & \textbf{0.0528} & 0.0861 & \textbf{0.1548} & 0.2105 \\
\textbf{03\_Resize\_High} & \textbf{0.0943} & 0.1266 & \textbf{0.0676} & 0.1054 & \textbf{0.1819} & 0.2229 \\
\textbf{04\_Resize\_Extreme} & \textbf{0.1155} & 0.1235 & \textbf{0.0987} & 0.1253 & 0.2165 & \textbf{0.2024} \\
\bottomrule
\end{tabular}
}
\caption{Global AUSE evolution under progressive visual degradation. As spatial resolution degrades and feature maps lose sharpness, baseline calibration deteriorates. Conversely, the GMM covariance volume reflects the increased spatial dispersion, remaining competitive under high degradation.}
\label{tab:ause_degradation}
\end{table*}

\begin{table*}[htbp]
\centering
\resizebox{\textwidth}{!}{
\begin{tabular}{l|ccc|ccc}
\toprule
 & \multicolumn{3}{c|}{\textbf{Unimodal ($K=1$)}} & \multicolumn{3}{c}{\textbf{Bimodal ($K=2$)}} \\
\textbf{Dataset} & \textbf{DARK ($1-P_{DARK}$)} & \textbf{Entropy} & \textbf{Ours ($\det(\boldsymbol{\Sigma}_{\text{final}})$)} & \textbf{DARK ($1-P_{DARK}$)} & \textbf{Entropy} & \textbf{Ours ($\det(\boldsymbol{\Sigma}_{\text{final}})$)} \\
\midrule
\textbf{COCO} & \textbf{0.0637} & 0.1493 & 0.0914 & \textbf{0.1364} & 0.1822 & 0.1479 \\
\textbf{CrowdPose} & \textbf{0.0507} & 0.1424 & 0.0912 & \textbf{0.1759} & 0.2429 & 0.1924 \\
\textbf{OCHuman} & \textbf{0.1588} & 0.2309 & 0.2174 & \textbf{0.2519} & 0.2995 & 0.2710 \\
\bottomrule
\end{tabular}
}
\caption{Global AUSE segregated by unimodal vs. bimodal predictions. Across all datasets and metrics, bimodal detections ($K=2$) yield a higher AUSE than unimodal ones ($K=1$), confirming that mixture bimodality serves as an indicator of heightened spatial ambiguity.}
\label{tab:ause_gaussians}
\end{table*}

\begin{figure*}[htbp]
    \centering
    \includegraphics[width=\textwidth]{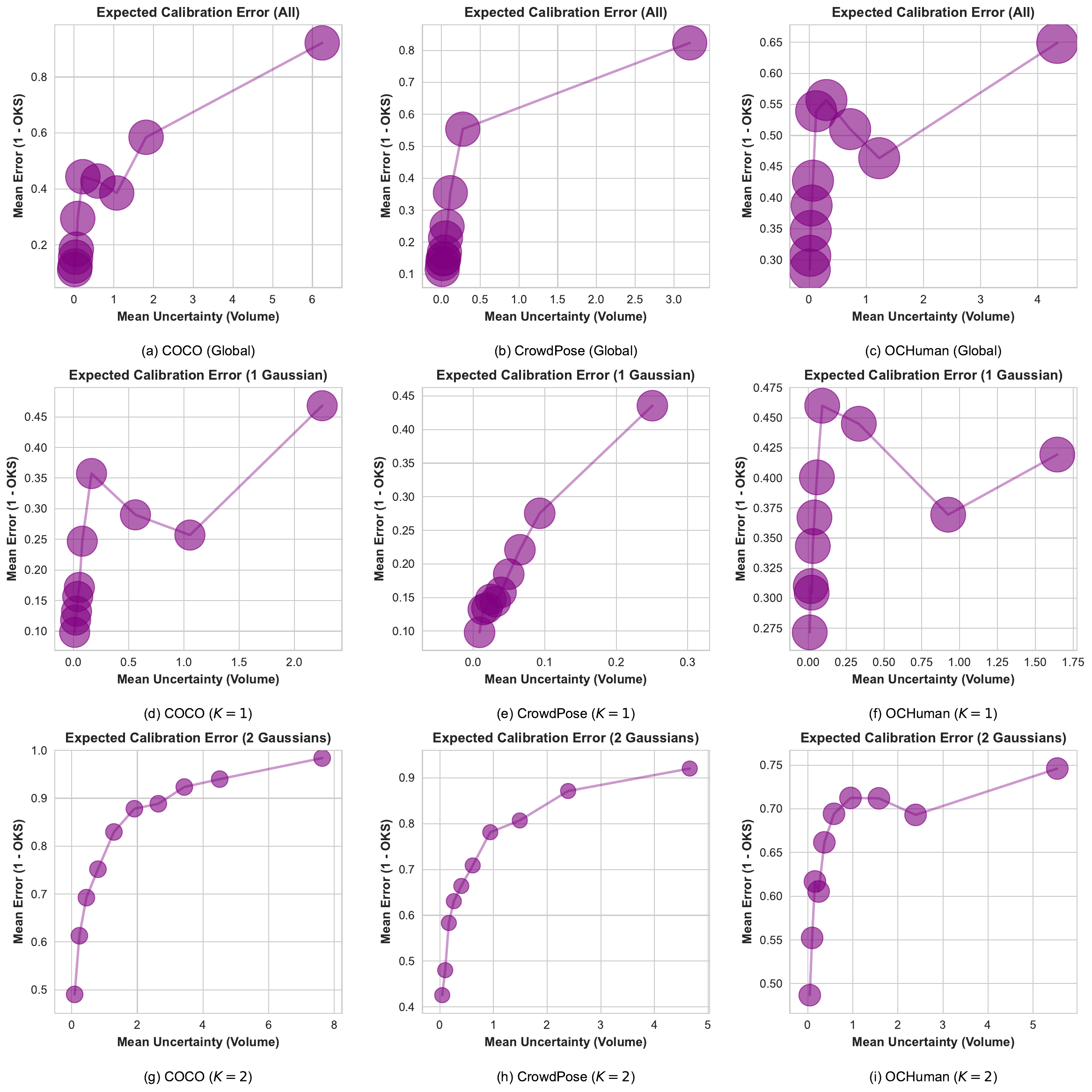}
    \caption{Continuous Expected Calibration Error (ECE) across distinct prediction regimes ($K=1$ vs. $K=2$). The x-axis represents log-volume $\log(1 + \det(\boldsymbol{\Sigma}_{\text{final}}))$ and the y-axis is the empirical mean error ($1 - \text{OKS}$). While the global evaluation (Top Row) mixes predictive states, decoupling the analysis clarifies the uncertainty response: the unimodal regime (Middle Row) clusters at low error values, whereas isolating bimodal predictions (Bottom Row) displays a monotonic relationship between covariance volume and error magnitude.}
    \label{fig:ece_combined_matrix}
\end{figure*}

\section{Adaptive Uncertainty Fusion}
\label{sec:supp_fusion}
This section provides the hyperparameter sensitivity analysis for the scaling parameter $\beta$ across Max-Pooling and Softmax fusion strategies. The results highlight the sensitivity of hard-routing mechanisms to threshold selection and evaluate the consistency of continuous Softmax fusion across degradation levels.

\begin{table*}[htbp]
\centering
\resizebox{\textwidth}{!}{
\begin{tabular}{l|ccc|ccc|ccc}
\toprule
 & \multicolumn{3}{c|}{\textbf{COCO}} & \multicolumn{3}{c|}{\textbf{CrowdPose}} & \multicolumn{3}{c}{\textbf{OCHuman}} \\
\textbf{Degradation} & \textbf{DARK (Base)} & \textbf{Max-Pool ($\beta$)} & \textbf{Softmax ($\beta$)} & \textbf{DARK (Base)} & \textbf{Max-Pool ($\beta$)} & \textbf{Softmax ($\beta$)} & \textbf{DARK (Base)} & \textbf{Max-Pool ($\beta$)} & \textbf{Softmax ($\beta$)} \\
\midrule
\textbf{00\_Baseline\_Clean} & 0.0528 & 0.0421 (0.19) & \textbf{0.0388} (0.12) & 0.0422 & 0.0418 (1.54) & \textbf{0.0413} (0.52) & 0.1382 & 0.1383 (0.01) & \textbf{0.1352} (0.04) \\
\textbf{01\_Resize\_Low} & 0.0545 & 0.0465 (0.22) & \textbf{0.0423} (0.10) & 0.0432 & 0.0432 (1.22) & \textbf{0.0425} (0.57) & 0.1430 & 0.1437 (0.00) & \textbf{0.1409} (0.03) \\
\textbf{02\_Resize\_Medium} & 0.0690 & 0.0613 (0.21) & \textbf{0.0554} (0.12) & 0.0528 & 0.0527 (0.90) & \textbf{0.0522} (0.66) & 0.1549 & 0.1551 (0.04) & \textbf{0.1517} (0.02) \\
\textbf{03\_Resize\_High} & 0.0945 & 0.0816 (0.31) & \textbf{0.0743} (0.19) & 0.0677 & 0.0678 (1.05) & \textbf{0.0671} (0.61) & 0.1820 & 0.1813 (0.16) & \textbf{0.1749} (0.08) \\
\textbf{04\_Resize\_Extreme} & 0.1162 & 0.1014 (0.45) & \textbf{0.0960} (0.31) & 0.0988 & 0.0983 (4.19) & \textbf{0.0948} (1.80) & 0.2216 & 0.1940 (0.77) & \textbf{0.1864} (7.18) \\
\bottomrule
\end{tabular}
}
\caption{Global AUSE comparison of fusion strategies under different degradation levels, accompanied by their respective oracle optimal $\beta$ values. Both Max-Pooling and Softmax consistently improve upon the baseline when $\beta$ is perfectly tuned.}
\label{tab:ause_fusion_deg_optimal}
\end{table*}

\begin{table*}[htbp]
    \centering
    \caption{Optimal empirical sensitivity hyperparameter ($\beta$) values determined via logarithmic Grid Search for the Max-Pooling fusion strategy. Clean scenarios require a small penalty (e.g., $0.0082$), while heavily occluded subsets require high sensitivity (e.g., $39.08$), demonstrating the difficulty of finding a single static operating point.}
    \label{tab:optimal_betas_combined}
    \resizebox{0.8\textwidth}{!}{
    \begin{tabular}{l|ccc|ccc}
        \toprule
        & \multicolumn{3}{c|}{\textbf{Global (All Points)}} & \multicolumn{3}{c}{\textbf{Occluded Joints Only}} \\
        \textbf{Degradation} & \textbf{COCO} & \textbf{CrowdPose} & \textbf{OCHuman} & \textbf{COCO} & \textbf{CrowdPose} & \textbf{OCHuman} \\
        \midrule
        \textbf{Clean} & 0.1924 & 1.5392 & 0.0082 & 0.1924 & 1.5392 & 39.0817 \\
        \textbf{Low} & 0.2245 & 1.2217 & 0.0000 & 0.2424 & 0.0707 & 45.5893 \\
        \textbf{Medium} & 0.2078 & 0.8978 & 0.0382 & 0.4157 & 0.0825 & 49.2388 \\
        \textbf{High} & 0.3055 & 1.0473 & 0.1650 & 0.1309 & 0.9697 & 45.5893 \\
        \textbf{Extreme} & 0.4489 & 4.1886 & 0.7696 & 0.2619 & 1.6624 & 36.1850 \\
        \bottomrule
    \end{tabular}
    }
\end{table*}

\begin{table*}[htbp]
\centering
\resizebox{0.8\textwidth}{!}{
\begin{tabular}{llccc}
\toprule
\textbf{Dataset} & \textbf{Degradation} & \textbf{DARK (Base) AUSE} & \textbf{Max-Pool AUSE ($\beta=0.1$)} & \textbf{Softmax AUSE ($\beta=0.1$)} \\
\midrule
\multirow{6}{*}{\textbf{COCO}} 
 & Clean & 0.0527 & 0.0435 & \textbf{0.0388} \\
 & Low & 0.0544 & 0.0477 & \textbf{0.0420} \\
 & Medium & 0.0689 & 0.0628 & \textbf{0.0555} \\
 & High & 0.0943 & 0.0862 & \textbf{0.0755} \\
 & Extreme & 0.1155 & 0.1098 & \textbf{0.0996} \\
 & \textit{General} & 0.0581 & 0.0524 & \textbf{0.0460} \\
\midrule
\multirow{6}{*}{\textbf{CrowdPose}} 
 & Clean & 0.0422 & 0.0423 & \textbf{0.0418} \\
 & Low & 0.0432 & 0.0434 & \textbf{0.0431} \\
 & Medium & 0.0528 & 0.0528 & \textbf{0.0525} \\
 & High & \textbf{0.0676} & 0.0680 & 0.0677 \\
 & Extreme & \textbf{0.0987} & 0.0999 & 0.0990 \\
 & \textit{General} & 0.0515 & 0.0518 & \textbf{0.0514} \\
\midrule
\multirow{6}{*}{\textbf{OCHuman}} 
 & Clean & \textbf{0.1381} & 0.1409 & 0.1390 \\
 & Low & \textbf{0.1429} & 0.1461 & 0.1435 \\
 & Medium & 0.1548 & 0.1557 & \textbf{0.1532} \\
 & High & 0.1819 & 0.1815 & \textbf{0.1756} \\
 & Extreme & 0.2165 & 0.2067 & \textbf{0.1945} \\
 & \textit{General} & 0.1573 & 0.1584 & \textbf{0.1540} \\
\bottomrule
\end{tabular}
}
\caption{Sparsification Error (AUSE) using a fixed $\beta=0.1$. Unlike Max-Pooling, the continuous Softmax strategy smoothly balances the contributions without per-image parameter tuning, matching or improving upon the DARK baseline across datasets and degradation levels.}
\label{tab:ause_fusion_final}
\end{table*}

\begin{figure*}[htbp]
    \centering
    \includegraphics[width=\textwidth]{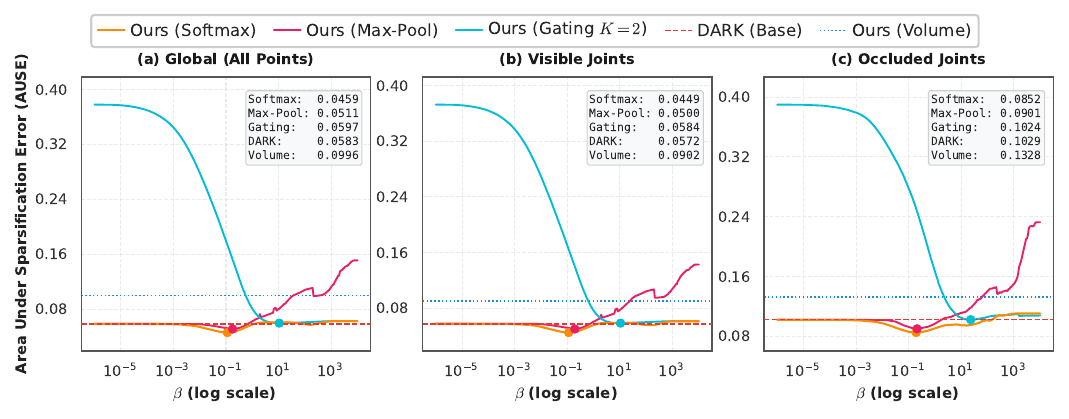}
    \caption{Grid Search optimization curves for the sensitivity hyperparameter $\beta$ evaluated on HRNet-W32 (COCO). Softmax fusion (orange curve) consistently achieves lower AUSE, with an empirical optimum near $\beta=0.1$.}
    \label{fig:beta_optimization}
\end{figure*}

\section{Out-of-Distribution (OoD) Anomaly Detection}
\label{sec:supp_ood}
This section provides extended Out-of-Distribution (OoD) detection results across visual degradation levels, evaluating the identification of absent or heavily occluded keypoints ($\text{vis} = 0$).

\begin{table*}[htbp]
    \centering
    \caption{Global AUROC for OoD Anomaly Detection (vis == 0).}
    \label{tab:ood_global_auc}
    \begin{tabular}{lcccc}
        \toprule
        \textbf{Dataset} & \textbf{Ours (Volume)} & \textbf{Ours (Uniform)} & \textbf{DARK (Entropy)} & \textbf{DARK ($1-P_{DARK}$)} \\
        \midrule
        COCO       & 0.7364 & 0.7387 & 0.7042 & \textbf{0.8122} \\
        CrowdPose  & \textbf{0.8787} & 0.5408 & 0.3256 & 0.5160 \\
        OCHuman    & \textbf{0.7209} & 0.5712 & 0.4643 & 0.6441 \\
        \bottomrule
    \end{tabular}
\end{table*}

\begin{table*}[htbp]
\centering
\caption{Evolution of AUROC for OoD detection across image degradation levels. As spatial resolution drops, deterministic baseline confidence degrades substantially. In dense datasets like CrowdPose, GMM covariance volume retains high discriminative performance ($0.8053$) under extreme degradation, whereas the baseline drops close to random classification ($0.5439$).}
\label{tab:ood_degradation_auc}
\begin{tabular}{llccccc}
\toprule
\textbf{Dataset} & \textbf{Metric} & \textbf{Clean} & \textbf{Low} & \textbf{Medium} & \textbf{High} & \textbf{Extreme} \\
\midrule
\multirow{2}{*}{COCO} & Ours (Volume) & 0.7663 & 0.7798 & 0.7832 & 0.7414 & 0.6753 \\
 & DARK ($1-P_{DARK}$) & \textbf{0.8974} & \textbf{0.9062} & \textbf{0.8861} & \textbf{0.8130} & \textbf{0.7053} \\
\midrule
\multirow{2}{*}{CrowdPose} & Ours (Volume) & \textbf{0.9231} & \textbf{0.9216} & \textbf{0.9139} & \textbf{0.8729} & \textbf{0.8053} \\
 & DARK ($1-P_{DARK}$) & 0.4858 & 0.4859 & 0.4928 & 0.5181 & 0.5439 \\
\midrule
\multirow{2}{*}{OCHuman} & Ours (Volume) & \textbf{0.7312} & \textbf{0.7295} & \textbf{0.7291} & \textbf{0.7268} & \textbf{0.6926} \\
 & DARK ($1-P_{DARK}$) & 0.6511 & 0.6496 & 0.6524 & 0.6529 & 0.6265 \\
\bottomrule
\end{tabular}
\end{table*}

\begin{figure*}[htbp]
    \centering
    \includegraphics[width=\textwidth]{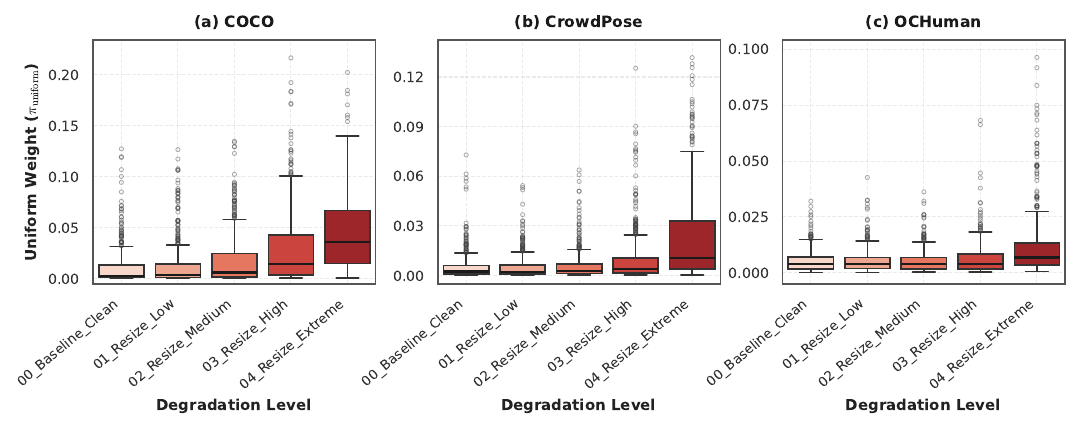}
    \caption{Distribution of the uniform weight ($\pi_{\text{uniform}}$) as visual degradation escalates. Boxplots demonstrate the metric's responsiveness to extreme noise.}
    \label{fig:uniform_weight_all}
\end{figure*}

\begin{figure*}[htbp]
    \centering
    \includegraphics[width=\textwidth]{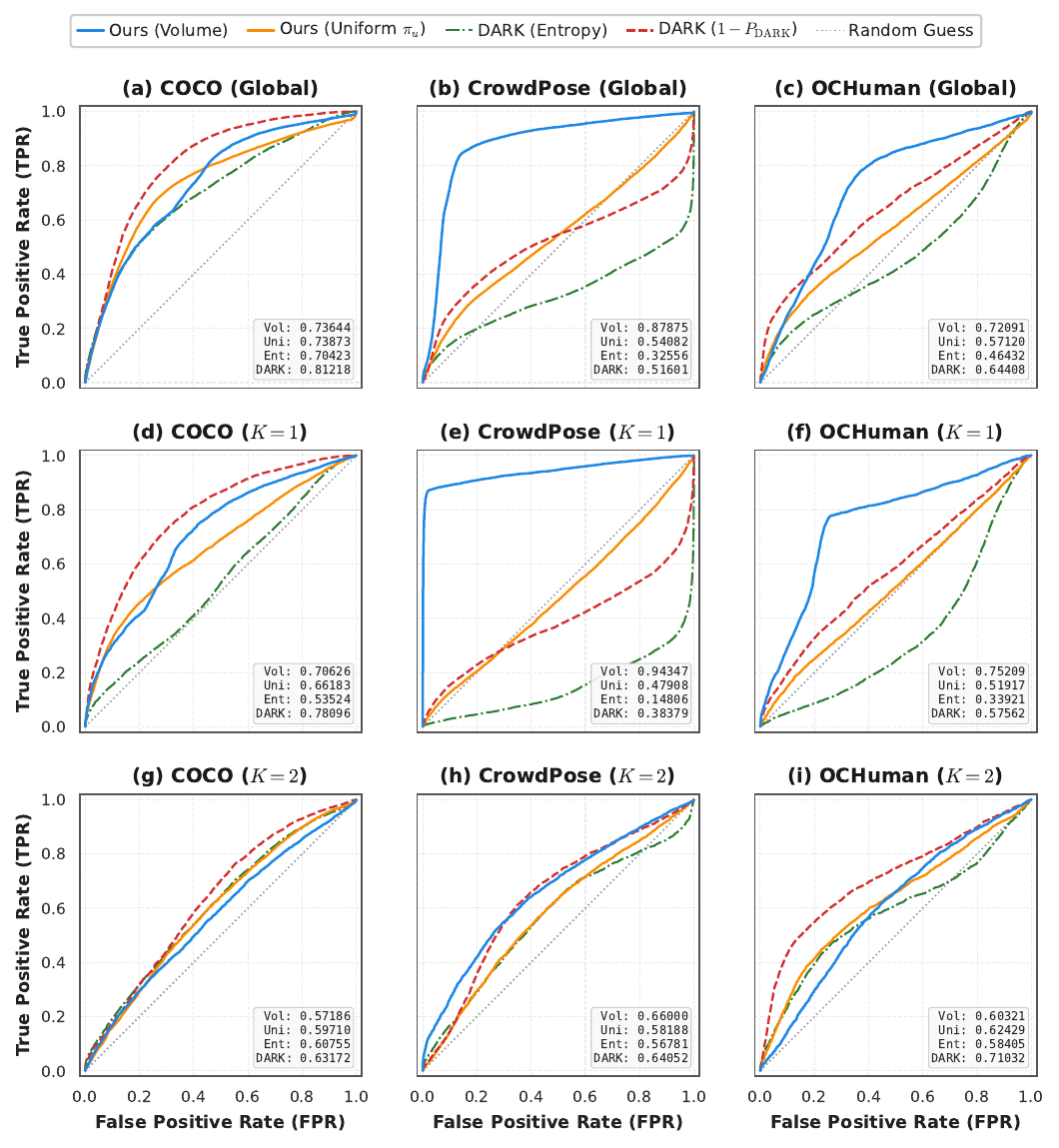}
    \caption{Out-of-Distribution (OoD) Anomaly Detection ROC curves decoupled by mixture cardinality ($K=1$ vs. $K=2$). In dense environments (CrowdPose, OCHuman) with unimodal predictions (Middle Row, $K=1$), baseline peak confidence degrades to $0.38$ AUROC due to confident detections on adjacent individuals. Conversely, continuous GMM covariance volume maintains strong discriminative ability ($0.94$ AUROC on CrowdPose) by evaluating spatial dispersion rather than peak intensity.}
    \label{fig:ood_combined_matrix}
\end{figure*}

\begin{figure*}[htbp]
    \centering
    \begin{subfigure}[b]{0.7\textwidth}
        \centering
        \includegraphics[width=\textwidth]{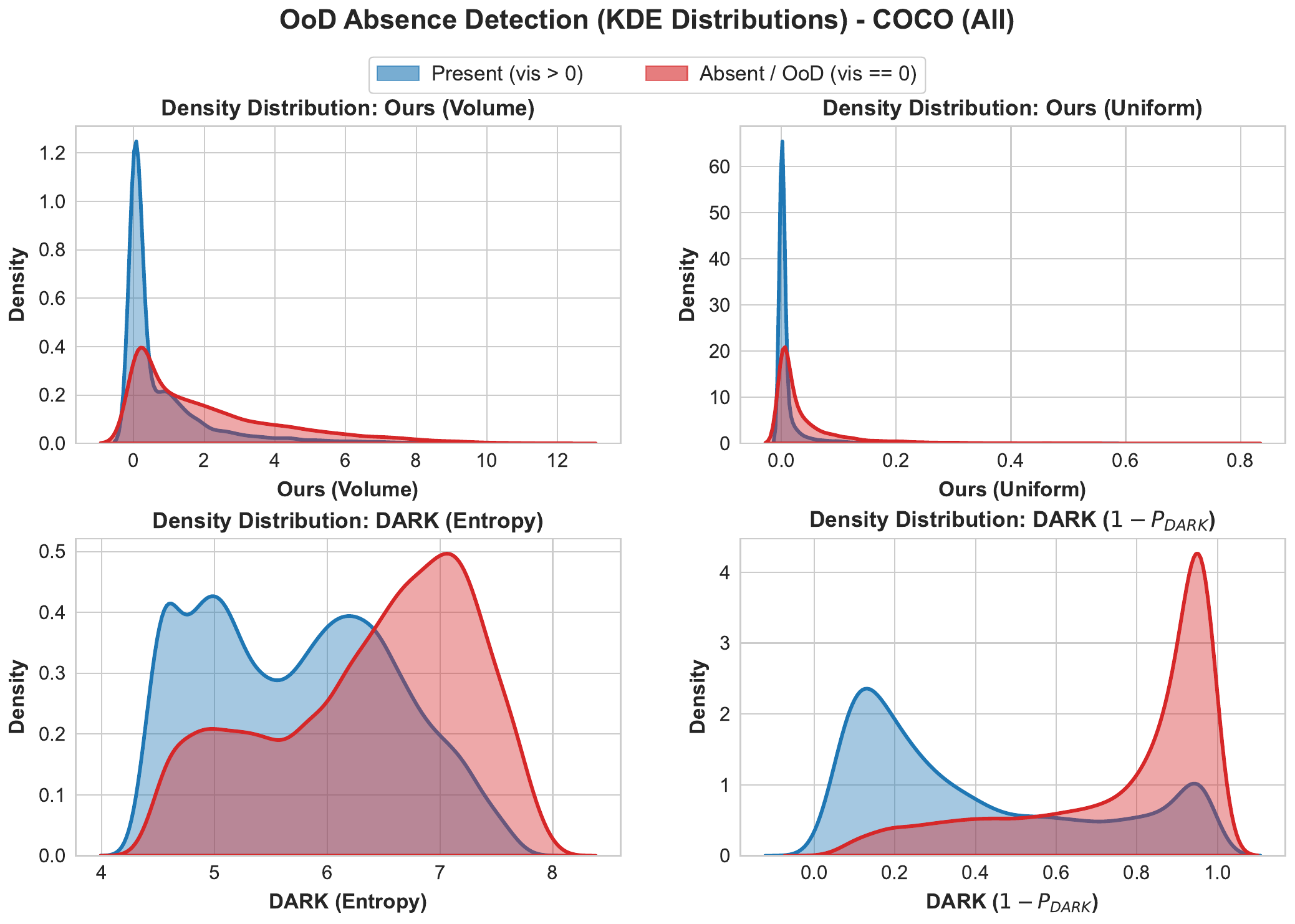}
        \caption{COCO (KDE)}
        \label{fig:ood_kde_coco}
    \end{subfigure}
    \vspace{4mm} 
    
    \begin{subfigure}[b]{0.7\textwidth}
        \centering
        \includegraphics[width=\textwidth]{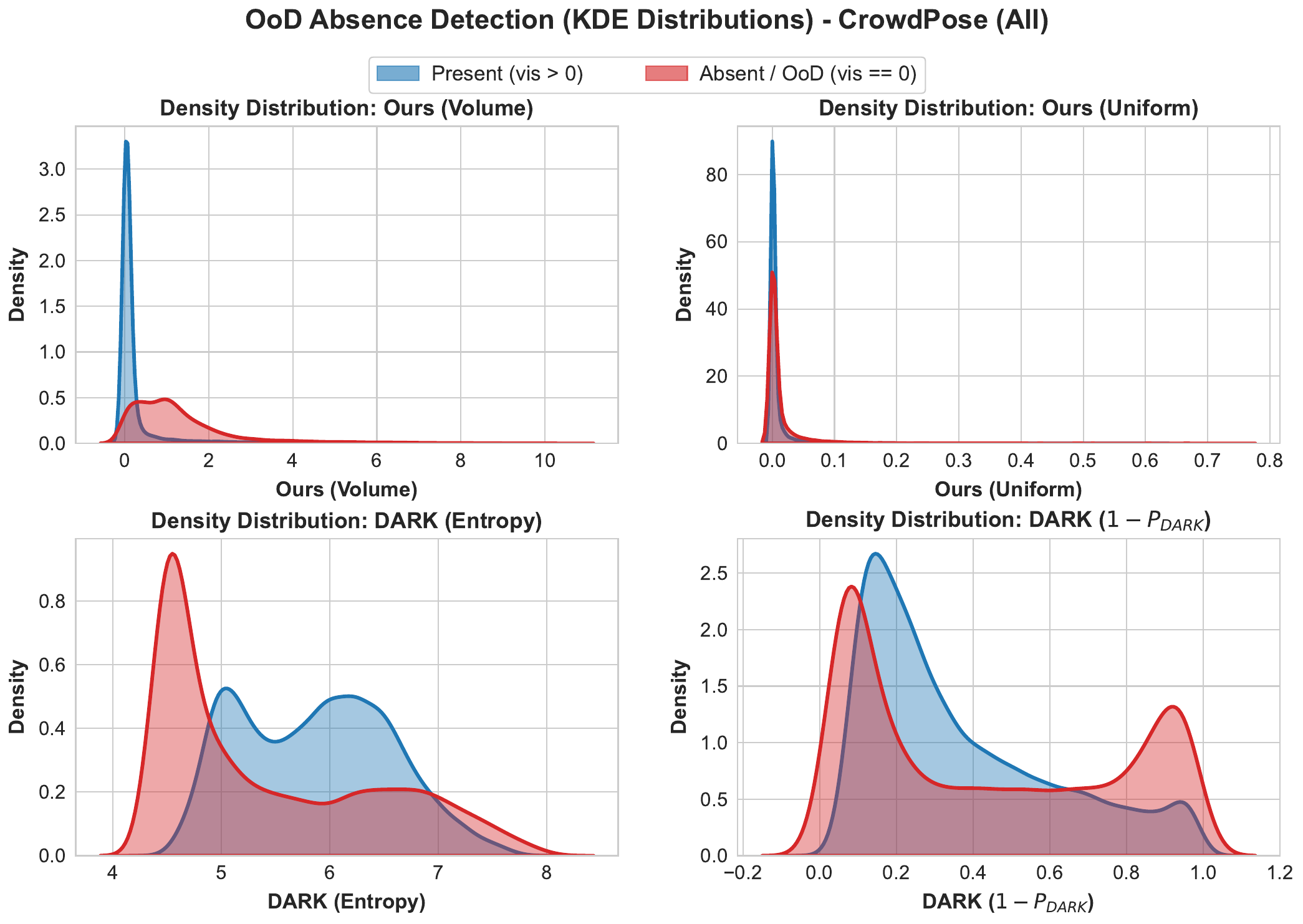}
        \caption{CrowdPose (KDE)}
        \label{fig:ood_kde_crowdpose}
    \end{subfigure}
    \vspace{4mm} 
    
    \begin{subfigure}[b]{0.7\textwidth}
        \centering
        \includegraphics[width=\textwidth]{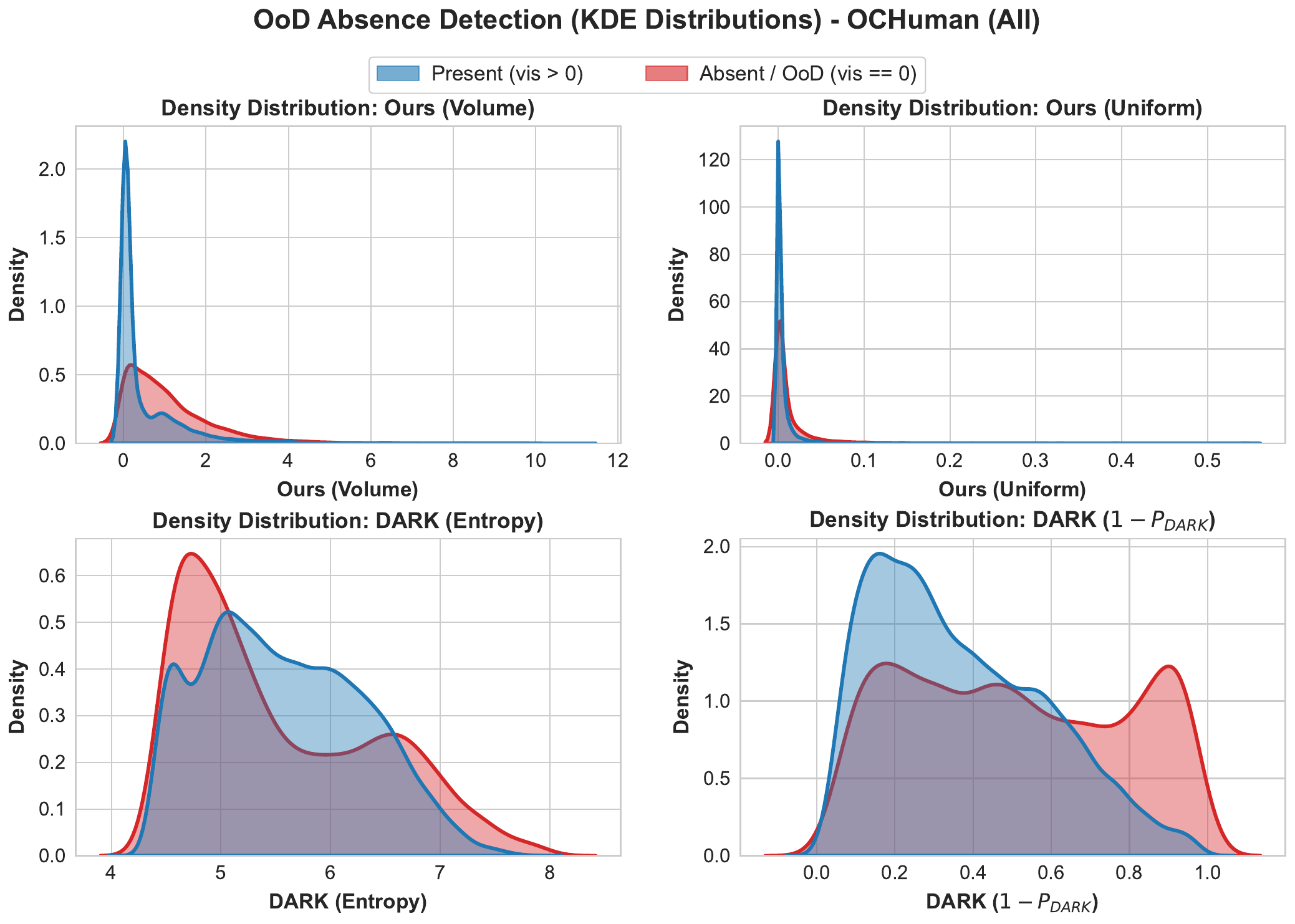}
        \caption{OCHuman (KDE)}
        \label{fig:ood_kde_ochuman}
    \end{subfigure}
    
    \caption{KDE density estimations comparing the distribution of In-Distribution versus OoD points. The bimodal separation achieved by covariance volume (blue curve on the right) accounts for its higher AUROC performance.}
    \label{fig:ood_kde_grid_all}
\end{figure*}

\begin{figure*}[htbp]
    \centering
    \includegraphics[width=\textwidth]{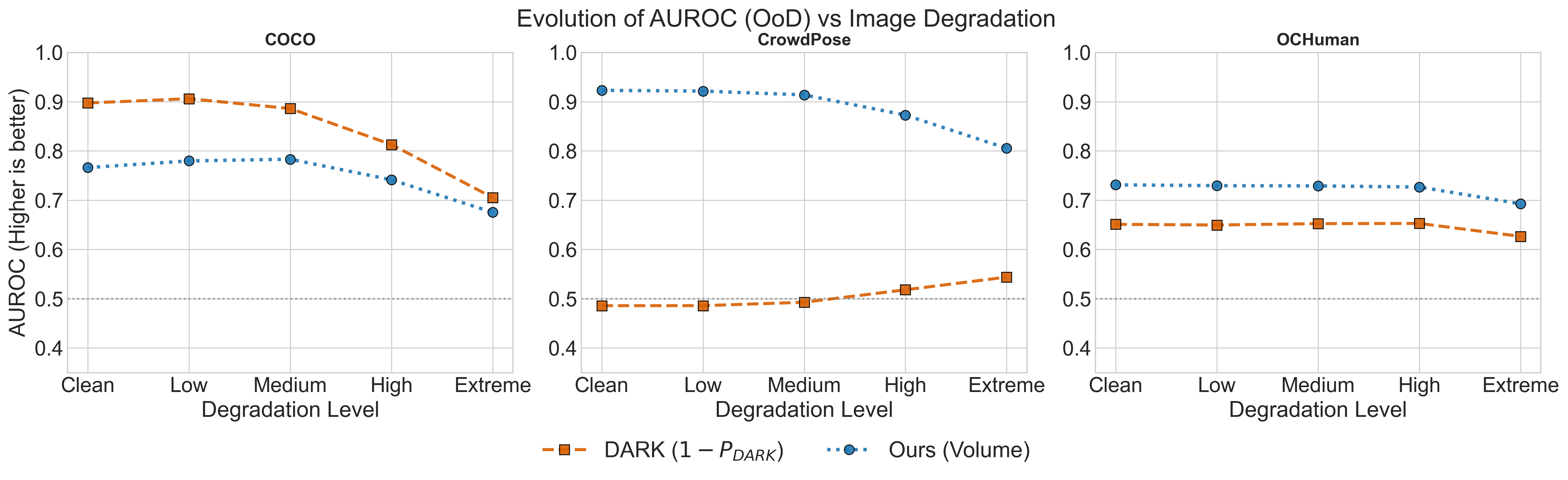}
    \caption{Impact of image degradation (Clean to Extreme) on OoD anomaly detection capabilities (AUROC).}
    \label{fig:ood_degradation_evolution}
\end{figure*}

\subsection{Convolutional Baseline: ResNet50}
We also evaluated the framework using a ResNet-50 backbone configured with the DARK sub-pixel decoder. The empirical results show consistent trends with the findings observed on HRNet-W32.

\subsubsection{Limb Swap Analysis and Stochastic Aggregation}
Table \ref{tab:resnet_supp_tta_comprehensive} reflects an architectural trade-off consistent with the HRNet-W32 experiments. Continuous GMM aggregation acts as an effective spatial filter for minor ambiguities, reducing Loose Swaps across degraded environments (e.g., 24 fewer loose swaps in OCHuman High and 22 in COCO Extreme).

However, continuous density fusion is more susceptible to left-right symmetric confusions (Strict Swaps) than discrete thresholding when visual features are less distinct. On a lighter convolutional backbone like ResNet-50, discrete TTA maintains lower Strict Swap rates under extreme degradation, whereas GMM aggregation introduces additional catastrophic swap errors (e.g., adding +13 Strict Swaps in OCHuman Extreme compared to DARK).

Table \ref{tab:resnet_supp_mrf_comprehensive} reports the impact of the MRF structural prior. In line with the HRNet evaluation, the MRF prior regularizes coordinates and improves overall precision (OKS). However, enforcing pairwise kinematic constraints over degraded heatmaps can propagate spatial errors, increasing Strict Swaps (e.g., adding +8 Strict Swaps in both OCHuman Extreme and COCO Extreme).

\begin{table*}[htbp]
\centering
\small
\resizebox{\textwidth}{!}{%
\setlength{\tabcolsep}{4.5pt} 
\begin{tabular}{@{}llccc|ccc|ccc@{}}
\toprule
\multirow{2}{*}{\textbf{Dataset}} & \multirow{2}{*}{\textbf{Degradation}} & \multicolumn{3}{c|}{\textbf{OKS}} & \multicolumn{3}{c|}{\textbf{Loose Swaps}} & \multicolumn{3}{c}{\textbf{Strict Swaps}} \\ \cmidrule(l){3-11}
 & & \textbf{DARK} & \textbf{GMM} & \textbf{$\Delta$} & \textbf{DARK} & \textbf{GMM} & \textbf{$\Delta$} & \textbf{DARK} & \textbf{GMM} & \textbf{$\Delta$} \\ \midrule

\multirow{5}{*}{COCO}
 & Baseline (Clean) & 0.6868 & \textbf{0.6870} & \textcolor{teal}{+0.0002} & \textbf{119} & 126 & \textcolor{red}{+7} & \textbf{22} & \textbf{22} & 0 \\
 & Resize Low & 0.6603 & \textbf{0.6607} & \textcolor{teal}{+0.0005} & \textbf{140} & 145 & \textcolor{red}{+5} & \textbf{32} & \textbf{32} & 0 \\
 & Resize Medium & \textbf{0.5767} & 0.5764 & \textcolor{red}{-0.0003} & \textbf{225} & 232 & \textcolor{red}{+7} & \textbf{77} & \textbf{77} & 0 \\
 & Resize High & \textbf{0.4140} & 0.4132 & \textcolor{red}{-0.0008} & 365 & \textbf{353} & \textcolor{teal}{-12} & \textbf{108} & 115 & \textcolor{red}{+7} \\
 & Resize Extreme & \textbf{0.2270} & 0.2267 & \textcolor{red}{-0.0003} & 381 & \textbf{359} & \textcolor{teal}{-22} & \textbf{108} & 115 & \textcolor{red}{+7} \\ \midrule
\multirow{5}{*}{CrowdPose}
 & Baseline (Clean) & 0.8172 & \textbf{0.8175} & \textcolor{teal}{+0.0003} & 169 & \textbf{168} & \textcolor{teal}{-1} & 41 & \textbf{37} & \textcolor{teal}{-4} \\
 & Resize Low & \textbf{0.8074} & 0.8065 & \textcolor{red}{-0.0009} & 176 & \textbf{175} & \textcolor{teal}{-1} & \textbf{46} & 47 & \textcolor{red}{+1} \\
 & Resize Medium & 0.7677 & \textbf{0.7686} & \textcolor{teal}{+0.0009} & 218 & \textbf{210} & \textcolor{teal}{-8} & 60 & \textbf{55} & \textcolor{teal}{-5} \\
 & Resize High & 0.6361 & \textbf{0.6367} & \textcolor{teal}{+0.0005} & 350 & \textbf{345} & \textcolor{teal}{-5} & \textbf{105} & \textbf{105} & 0 \\
 & Resize Extreme & \textbf{0.4152} & 0.4142 & \textcolor{red}{-0.0010} & \textbf{477} & \textbf{477} & 0 & \textbf{145} & 148 & \textcolor{red}{+3} \\ \midrule
\multirow{5}{*}{OCHuman}
 & Baseline (Clean) & \textbf{0.5999} & 0.5996 & \textcolor{red}{-0.0003} & 463 & \textbf{458} & \textcolor{teal}{-5} & \textbf{116} & 119 & \textcolor{red}{+3} \\
 & Resize Low & 0.5953 & \textbf{0.5967} & \textcolor{teal}{+0.0014} & 469 & \textbf{463} & \textcolor{teal}{-6} & 121 & \textbf{119} & \textcolor{teal}{-2} \\
 & Resize Medium & \textbf{0.5756} & 0.5755 & \textcolor{red}{-0.0001} & 487 & \textbf{482} & \textcolor{teal}{-5} & \textbf{127} & 128 & \textcolor{red}{+1} \\
 & Resize High & 0.5029 & \textbf{0.5041} & \textcolor{teal}{+0.0012} & 585 & \textbf{561} & \textcolor{teal}{-24} & 166 & \textbf{160} & \textcolor{teal}{-6} \\
 & Resize Extreme & \textbf{0.3399} & 0.3399 & \textcolor{red}{-0.0000} & 674 & \textbf{669} & \textcolor{teal}{-5} & \textbf{183} & 196 & \textcolor{red}{+13} \\ \bottomrule

\end{tabular}%
}
\vspace{2mm}
\caption{ResNet50 Ablation on Stochastic Aggregation: Comprehensive comparison of DARK (Discrete TTA) vs our continuous GMM aggregation across three datasets and five degradation levels (\textbf{DARK}: DARK (Discrete TTA), \textbf{GMM}: Ours (GMM).}
\label{tab:resnet_supp_tta_comprehensive}
\end{table*}

\begin{table*}[htbp]
\centering
\small
\resizebox{\textwidth}{!}{%
\setlength{\tabcolsep}{4.5pt} 
\begin{tabular}{@{}llcccc|cccc|cccc@{}}
\toprule
\multirow{2}{*}{\textbf{Dataset}} & \multirow{2}{*}{\textbf{Degradation}} & \multicolumn{4}{c|}{\textbf{OKS}} & \multicolumn{4}{c|}{\textbf{Loose Swaps}} & \multicolumn{4}{c}{\textbf{Strict Swaps}} \\ \cmidrule(l){3-14}
 & & \textbf{DARK} & \textbf{GMM} & \textbf{+MRF} & \textbf{$\Delta_{M,G}$} & \textbf{DARK} & \textbf{GMM} & \textbf{+MRF} & \textbf{$\Delta_{M,G}$} & \textbf{DARK} & \textbf{GMM} & \textbf{+MRF} & \textbf{$\Delta_{M,G}$} \\ \midrule
\multirow{5}{*}{COCO}
 & Baseline (Clean) & 0.6868 & 0.6870 & \textbf{0.6875} & \textcolor{teal}{+0.0005} & \textbf{119} & 126 & 122 & \textcolor{teal}{-4} & \textbf{22} & \textbf{22} & 23 & \textcolor{red}{+1} \\
 & Resize Low & 0.6603 & 0.6607 & \textbf{0.6615} & \textcolor{teal}{+0.0008} & \textbf{140} & 145 & 146 & \textcolor{red}{+1} & \textbf{32} & \textbf{32} & 34 & \textcolor{red}{+2} \\
 & Resize Medium & \textbf{0.5767} & 0.5764 & 0.5765 & \textcolor{teal}{+0.0001} & \textbf{225} & 232 & 231 & \textcolor{teal}{-1} & \textbf{77} & \textbf{77} & 81 & \textcolor{red}{+4} \\
 & Resize High & \textbf{0.4140} & 0.4132 & 0.4139 & \textcolor{teal}{+0.0007} & 365 & \textbf{353} & 359 & \textcolor{red}{+6} & \textbf{108} & 115 & 115 & 0 \\
 & Resize Extreme & 0.2270 & 0.2267 & \textbf{0.2277} & \textcolor{teal}{+0.0010} & 381 & \textbf{359} & 381 & \textcolor{red}{+22} & \textbf{108} & 115 & 123 & \textcolor{red}{+8} \\ \midrule
\multirow{5}{*}{CrowdPose}
 & Baseline (Clean) & 0.8172 & 0.8175 & \textbf{0.8183} & \textcolor{teal}{+0.0007} & 169 & 168 & \textbf{166} & \textcolor{teal}{-2} & 41 & \textbf{37} & 38 & \textcolor{red}{+1} \\
 & Resize Low & 0.8074 & 0.8065 & \textbf{0.8079} & \textcolor{teal}{+0.0014} & 176 & 175 & \textbf{169} & \textcolor{teal}{-6} & 46 & 47 & \textbf{45} & \textcolor{teal}{-2} \\
 & Resize Medium & 0.7677 & 0.7686 & \textbf{0.7688} & \textcolor{teal}{+0.0002} & 218 & \textbf{210} & 213 & \textcolor{red}{+3} & 60 & \textbf{55} & 57 & \textcolor{red}{+2} \\
 & Resize High & 0.6361 & \textbf{0.6367} & 0.6365 & \textcolor{red}{-0.0002} & 350 & \textbf{345} & \textbf{345} & 0 & \textbf{105} & \textbf{105} & 108 & \textcolor{red}{+3} \\
 & Resize Extreme & \textbf{0.4152} & 0.4142 & 0.4144 & \textcolor{teal}{+0.0002} & \textbf{477} & \textbf{477} & 488 & \textcolor{red}{+11} & \textbf{145} & 148 & 156 & \textcolor{red}{+8} \\ \midrule
\multirow{5}{*}{OCHuman}
 & Baseline (Clean) & \textbf{0.5999} & 0.5996 & 0.5996 & \textcolor{red}{-0.0000} & 463 & 458 & \textbf{456} & \textcolor{teal}{-2} & \textbf{116} & 119 & 120 & \textcolor{red}{+1} \\
 & Resize Low & 0.5953 & 0.5967 & \textbf{0.5976} & \textcolor{teal}{+0.0009} & 469 & \textbf{463} & 467 & \textcolor{red}{+4} & 121 & \textbf{119} & 125 & \textcolor{red}{+6} \\
 & Resize Medium & 0.5756 & 0.5755 & \textbf{0.5776} & \textcolor{teal}{+0.0021} & 487 & 482 & \textbf{477} & \textcolor{teal}{-5} & \textbf{127} & 128 & 130 & \textcolor{red}{+2} \\
 & Resize High & 0.5029 & 0.5041 & \textbf{0.5042} & \textcolor{teal}{+0.0000} & 585 & \textbf{561} & 571 & \textcolor{red}{+10} & 166 & \textbf{160} & 166 & \textcolor{red}{+6} \\
 & Resize Extreme & \textbf{0.3399} & 0.3399 & 0.3395 & \textcolor{red}{-0.0004} & 674 & \textbf{669} & \textbf{669} & 0 & \textbf{183} & 196 & 204 & \textcolor{red}{+8} \\ \bottomrule

\end{tabular}%
}
\vspace{2mm}
\caption{ResNet50 Ablation on Global Structural Priors: Isolating the effect of the MRF structural constraint applied over the continuous GMM + TTA probability distributions. (\textbf{DARK}: DARK (Discrete TTA), \textbf{GMM}: Ours (GMM), \textbf{+MRF}: Ours (+MRF)).}
\label{tab:resnet_supp_mrf_comprehensive}
\end{table*}

\subsubsection{Uncertainty Calibration and OoD Detection}
Tables \ref{tab:resnet_ause_fusion_deg_optimal} and \ref{tab:resnet_ause_fusion_final} show that Adaptive Softmax Fusion balances baseline confidence with GMM uncertainty, reducing the Sparsification Error (AUSE) without requiring image-specific hyperparameter search.

\begin{table}[htbp]
\centering
\resizebox{\textwidth}{!}{
\begin{tabular}{l|ccc|ccc|ccc}
\toprule
 & \multicolumn{3}{c|}{\textbf{COCO}} & \multicolumn{3}{c|}{\textbf{CrowdPose}} & \multicolumn{3}{c}{\textbf{OCHuman}} \\
\textbf{Degradation} & \textbf{DARK (Base)} & \textbf{Max-Pool ($\beta$)} & \textbf{Softmax ($\beta$)} & \textbf{DARK (Base)} & \textbf{Max-Pool ($\beta$)} & \textbf{Softmax ($\beta$)} & \textbf{DARK (Base)} & \textbf{Max-Pool ($\beta$)} & \textbf{Softmax ($\beta$)} \\
\midrule
\textbf{00\_Baseline\_Clean} & 0.0641 & 0.0458 (0.28) & \textbf{0.0416} (0.15) & 0.0429 & 0.0422 (1.94) & \textbf{0.0408} (1.22) & 0.1598 & 0.1596 (0.06) & \textbf{0.1547} (0.05) \\
\textbf{01\_Resize\_Low} & 0.0677 & 0.0507 (0.28) & \textbf{0.0456} (0.15) & 0.0459 & 0.0448 (2.09) & \textbf{0.0430} (1.43) & 0.1646 & 0.1618 (0.08) & \textbf{0.1571} (0.05) \\
\textbf{02\_Resize\_Medium} & 0.0782 & 0.0668 (0.26) & \textbf{0.0606} (0.14) & 0.0498 & 0.0494 (2.09) & \textbf{0.0474} (1.32) & 0.1677 & 0.1663 (0.12) & \textbf{0.1608} (0.06) \\
\textbf{03\_Resize\_High} & 0.1053 & 0.0887 (0.33) & \textbf{0.0821} (0.19) & 0.0672 & 0.0676 (1.94) & \textbf{0.0660} (0.71) & 0.1873 & 0.1823 (0.22) & \textbf{0.1764} (0.10) \\
\textbf{04\_Resize\_Extreme} & 0.1168 & 0.0962 (0.71) & \textbf{0.0913} (0.38) & 0.0971 & 0.0965 (0.83) & \textbf{0.0933} (2.09) & 0.2172 & 0.1955 (9.77) & \textbf{0.1865} (13.30) \\
\bottomrule
\end{tabular}
}
\caption{ResNet50 Global AUSE comparison of fusion strategies accompanied by their oracle optimal $\beta$ values.}
\label{tab:resnet_ause_fusion_deg_optimal}
\end{table}

\begin{table*}[htbp]
\centering
\resizebox{0.8\textwidth}{!}{
\begin{tabular}{llccc}
\toprule
\textbf{Dataset} & \textbf{Degradation} & \textbf{DARK (Base) AUSE} & \textbf{Max-Pool AUSE ($\beta=0.1$)} & \textbf{Softmax AUSE ($\beta=0.1$)} \\
\midrule
\multirow{6}{*}{\textbf{COCO}}
 & Clean & 0.0639 & 0.0489 & \textbf{0.0420} \\
 & Low & 0.0673 & 0.0542 & \textbf{0.0460} \\
 & Medium & 0.0782 & 0.0659 & \textbf{0.0563} \\
 & High & 0.1002 & 0.0923 & \textbf{0.0793} \\
 & Extreme & 0.1167 & 0.1107 & \textbf{0.0969} \\
 & \textit{General} & 0.0662 & 0.0567 & \textbf{0.0484} \\
\midrule
\multirow{6}{*}{\textbf{CrowdPose}}
 & Clean & 0.0428 & 0.0432 & \textbf{0.0426} \\
 & Low & 0.0459 & 0.0460 & \textbf{0.0452} \\
 & Medium & 0.0497 & 0.0501 & \textbf{0.0494} \\
 & High & \textbf{0.0671} & 0.0677 & \textbf{0.0671} \\
 & Extreme & 0.0967 & 0.0965 & \textbf{0.0955} \\
 & \textit{General} & 0.0519 & 0.0521 & \textbf{0.0515} \\
\midrule
\multirow{6}{*}{\textbf{OCHuman}}
 & Clean & 0.1597 & 0.1601 & \textbf{0.1566} \\
 & Low & 0.1642 & 0.1621 & \textbf{0.1584} \\
 & Medium & 0.1676 & 0.1669 & \textbf{0.1618} \\
 & High & 0.1868 & 0.1855 & \textbf{0.1765} \\
 & Extreme & 0.2157 & 0.2059 & \textbf{0.1937} \\
 & \textit{General} & 0.1694 & 0.1686 & \textbf{0.1624} \\
\bottomrule
\end{tabular}
}
\caption{ResNet50 Sparsification Error (AUSE) using a static $\beta=0.1$. The continuous Softmax strategy consistently outperforms or matches the DARK baseline.}
\label{tab:resnet_ause_fusion_final}
\end{table*}

\begin{figure*}[htbp]
    \centering
    \includegraphics[width=\textwidth]{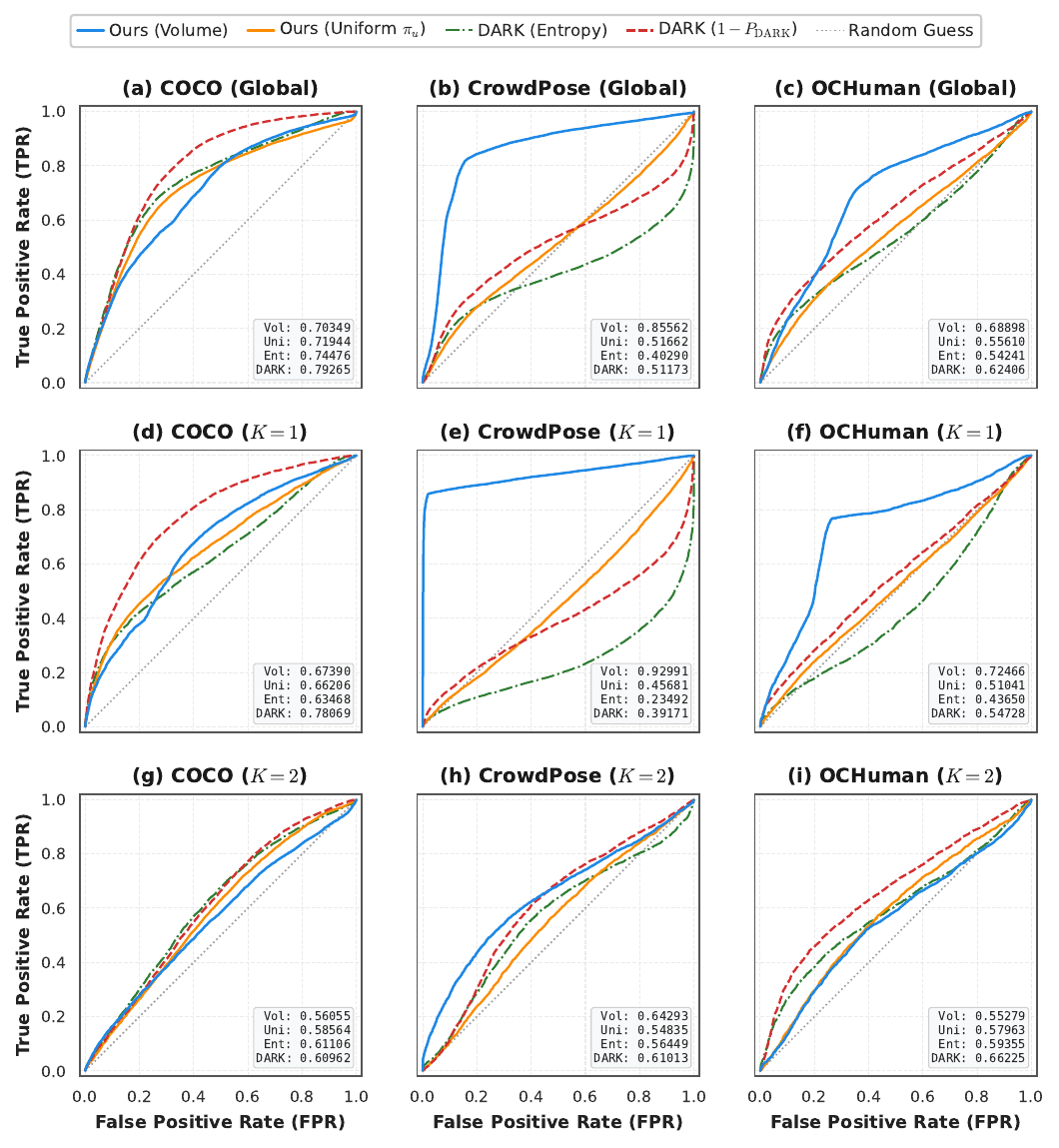}
    \caption{ResNet50 Out-of-Distribution (OoD) Anomaly Detection ROC curves decoupled by mixture cardinality ($K=1$ vs. $K=2$).}
    \label{fig:resnet_ood_combined_matrix}
\end{figure*}

\begin{figure*}[htbp]
    \centering
    \includegraphics[width=\textwidth]{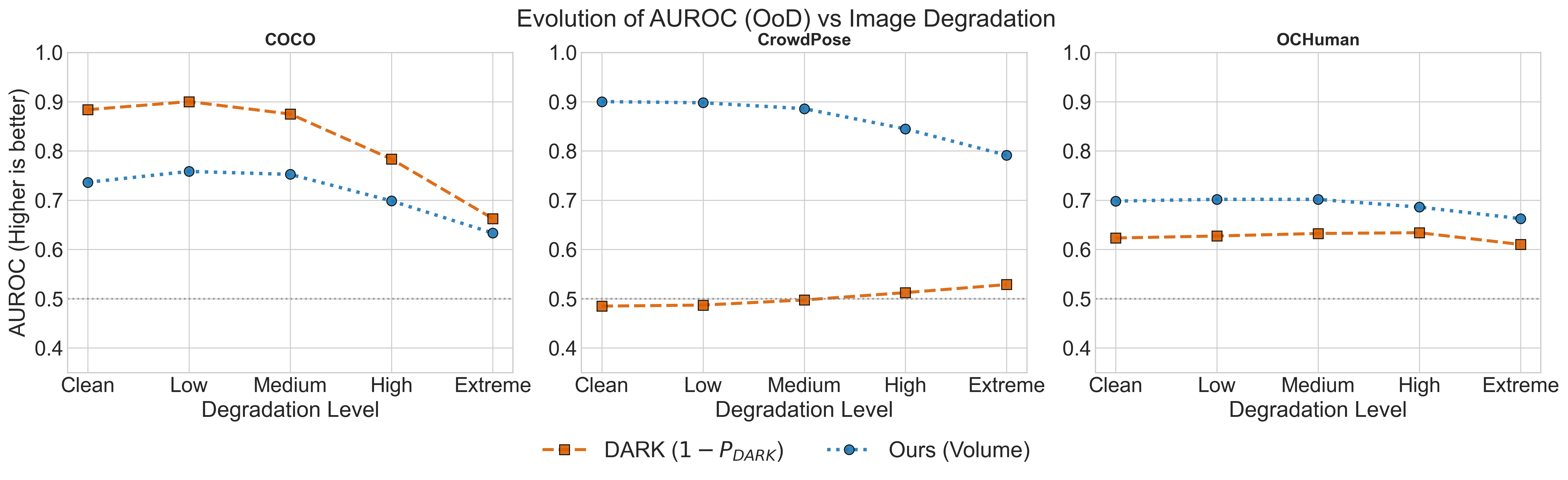}
    \caption{ResNet50: Impact of image degradation (Clean to Extreme) on OoD anomaly detection capabilities (AUROC).}
    \label{fig:resnet_ood_degradation_evolution}
\end{figure*}

Regarding Out-of-Distribution detection, Figures \ref{fig:resnet_ood_combined_matrix} and \ref{fig:resnet_ood_degradation_evolution} demonstrate that the explicit geometric covariance \textbf{Ours (Volume)} consistently outperforms the heuristic \textbf{DARK ($1 - P_{DARK}$)} in identifying absent or occluded keypoints. This confirms that the OoD detection advantage stems directly from evaluating spatial dispersion rather than peak heatmap intensity.

\subsection{Vision Transformer: ViTPose-Small}
To evaluate the methodology on non-convolutional architectures, we integrated the framework with ViTPose-Small. This architecture uses the UDP codec rather than DARK, testing whether the framework generalizes beyond DARK-based coordinate representations.

\subsubsection{Limb Swap Analysis and Stochastic Aggregation}
Because self-attention mechanisms process spatial relationships globally, vision transformers can produce activation profiles with different spatial noise characteristics than localized convolutions under heavy occlusion. Table \ref{tab:vitpose_supp_tta_comprehensive} details the stochastic aggregation performance on ViTPose-Small.

Continuous GMM aggregation consistently filters minor localization ambiguities, reducing Loose Swaps (e.g., 13 fewer Loose Swaps in COCO High and 11 in OCHuman Medium). Furthermore, unlike on the lighter ResNet-50 backbone, continuous aggregation over ViTPose heatmaps reduces catastrophic anatomical inversions, eliminating 9 Strict Swaps in COCO Extreme and 7 in OCHuman High compared to the \textbf{UDP (Discrete TTA)} baseline.

Table \ref{tab:vitpose_supp_mrf_comprehensive} shows the effect of the MRF structural prior when combined with UDP decoding. While the MRF improves absolute coordinate precision ($\Delta_{M,G}$ OKS is positive across most scenarios), applying 2D kinematic constraints over degraded attention maps can propagate structural errors, increasing Strict Swaps (+6 in COCO Extreme and +3 in OCHuman High).

\begin{table*}[htbp]
\centering
\small
\resizebox{\textwidth}{!}{%
\setlength{\tabcolsep}{4.5pt} 
\begin{tabular}{@{}llccc|ccc|ccc@{}}
\toprule
\multirow{2}{*}{\textbf{Dataset}} & \multirow{2}{*}{\textbf{Degradation}} & \multicolumn{3}{c|}{\textbf{OKS}} & \multicolumn{3}{c|}{\textbf{Loose Swaps}} & \multicolumn{3}{c}{\textbf{Strict Swaps}} \\ \cmidrule(l){3-11}
 & & \textbf{UDP} & \textbf{GMM} & \textbf{$\Delta$} & \textbf{UDP} & \textbf{GMM} & \textbf{$\Delta$} & \textbf{UDP} & \textbf{GMM} & \textbf{$\Delta$} \\ \midrule

\multirow{5}{*}{COCO}
 & Baseline (Clean) & 0.6926 & \textbf{0.6936} & \textcolor{teal}{+0.0011} & 123 & \textbf{116} & \textcolor{teal}{-7} & \textbf{25} & \textbf{25} & 0 \\
 & Resize Low & \textbf{0.6700} & 0.6697 & \textcolor{red}{-0.0003} & 140 & \textbf{133} & \textcolor{teal}{-7} & \textbf{30} & \textbf{30} & 0 \\
 & Resize Medium & \textbf{0.5922} & 0.5886 & \textcolor{red}{-0.0036} & \textbf{225} & 229 & \textcolor{red}{+4} & \textbf{73} & 77 & \textcolor{red}{+4} \\
 & Resize High & \textbf{0.4377} & 0.4313 & \textcolor{red}{-0.0064} & 385 & \textbf{372} & \textcolor{teal}{-13} & \textbf{140} & \textbf{140} & 0 \\
 & Resize Extreme & \textbf{0.2607} & 0.2567 & \textcolor{red}{-0.0040} & 423 & \textbf{415} & \textcolor{teal}{-8} & 142 & \textbf{133} & \textcolor{teal}{-9} \\ \midrule
\multirow{5}{*}{CrowdPose}
 & Baseline (Clean) & \textbf{0.8160} & 0.8134 & \textcolor{red}{-0.0026} & \textbf{155} & 165 & \textcolor{red}{+10} & \textbf{44} & 45 & \textcolor{red}{+1} \\
 & Resize Low & \textbf{0.8092} & 0.8083 & \textcolor{red}{-0.0009} & 163 & \textbf{159} & \textcolor{teal}{-4} & 48 & \textbf{46} & \textcolor{teal}{-2} \\
 & Resize Medium & \textbf{0.7738} & 0.7714 & \textcolor{red}{-0.0024} & \textbf{204} & 206 & \textcolor{red}{+2} & \textbf{68} & 70 & \textcolor{red}{+2} \\
 & Resize High & \textbf{0.6585} & 0.6534 & \textcolor{red}{-0.0051} & 337 & \textbf{330} & \textcolor{teal}{-7} & 109 & \textbf{104} & \textcolor{teal}{-5} \\
 & Resize Extreme & \textbf{0.4358} & 0.4317 & \textcolor{red}{-0.0041} & \textbf{492} & 497 & \textcolor{red}{+5} & 186 & \textbf{181} & \textcolor{teal}{-5} \\ \midrule
\multirow{5}{*}{OCHuman}
 & Baseline (Clean) & \textbf{0.6292} & 0.6270 & \textcolor{red}{-0.0022} & 438 & \textbf{433} & \textcolor{teal}{-5} & 117 & \textbf{116} & \textcolor{teal}{-1} \\
 & Resize Low & \textbf{0.6263} & 0.6232 & \textcolor{red}{-0.0031} & 443 & \textbf{436} & \textcolor{teal}{-7} & \textbf{113} & 118 & \textcolor{red}{+5} \\
 & Resize Medium & \textbf{0.6023} & 0.5985 & \textcolor{red}{-0.0037} & 454 & \textbf{443} & \textcolor{teal}{-11} & 123 & \textbf{122} & \textcolor{teal}{-1} \\
 & Resize High & \textbf{0.5314} & 0.5246 & \textcolor{red}{-0.0068} & 562 & \textbf{557} & \textcolor{teal}{-5} & 166 & \textbf{159} & \textcolor{teal}{-7} \\
 & Resize Extreme & \textbf{0.3544} & 0.3505 & \textcolor{red}{-0.0039} & 781 & \textbf{778} & \textcolor{teal}{-3} & \textbf{261} & 270 & \textcolor{red}{+9} \\ \bottomrule
\end{tabular}%
}
\vspace{2mm}
\caption{ViTPose-Small Ablation on Stochastic Aggregation: Comparison of UDP (Discrete TTA) vs our continuous GMM aggregation across three datasets and five degradation levels (\textbf{UDP}: UDP (Discrete TTA), \textbf{GMM}: Ours (GMM).}
\label{tab:vitpose_supp_tta_comprehensive}
\end{table*}

\begin{table*}[htbp]
\centering
\small
\resizebox{\textwidth}{!}{%
\setlength{\tabcolsep}{4.5pt} 
\begin{tabular}{@{}llcccc|cccc|cccc@{}}
\toprule
\multirow{2}{*}{\textbf{Dataset}} & \multirow{2}{*}{\textbf{Degradation}} & \multicolumn{4}{c|}{\textbf{OKS}} & \multicolumn{4}{c|}{\textbf{Loose Swaps}} & \multicolumn{4}{c}{\textbf{Strict Swaps}} \\ \cmidrule(l){3-14}
 & & \textbf{UDP} & \textbf{GMM} & \textbf{+MRF} & \textbf{$\Delta_{M,G}$} & \textbf{UDP} & \textbf{GMM} & \textbf{+MRF} & \textbf{$\Delta_{M,G}$} & \textbf{UDP} & \textbf{GMM} & \textbf{+MRF} & \textbf{$\Delta_{M,G}$} \\ \midrule

\multirow{5}{*}{COCO}
 & Baseline (Clean) & 0.6926 & 0.6936 & \textbf{0.6944} & \textcolor{teal}{+0.0008} & 123 & \textbf{116} & 120 & \textcolor{red}{+4} & \textbf{25} & \textbf{25} & 27 & \textcolor{red}{+2} \\
 & Resize Low & 0.6700 & 0.6697 & \textbf{0.6712} & \textcolor{teal}{+0.0015} & 140 & \textbf{133} & 137 & \textcolor{red}{+4} & \textbf{30} & \textbf{30} & 31 & \textcolor{red}{+1} \\
 & Resize Medium & \textbf{0.5922} & 0.5886 & 0.5900 & \textcolor{teal}{+0.0015} & \textbf{225} & 229 & 226 & \textcolor{teal}{-3} & \textbf{73} & 77 & 77 & 0 \\
 & Resize High & \textbf{0.4377} & 0.4313 & 0.4327 & \textcolor{teal}{+0.0014} & 385 & \textbf{372} & 377 & \textcolor{red}{+5} & \textbf{140} & \textbf{140} & 145 & \textcolor{red}{+5} \\
 & Resize Extreme & \textbf{0.2607} & 0.2567 & 0.2569 & \textcolor{teal}{+0.0002} & 423 & \textbf{415} & 430 & \textcolor{red}{+15} & 142 & \textbf{133} & 139 & \textcolor{red}{+6} \\ \midrule
\multirow{5}{*}{CrowdPose}
 & Baseline (Clean) & \textbf{0.8160} & 0.8134 & 0.8144 & \textcolor{teal}{+0.0010} & \textbf{155} & 165 & 168 & \textcolor{red}{+3} & \textbf{44} & 45 & 47 & \textcolor{red}{+2} \\
 & Resize Low & \textbf{0.8092} & 0.8083 & 0.8086 & \textcolor{teal}{+0.0003} & 163 & \textbf{159} & 166 & \textcolor{red}{+7} & 48 & \textbf{46} & 51 & \textcolor{red}{+5} \\
 & Resize Medium & \textbf{0.7738} & 0.7714 & 0.7710 & \textcolor{red}{-0.0004} & \textbf{204} & 206 & 212 & \textcolor{red}{+6} & \textbf{68} & 70 & 76 & \textcolor{red}{+6} \\
 & Resize High & \textbf{0.6585} & 0.6534 & 0.6553 & \textcolor{teal}{+0.0019} & 337 & \textbf{330} & 334 & \textcolor{red}{+4} & 109 & \textbf{104} & 110 & \textcolor{red}{+6} \\
 & Resize Extreme & \textbf{0.4358} & 0.4317 & 0.4325 & \textcolor{teal}{+0.0008} & \textbf{492} & 497 & 498 & \textcolor{red}{+1} & 186 & \textbf{181} & 183 & \textcolor{red}{+2} \\ \midrule
\multirow{5}{*}{OCHuman}
 & Baseline (Clean) & \textbf{0.6292} & 0.6270 & 0.6266 & \textcolor{red}{-0.0005} & 438 & \textbf{433} & 435 & \textcolor{red}{+2} & 117 & \textbf{116} & 117 & \textcolor{red}{+1} \\
 & Resize Low & \textbf{0.6263} & 0.6232 & 0.6232 & \textcolor{red}{-0.0000} & 443 & 436 & \textbf{435} & \textcolor{teal}{-1} & \textbf{113} & 118 & 120 & \textcolor{red}{+2} \\
 & Resize Medium & \textbf{0.6023} & 0.5985 & 0.5978 & \textcolor{red}{-0.0007} & 454 & \textbf{443} & 445 & \textcolor{red}{+2} & 123 & \textbf{122} & 124 & \textcolor{red}{+2} \\
 & Resize High & \textbf{0.5314} & 0.5246 & 0.5265 & \textcolor{teal}{+0.0019} & 562 & 557 & \textbf{556} & \textcolor{teal}{-1} & 166 & \textbf{159} & 162 & \textcolor{red}{+3} \\
 & Resize Extreme & \textbf{0.3544} & 0.3505 & 0.3484 & \textcolor{red}{-0.0021} & 781 & 778 & \textbf{775} & \textcolor{teal}{-3} & \textbf{261} & 270 & 273 & \textcolor{red}{+3} \\ \bottomrule
\end{tabular}%
}
\vspace{2mm}
\caption{ViTPose-Small Ablation on Global Structural Priors: Isolating the effect of the MRF structural constraint applied over the continuous GMM + TTA probability distributions (\textbf{UDP}: UDP (Discrete TTA), \textbf{GMM}: Ours (GMM), \textbf{+MRF}: Ours (+MRF)).}
\label{tab:vitpose_supp_mrf_comprehensive}
\end{table*}

\subsubsection{Uncertainty Calibration and OoD Detection}
Tables \ref{tab:vitpose_ause_fusion_deg_optimal} and \ref{tab:vitpose_ause_fusion_final} report In-Distribution calibration performance, demonstrating that Adaptive Softmax Fusion consistently improves the Sparsification Error (AUSE) across transformer-generated activation maps.

\begin{table}[htbp]
\centering
\resizebox{\textwidth}{!}{
\begin{tabular}{l|ccc|ccc|ccc}
\toprule
 & \multicolumn{3}{c|}{\textbf{COCO}} & \multicolumn{3}{c|}{\textbf{CrowdPose}} & \multicolumn{3}{c}{\textbf{OCHuman}} \\
\textbf{Degradation} & \textbf{UDP (Base)} & \textbf{Max-Pool ($\beta$)} & \textbf{Softmax ($\beta$)} & \textbf{UDP (Base)} & \textbf{Max-Pool ($\beta$)} & \textbf{Softmax ($\beta$)} & \textbf{UDP (Base)} & \textbf{Max-Pool ($\beta$)} & \textbf{Softmax ($\beta$)} \\
\midrule
\textbf{00\_Baseline\_Clean} & 0.0604 & 0.0448 (0.21) & \textbf{0.0413} (0.18) & 0.0438 & 0.0435 (1.43) & \textbf{0.0416} (1.94) & 0.1389 & 0.1441 (0.06) & \textbf{0.1368} (0.06) \\
\textbf{01\_Resize\_Low} & 0.0640 & 0.0489 (0.24) & \textbf{0.0445} (0.15) & 0.0458 & 0.0460 (1.66) & \textbf{0.0442} (1.54) & 0.1434 & 0.1471 (0.08) & \textbf{0.1395} (0.07) \\
\textbf{02\_Resize\_Medium} & 0.0766 & 0.0599 (0.31) & \textbf{0.0546} (0.18) & \textbf{0.0570} & 0.0586 (1.54) & \textbf{0.0570} (1.43) & 0.1548 & 0.1544 (0.12) & \textbf{0.1499} (0.08) \\
\textbf{03\_Resize\_High} & 0.1100 & 0.0906 (0.38) & \textbf{0.0843} (0.22) & \textbf{0.0728} & 0.0751 (1.94) & 0.0730 (0.97) & 0.1886 & 0.1815 (0.22) & \textbf{0.1761} (0.15) \\
\textbf{04\_Resize\_Extreme} & 0.1280 & 0.1023 (0.97) & \textbf{0.0980} (0.45) & 0.1079 & 0.1063 (5.28) & \textbf{0.1024} (3.88) & 0.2261 & 0.1959 (0.83) & \textbf{0.1909} (4.89) \\
\bottomrule
\end{tabular}
}
\caption{ViTPose-Small Global AUSE comparison of fusion strategies accompanied by their oracle optimal $\beta$ values.}
\label{tab:vitpose_ause_fusion_deg_optimal}
\end{table}

\begin{table*}[htbp]
\centering
\resizebox{0.8\textwidth}{!}{
\begin{tabular}{llccc}
\toprule
\textbf{Dataset} & \textbf{Degradation} & \textbf{UDP (Base) AUSE} & \textbf{Max-Pool AUSE ($\beta=0.1$)} & \textbf{Softmax AUSE ($\beta=0.1$)} \\
\midrule
\multirow{6}{*}{\textbf{COCO}}
 & Clean & 0.0597 & 0.0460 & \textbf{0.0415} \\
 & Low & 0.0640 & 0.0512 & \textbf{0.0451} \\
 & Medium & 0.0766 & 0.0655 & \textbf{0.0561} \\
 & High & 0.1097 & 0.1017 & \textbf{0.0877} \\
 & Extreme & 0.1277 & 0.1233 & \textbf{0.1055} \\
 & \textit{General} & 0.0670 & 0.0582 & \textbf{0.0506} \\
\midrule
\multirow{6}{*}{\textbf{CrowdPose}}
 & Clean & \textbf{0.0438} & 0.0448 & 0.0442 \\
 & Low & \textbf{0.0456} & 0.0472 & 0.0465 \\
 & Medium & \textbf{0.0567} & 0.0595 & 0.0589 \\
 & High & \textbf{0.0728} & 0.0754 & 0.0746 \\
 & Extreme & \textbf{0.1061} & 0.1088 & 0.1076 \\
 & \textit{General} & \textbf{0.0551} & 0.0569 & 0.0563 \\
\midrule
\multirow{6}{*}{\textbf{OCHuman}}
 & Clean & 0.1386 & 0.1414 & \textbf{0.1376} \\
 & Low & 0.1434 & 0.1440 & \textbf{0.1403} \\
 & Medium & 0.1548 & 0.1544 & \textbf{0.1502} \\
 & High & 0.1844 & 0.1811 & \textbf{0.1726} \\
 & Extreme & 0.2231 & 0.2111 & \textbf{0.1996} \\
 & \textit{General} & 0.1564 & 0.1566 & \textbf{0.1512} \\
\bottomrule
\end{tabular}
}
\caption{ViTPose-Small Sparsification Error (AUSE) using a static $\beta=0.1$.}
\label{tab:vitpose_ause_fusion_final}
\end{table*}

\begin{figure*}[htbp]
    \centering
    \includegraphics[width=\textwidth]{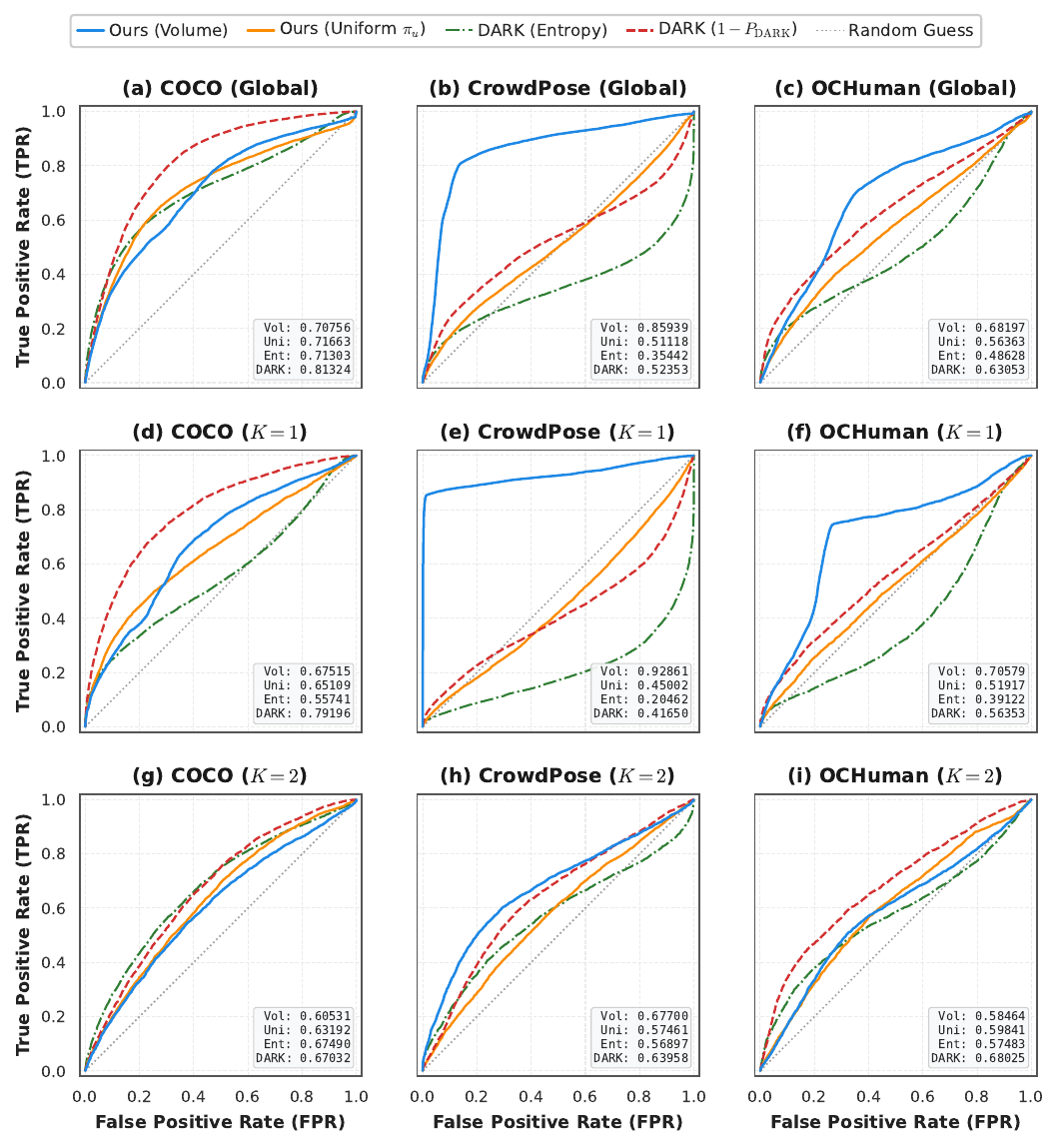}
    \caption{ViTPose-Small Out-of-Distribution (OoD) Anomaly Detection ROC curves decoupled by mixture cardinality ($K=1$ vs. $K=2$). Note that the baseline heuristic curve is labeled as DARK in the plot legend, representing the native UDP confidence ($1 - P_{\text{UDP}}$) in this architecture.}
    \label{fig:vitpose_ood_combined_matrix}
\end{figure*}

\begin{figure*}[htbp]
    \centering
    \includegraphics[width=\textwidth]{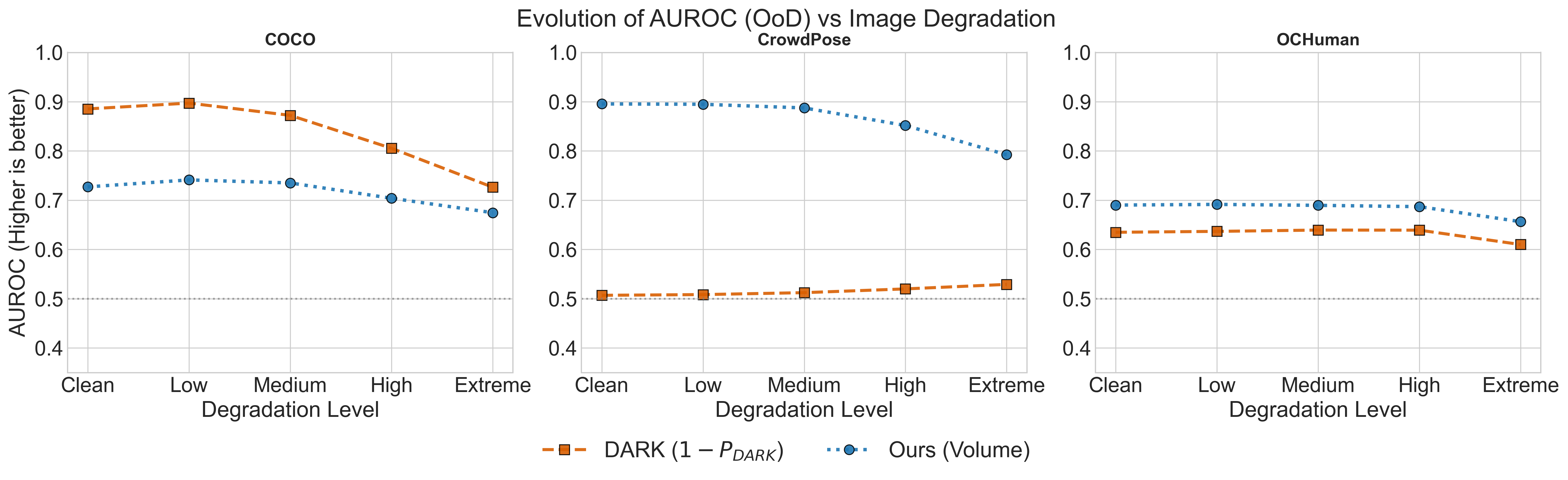}
    \caption{ViTPose-Small: Impact of image degradation (Clean to Extreme) on OoD anomaly detection capabilities (AUROC).}
    \label{fig:vitpose_ood_degradation_evolution}
\end{figure*}

Finally, Figures \ref{fig:vitpose_ood_combined_matrix} and \ref{fig:vitpose_ood_degradation_evolution} evaluate Out-of-Distribution detection on ViTPose-Small. While heuristic confidence of the baseline (\textbf{Baseline ($1 - P$)}, red curve; labeled as DARK in the plot legend) performs well on isolated subjects in COCO, its discriminative capacity drops in dense, overlapping environments. As shown in Figure \ref{fig:vitpose_ood_combined_matrix} (Middle Row, $K=1$), on CrowdPose baseline confidence drops close to random chance due to overconfident detections on neighboring individuals.

In contrast, \textbf{Ours (Volume)} (blue curve) evaluates the spatial dispersion of the attention heatmap, identifying misplaced detections and retaining high AUROC across all evaluated datasets. Furthermore, Figure \ref{fig:vitpose_ood_degradation_evolution} shows that while baseline reliability degrades as spatial resolution decreases, GMM covariance volume maintains a stable anomaly detection signal from clean inputs down to extreme degradation.

\section{Extended Computational Complexity and Profiling Breakdown}
\label{sec:supp_profiling}
This section provides a hardware profiling breakdown of the execution pipeline, detailing the distribution of computational load between GPU tensor operations, host-device PCIe memory transfers, and CPU post-processing stages.

\subsection{Experimental Hardware and Execution Environment}
All computational benchmarks, latency profiles, and memory measurements were conducted on a standardized workstation equipped with:
\begin{itemize}
    \item \textbf{GPU:} NVIDIA GeForce RTX 2080 SUPER (8 GB GDDR6 VRAM) for neural network batch forward tensor operations (CUDA 12.1, PyTorch 2.1.2).
    \item \textbf{CPU:} Intel Core i7-9700K @ 3.60 GHz (8 cores, up to 4.90 GHz Turbo) with 16 GB DDR4 RAM, evaluating CPU execution under Python 3.11 / NumPy 1.26.4.
    \item \textbf{Measurement Protocol:} Latencies are averaged across 25 consecutive inference cycles on standard COCO resolution ($256 \times 192$) following a 10-iteration warm-up phase to reach stable GPU clock frequencies.
\end{itemize}

\subsection{Stage-by-Stage Pipeline Latency Breakdown}
\begin{figure*}[htbp]
    \centering
    \includegraphics[width=\textwidth]{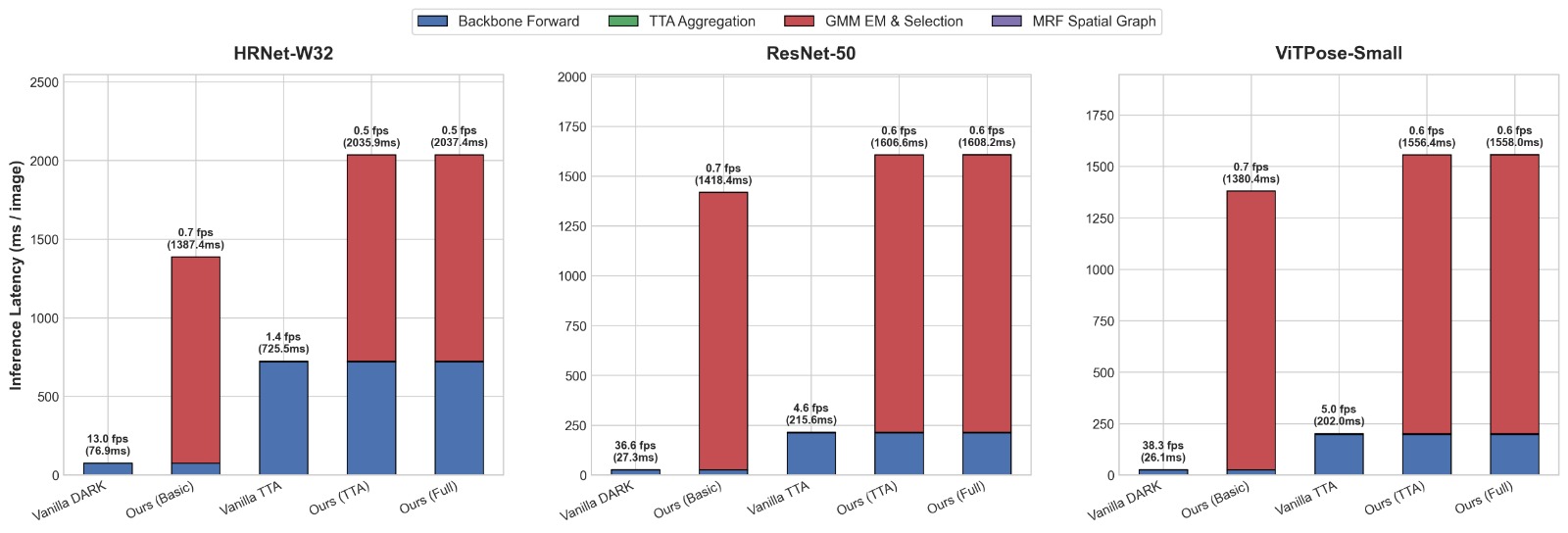}
    \caption{\textbf{Modular latency and throughput breakdown across backbones and execution configurations.} Stacked bar charts illustrate the relative time consumed by Backbone Forward passes (blue), TTA Aggregation (green), GMM EM fitting and model selection (red), and MRF spatial graph decoding (purple). While the single-threaded CPU implementation of the EM loop currently limits throughput to $0.5 - 0.7\text{ FPS}$, batch-tensor GPU parallelization across all 17 joints provides a direct path to higher throughput toward interactive frame rates.}
    \label{fig:supp_computational_breakdown}
\end{figure*}

To characterize performance bottlenecks, the execution flow is divided into seven operational stages (Figure \ref{fig:supp_computational_breakdown}):
\begin{enumerate}
    \item \textbf{Input Preparation and Augmentation Generation (CPU):} The bounding box is scaled ($0.85\times, 0.925\times, 1.0\times, 1.075\times, 1.15\times$) with horizontal flips, generating 10 inference crops ($< 0.5\text{ ms}$).
    \item \textbf{Neural Network Forward Passes (GPU):} Batch tensor inference through the backbone. Single forward passes require $25.6 - 75.7\text{ ms}$, while 10-view TTA batches require $197.9 - 720.9\text{ ms}$ depending on network depth (Table \ref{tab:computational_cost}).
    \item \textbf{Host-to-Device Memory Transfer (PCIe):} Downloading heatmaps to host RAM via asynchronous copy operations consumes $< 0.1\%$ ($< 0.5\text{ ms}$) of total runtime.
    \item \textbf{Continuous Scale-Aware TTA Fusion (CPU):} Applying inverse affine transformations and sigmoidal sharpness confidence weighting across all 10 views takes $\approx 3.2 - 3.6\text{ ms}$.
    \item \textbf{Vectorized Monte Carlo Sampling (CPU):} Von Neumann rejection sampling ($N=1000$ points per joint) with temperature sharpening ($T=0.3$) and uniform sub-pixel dequantization takes $\approx 80 - 100\text{ ms}$ across all 17 keypoints.
    \item \textbf{Sequential Expectation-Maximization and BIC Selection (CPU):} Fitting $17 \times 2 = 34$ candidate models ($K=1$ vs. $K=2$) sequentially in single-threaded Python constitutes the main bottleneck, taking $\approx 1300 - 1390\text{ ms}$ (accounting for $\approx 80.8\%$ of total latency on ResNet-50).
    \item \textbf{Kinematic Tree Belief Propagation (CPU):} MAP message-passing over the COCO skeletal graph takes only $\approx 1.58 - 1.62\text{ ms}$.
\end{enumerate}

\subsection{Memory Footprint and Hardware Scaling}
As demonstrated in Table \ref{tab:computational_cost}, our post-hoc probabilistic decoding requires no additional neural network parameters and maintains a low memory footprint ($118\text{ MB}$ for ViTPose-Small, $247\text{ MB}$ for HRNet-W32, and $265\text{ MB}$ for ResNet-50). The memory overhead is invariant to whether GMM modeling or MRF inference is enabled, showing that continuous uncertainty quantification can be integrated into resource-constrained deployments without increasing peak VRAM requirements.

\begin{table*}[t]
\centering
\small
\caption{\textbf{Computational Efficiency and Modular Latency Breakdown across Pipeline Configurations.} Comparison of inference latency (ms), modular stage breakdown, throughput (FPS), and peak GPU memory (VRAM) across standard backbones on COCO resolution ($256\times 192$). Evaluated under the production execution pipeline matching the experimental benchmark protocol.}

\label{tab:computational_cost}
\resizebox{\textwidth}{!}{%
\begin{tabular}{llccccccc}
\toprule
\textbf{Backbone} & \textbf{Configuration} & \textbf{Forward (ms)} & \textbf{TTA Fuse (ms)} & \textbf{GMM EM (ms)} & \textbf{MRF (ms)} & \textbf{Total Latency (ms)} & \textbf{Throughput (FPS)} & \textbf{Peak VRAM (MB)} \\
\midrule
\textbf{HRNet-W32} & Vanilla DARK (Baseline) & 75.7 & -- & -- & -- & \textbf{76.9 $\pm$ 4.3} & 13.0 & 247 \\
 & Ours (Basic: Single + GMM) & 75.7 & -- & 1311.7 & -- & \textbf{1387.4 $\pm$ 8.6} & 0.7 & 247 \\
 & Vanilla TTA (Multi-scale + DARK) & 720.9 & 3.3 & -- & -- & \textbf{725.5 $\pm$ 10.0} & 1.4 & 247 \\
 & Ours (TTA + Continuous GMM) & 720.9 & 3.3 & 1311.7 & -- & \textbf{2035.9 $\pm$ 12.5} & 0.5 & 247 \\
 & Ours (Full: TTA + GMM + MRF) & 720.9 & 3.3 & 1311.7 & 1.58 & \textbf{2037.4 $\pm$ 12.5} & 0.5 & 247 \\
\midrule
\textbf{ResNet-50} & Vanilla DARK (Baseline) & 26.8 & -- & -- & -- & \textbf{27.3 $\pm$ 17.5} & 36.6 & 265 \\
 & Ours (Basic: Single + GMM) & 26.8 & -- & 1391.6 & -- & \textbf{1418.4 $\pm$ 20.8} & 0.7 & 265 \\
 & Vanilla TTA (Multi-scale + DARK) & 211.9 & 3.2 & -- & -- & \textbf{215.6 $\pm$ 54.1} & 4.6 & 265 \\
 & Ours (TTA + Continuous GMM) & 211.9 & 3.2 & 1391.6 & -- & \textbf{1606.6 $\pm$ 55.3} & 0.6 & 265 \\
 & Ours (Full: TTA + GMM + MRF) & 211.9 & 3.2 & 1391.6 & 1.62 & \textbf{1608.2 $\pm$ 55.3} & 0.6 & 265 \\
\midrule
\textbf{ViTPose-Small} & Vanilla DARK (Baseline) & 25.6 & -- & -- & -- & \textbf{26.1 $\pm$ 8.6} & 38.3 & 118 \\
 & Ours (Basic: Single + GMM) & 25.6 & -- & 1354.9 & -- & \textbf{1380.4 $\pm$ 10.6} & 0.7 & 118 \\
 & Vanilla TTA (Multi-scale + DARK) & 197.9 & 3.6 & -- & -- & \textbf{202.0 $\pm$ 25.2} & 5.0 & 118 \\
 & Ours (TTA + Continuous GMM) & 197.9 & 3.6 & 1354.9 & -- & \textbf{1556.4 $\pm$ 25.9} & 0.6 & 118 \\
 & Ours (Full: TTA + GMM + MRF) & 197.9 & 3.6 & 1354.9 & 1.61 & \textbf{1558.0 $\pm$ 25.9} & 0.6 & 118 \\
\bottomrule
\end{tabular}%
}
\end{table*}

\end{document}